\documentclass{article}
\pdfoutput=1
\usepackage[preprint]{tmlr}

\usepackage{tikz}
\usepackage{comment}
\usepackage{amsmath,amssymb,amsfonts} 
\usepackage{color}
\usepackage[breaklinks]{hyperref}
\usepackage{url}
\usepackage[utf8]{inputenc}
\usepackage{xcolor}
\usepackage{hyperref}
\usepackage{subcaption}
\usepackage{tabularx}
\usepackage{booktabs}
\usepackage{longtable}
\usepackage{xspace}
\usepackage{microtype}
\usepackage{multirow}
\usepackage{wrapfig}
\usepackage{cleveref}
\usepackage{soul} 

\usepackage{natbib}

\crefformat{section}{\S#2\color{blue}#1#3} 
\crefformat{subsection}{\S#2\color{blue}#1#3}
\crefformat{subsubsection}{\S#2#1#3}

\usepackage[T1]{fontenc}    
\usepackage{url}            
\usepackage{nicefrac}       

\usepackage[accsupp]{axessibility}  

\usepackage{amsmath,amsfonts,bm}

\def\eqref#1{equation~\ref{#1}}

\def\1{\bm{1}}

\def\vx{{\bm{x}}}
\def\vy{{\bm{y}}}

\def\mH{{\bm{H}}}
\def\mI{{\bm{I}}}

\def\mK{{\bm{K}}}
\def\mL{{\bm{L}}}

\def\mX{{\bm{X}}}
\def\mY{{\bm{Y}}}

\DeclareMathAlphabet{\mathsfit}{\encodingdefault}{\sfdefault}{m}{sl}
\SetMathAlphabet{\mathsfit}{bold}{\encodingdefault}{\sfdefault}{bx}{n}

\def\sR{{\mathbb{R}}}

\newcommand{\R}{\mathbb{R}}

\DeclareMathOperator*{\argmax}{arg\,max}

\definecolor{lightgray}{rgb}{0.85, 0.85, 0.85}

\newcommand{\wrnte}[2]{\textsc{Wrn-#1-#2}}

\newcommand{\vikash}[1]{}

\newcommand{\deltext}[1]{}

\title{Understanding Robust Learning through the \\ Lens of Representation Similarities}

\author{\name Christian Cianfarani\thanks{Equal Contribution}  \email crc@uchicago.edu \\
        \addr Department of Computer Science\\
        University of Chicago
      \AND
      \name Arjun Nitin Bhagoji$^*$\thanks{Corresponding Author} \email abhagoji@uchicago.edu \\
      \addr Department of Computer Science\\
      University of Chicago
      \AND
      \name Vikash Sehwag$^*$ \email vvikash@princeton.edu\\
      \addr Department of Electrical Engineering \\
      Princeton University
      \AND
      \name Ben Y. Zhao \email ravenben@cs.uchicago.edu \\
      \addr Department of Computer Science\\
      University of Chicago
      \AND
      \name Prateek Mittal \email pmittal@princeton.edu\\
      \addr Department of Electrical Engineering \\
      Princeton University
      \AND
      \name Haitao Zheng \email htzheng@cs.uchicago.edu \\
      \addr Department of Computer Science\\
      University of Chicago
}

\begin{document}

	
\maketitle
\begin{abstract}
  Representation learning, \textit{i.e.} the generation of representations useful
  for downstream applications, is a task of fundamental importance that underlies
  much of the success of deep neural networks (DNNs). Recently, \emph{robustness
  to adversarial examples} has emerged as a desirable property for DNNs, spurring
  the development of robust training methods that account for adversarial
  examples. In this paper, we aim to understand how the properties of
  representations learned by robust training differ from those obtained from
  standard, non-robust training. This is critical to diagnosing numerous salient
  pitfalls in robust networks, such as, degradation of performance on benign
  inputs, poor generalization of robustness, and increase in over-fitting. We
  utilize a powerful set of tools known as representation similarity metrics,
  across three vision datasets, to obtain layer-wise comparisons between robust and
  non-robust DNNs with different training procedures, architectural parameters and
  adversarial constraints. Our experiments highlight hitherto unseen properties of
  robust representations that we posit underlie the behavioral differences of
  robust networks. We discover a lack of specialization in robust networks'
  representations along with a disappearance of `block structure'. We also find
  overfitting during robust training largely impacts deeper layers. These, along
  with other findings, suggest ways forward for the design and training of better
  robust networks.
\end{abstract}

\section{Introduction}\label{sec: intro}

\begin{wrapfigure}{r}{0.4\textwidth}
\vspace{-20pt}
\begin{center}
\includegraphics[width=\linewidth]{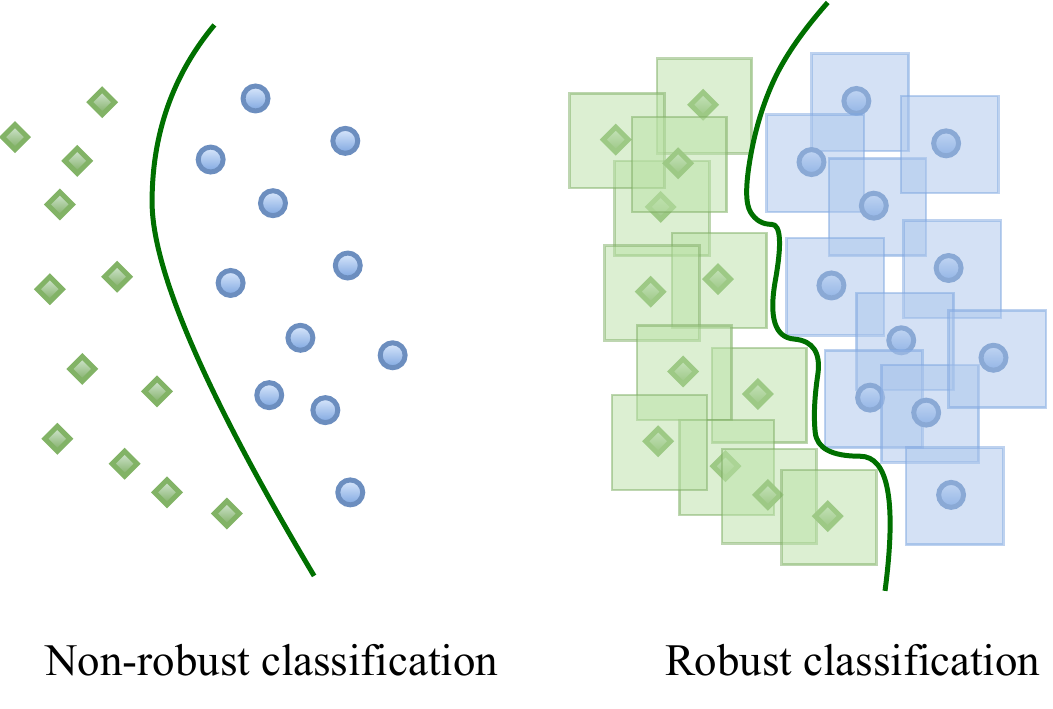}
\end{center}
\caption{\textbf{Non-robust vs robust training.} Adversarially robust training aims to learn decision boundaries that are robust to adversarial perturbations ($\ell_{\infty}$ in this sketch), naturally learning more complex decision boundaries than non-robust training.}
\label{fig: intro_train_viz}
\vspace{-10pt}
\end{wrapfigure}

Representation learning is fundamental to machine learning and a task on which
deep neural networks perform remarkably well. Given complex high-dimensional
input signals, such as images, speech, or text, deep neural networks (DNNs) can
learn meaningful, lower-dimensional representations \citep{krizhevsky2012imagenet, conneau2019unsupervised, vaswani2017attention, devlin2018bert,  hinton2012deep, deng2019arcface, zhou2020graph}. The downstream utility of these representations underlies DNNs' success on different tasks
\citep{brown2020language,dosovitskiy2021an,shen2018natural, mahajan2018exploring, beal2022billion}. For standard image classification, neural networks learn well-separated representations of inputs from different classes to achieve
high accuracy. The discovery of adversarial examples, perturbed inputs that
induce high error in model predictions, even for well-trained classifiers has
motivated the need for another desirable property: \emph{adversarial robustness}. Robust
training methods typically protect against adversarial examples during the
training phase by converting the standard empirical risk minimization objective
to a min-max one (Figure \ref{fig: intro_train_viz}) \citep{madry2017towards, raghunathan2018certified, zhang2019theoretically}. However, they typically
have far worse performance on benign inputs, converge slower and appear to
require larger architectures to be effective \citep{gao2019convergence, nakkiran2019adversarial,tsipras2018robustness,xie_intriguing_2019}. Given the importance of
representation learning for good downstream performance, we posit that an
effective method to identify potential improvements for robust training is via developing a better understanding of robustly learned representations \footnote{We note that we use the word `robust' as shorthand for `adversarially robust' throughout the paper. When considering other types of robustness, such as to common corruptions, we explicitly mention it.}.




Our approach is to juxtapose representations of non-robust and robust  networks and
analyze them through a unified lens. Aggregate metrics like loss and accuracy do
shed some light on the differences, but these are task-driven and do not admit a
layer-wise comparison of representations \citep{andriushchenko2020understanding,gowal2020uncovering}. We need a comprehensive analysis of
the impact of choice of architectural parameters like depth and, nature of threat model and corresponding
attack strength, and the choice of dataset on robustly learned representations
and their link to downstream performance.

To this end, we leverage recent work on representation similiarity (RS) that
enables fine-grained comparisons of representations obtained across all layers
in a deep neural network (DNN) \citep{kornblith2019similarity,raghu2017svcca}.
Our key findings, using both benign and adversarial inputs as probes over CIFAR-10
\citep{krizhevsky2009learning} and the Imagenette and Imagewoof \citep{imagenette} subsets of the Imagenet \citep{imagenet_cvpr09} dataset, are:


\noindent \textbf{Robust representations are less specialized (\cref{sec:
robust_net_benign_reps}):} Robust networks
exhibit much higher similarity across spatially distant layers compared to non-robust
networks, which have far more localized similarity. The latter thus exhibit the expected
specialization in layers, which is unexpectedly missing in the former. Further,
we find reducing the width of the network has a similar effect to increasing the
budget used in robust training, with both changes increasing long-range
similarity among the learned representations. This indicates that the relative
capacity of a network drops as the budget is increased.


\noindent \textbf{Early layers are largely unaffected by adversarial examples (\cref{sec: perturbed_reps}):} The representations of perturbed inputs from non-robust
and robust networks are, as expected, starkly different. On the other hand,
the representations of benign and perturbed inputs from robust networks are
indistinguishable from one another with regards to representation similarity metrics. 

\noindent \textbf{Deeper layers overfit (\cref{sec: time}):} For both
non-robust and robust training, early layers in the network converge within the
first few epochs, and exhibit very little variation in subsequent training. This
indicates they perform similar roles in both benign and robust networks. Deeper
layers converge slower while visiting drastically different representations over
the course of training. These different representations usually occur when
the loss indicates overfitting, leading us to conclude that \emph{deeper layers
overfit during robust training}.

\noindent \textbf{Visually dissimilar adversarial constraints can lead to
similar robust representations (\cref{sec: threats}):} Regardless of the
adversarial constraint, we find a high degree of similarity between the
representations of perturbed and benign inputs for robust networks.
Representations from networks robust to different, but allied (\textit{e.g.}
$\ell_p$), threat models are similar across attack strength. We find surprising
similarities between the representations of visually unrelated threat models at
lower attack strengths.





\noindent \textbf{What do our findings imply?} The lack of specialization in
robust networks even with heavily overparametrized architectures suggests a need
for explicit tools to differentiate between layers during training. Further, the
overfitting we find in deeper layers of neural networks is an indication that
layer-wise specialized regularization methods need to be adopted for robust
training.  We hope the lessons from this paper and the accompanying open-source
code \footnote{\url{https://github.com/uchicago-sandlab/robust_representation_similarity}} spur the development of robust learning-specific architectures and training
methods. 


\noindent \textbf{Paper Layout.} We provide an overview of RS tools and robust
training methods in \cref{sec: background}. We compare representations from
non-robust and robust networks using benign inputs in \cref{sec:
robust_net_benign_reps}. \cref{sec: perturbed_reps} repeats this analysis with
perturbed inputs. The evolution of robust representations over training is
studied in \cref{sec: time} while the impact of different threat models is in
\cref{sec: threats}. \cref{sec: discussion} discusses lessons and limitations.

\section{Background and Setup}\label{sec: background}
We provide a brief overview of the techniques we use to train robust
networks and the tools we use to probe the intermediate
representations of neural networks. Further related work is in Appendix \ref{appsec: extra_background}.


\subsection{Adversarial Examples and Robust Learning} \label{subsec: adv_back}
Adversarial examples are perturbed inputs to machine learning models that
intentionally induce misclassification
\citep{biggio2014security,szegedy2013intriguing}. For an input $\mathbf{x}$, an
untargeted adversarial example with respect to a model $f$ and loss function
$\ell(\cdot)$ is generated by adding a perturbation $\bm{\delta}$ such that
$\bm{\delta} = \argmax_{\mathbf{x}+\bm{\delta} \in N(\mathbf{x})}
\ell(f(\mathbf{x}+\bm{\delta}),y)$, where $y$ is the label of the original class
and $N(\cdot)$ represents the adversarial constraint or neighborhood. The most
commonly used constraints are $L_p$-ball based, although we also experiment with
other unrelated classes of attacks, such as JPEG, Gabor, or Snow based
perturbations \citep{kang2019testing}. We characterize activations of internal
layer in a network as: 1) \textit{Benign representation:} if input is
non-adversarial 2) \textit{Adversarial representations:} if input is an
adversarial example.  


\begin{wraptable}{r}{0.4\textwidth}
	\centering
	\vspace{-10pt}
	\caption{Performance of non-robust and robust  networks on CIFAR-10 dataset, in particular, the high generalization gap in robust network (\wrnte{28}{10}).}
	\def\arraystretch{1.2}
	\resizebox{\linewidth}{!}{
		\begin{tabular}{ccccccc}
			\toprule
			Network  & \multicolumn{3}{c}{ Benign accuracy} &  \multicolumn{3}{c}{Robust accuracy} \\ \midrule
			& Train  & Test  & $\Delta$  & Train  & Test  & $\Delta$  \\
			Non-robust  & $100.0\%$ & $94.9\%$ & $5.1\%$  & $0.0\%$ & $0.0\%$ & $0.0\%$ \\
			Robust & $100.0\%$ & $87.6\%$ & $12.4\%$ & $98.2\%$ & $44.2\%$ & $54.0\%$ \\  \bottomrule
	\end{tabular}}
	\label{tab: cifar10_data}
	\vspace{-10pt}
\end{wraptable}
Adversarial training \citep{madry2017towards} has emerged as the most effective methods to train robust classifiers \citep{croce2020robustbench}, where it accounts for adversarial examples during training. 
It modifies the empirical risk minimization procedure in non-robust training to a min-max optimization, where the inner maximization focuses on finding the strongest adversarial example for each input during training:
\begin{align} \label{eq:adv_training}
\min_\theta \mathbb{E}_{(\mathbf{x},y)\in \mathcal{D}} \left[ \max_{\mathbf{x} + \bm{\delta} \in N(\mathbf{x})} \ell(f_\theta (\mathbf{x} + \bm{\delta}), y) \right],
\end{align}
where $\mathcal{D}$ is a data distribution and $f_\theta$ is a model with parameters $\theta$. Newer methods, like TRADES
\citep{zhang2019theoretically}, attempt to improve on adversarial training by better balancing the tradeoff between robustness and accuracy. We refer to models trained using methods that account for adversarial examples as \textit{robust} models and those trained without these methods as \textit{non-robust} models. 

We refer to accuracy on non-perturbed input as \textit{benign accuracy} and on adversarial examples as \textit{robust accuracy}. Adversarial training is well known to trade-off benign accuracy to gain robustness and exhibits a high generalization gap in robust accuracy (as we show in Table~\ref{tab: cifar10_data} for CIFAR-10 dataset).

\subsection{Probing Neural Network Representations} \label{subsec: rs_back}
Representation similarity (RS) metrics allow for the quantitative comparison of the activations
of different sets of neurons on a given dataset. 
Comparing learned, internal representations within and between neural networks of
different architectures is important for understanding the performance and
behavior of these networks under changing conditions. Several metrics have
 been proposed for this purpose: Canonical Correlation Analysis (CCA)
\citep{hardoon2004canonical} and its extensions, Singular Vector CCA (SVCCA) \citep{raghu2017svcca} and
Projection Weighted CCA (PWCCA) \citep{morcos2018insights}, Centered Kernel Alignment
(CKA) \citep{kornblith2019similarity}, the Orthogonal Procrustes distance
\citep{ding2021grounding} and model stitching
\citep{bansal2021revisiting,csiszarik2021similarity}. These metrics can all
handle different sized representations, but require that the number of
datapoints be the same.

We choose CKA to measure representation similarity, due to its desirable
properties such as orthogonal and isotropic invariance, and wide use in previous
work \citep{neyshabur2020being,nguyen2020wide,Raghu2020Rapid,vulic-etal-2020-probing}. 
\citet{ding2021grounding} have shown that CKA is nevertheless insensitive
to deletions of principal components of representations that strongly correlate
with model accuracy. However, this property does not impact the
manner in which we use CKA. Appendix \ref{appsec:
other_rs} contains experiments with
other RS metrics, yielding similar discoveries.

\noindent \textbf{Centred Kernel Alignment (CKA):} Let $\mX \in \sR^{n \times
p_1}$ with rows $\vx_i$, and $\mY \in \sR^{n \times p_2}$ with rows $\vy_i$ be
the activation matrices of layers with $p_1$ and $p_2$ neurons respectively,
defined on the same set of $n$ points. These activation matrices are then the
representations induced by a particular layer. We also define a $n \times n$
centering matrix $\mH=\mI_n - \bm{1}\bm{1}^\top$. The CKA between these
activation matrices is then
\begin{align}
	\text{CKA}(\mK,\mL) = \frac{\text{tr}(\mK\mH \mL
	\mH)}{\sqrt{\text{tr}(\mK\mH \mK\mH)\text{tr}(\mL\mH \mL\mH)}},
\end{align}
where $\mK_{ij}=k(\vx_i,\vx_j)$, $\mL_{ij}=l(\vy_i,\vy_j)$, with $k$ and $l$
being kernels. When the kernels are linear and the activation matrices are
centered, we get the Linear CKA between activation matrices to be 
\begin{align} \label{eq: linCka}
	\text{Linear CKA} = \frac{ \| \mY^\top \mX \|_F^2}{\| \mX^\top \mX \|_F \| \mY^\top\mY \|_F} =  \frac{ \langle \text{vec}(\mX \mX^\top),\text{vec}(\mY \mY^\top) \rangle}{\| \mX^\top \mX \|_F \| \mY^\top\mY \|_F}.
\end{align}
Since \citet{kornblith2019similarity} find the use of a linear kernel to be as
accurate as and much faster than the RBF kernel, we focus only on linear
kernels. To reduce memory usage, we use online CKA which uses batching to get an
unbiased estimate of CKA \citep{nguyen2020wide}.


\subsection{Experimental Setup} \label{subsec: exp_back} We provide a brief
overview of our experimental setup here. All of our
experiments were run on machines with either NVidia Titan RTX GPUs with 24GB of memory or RTX A4000 GPUs
with 16GB of memory. Further details are in Appendix \ref{appsec: extra_background}.

\noindent \textbf{Models and datasets.} We consider three commonly used image
dataset: CIFAR-10~\citep{krizhevsky2009learning}, ImageNette~\citep{imagenette},
and ImageWoof~\citep{imagenette}, where latter two are subsets of ImageNet
dataset~\citep{deng2009imagenet}, which we use due to the high cost of
adversarial training on the full ImageNet dataset. We use the Wide
Resnet~\citep{zagoruyko2016wide} class of models on CIFAR-10 dataset and
ImageNet-based ResNets for ImageNette and Imagewoof datasets.

%

\noindent \textbf{Adversarial training.} We follow standard convention with $\ell_{\infty}$ perturbations and use $\epsilon = \frac{8}{255}$ for  CIFAR-10 and $\epsilon = \frac{4}{255}$ for ImageNet based datasets. We use 10-step projected gradient descent (PGD) attack during training and use 20 steps at test time to calculate adversarial representations. 
%

\noindent \textbf{Representation similarity.} We will be using mainly using CKA
to compare activation matrices derived from different benign or adversarial
images. In online CKA, we use a batch-size of 1024 and take 3 passes over the
dataset to reduce any stochasticity in the output similarity score. In
layer-wise similarity analyses, we consider all layers, including convolutions,
pooling, activation, and batch-normalization layers.




\begin{figure}[t]
	\centering
	\begin{subfigure}[b]{0.48\textwidth}
		\centering
		\includegraphics[width=\linewidth]{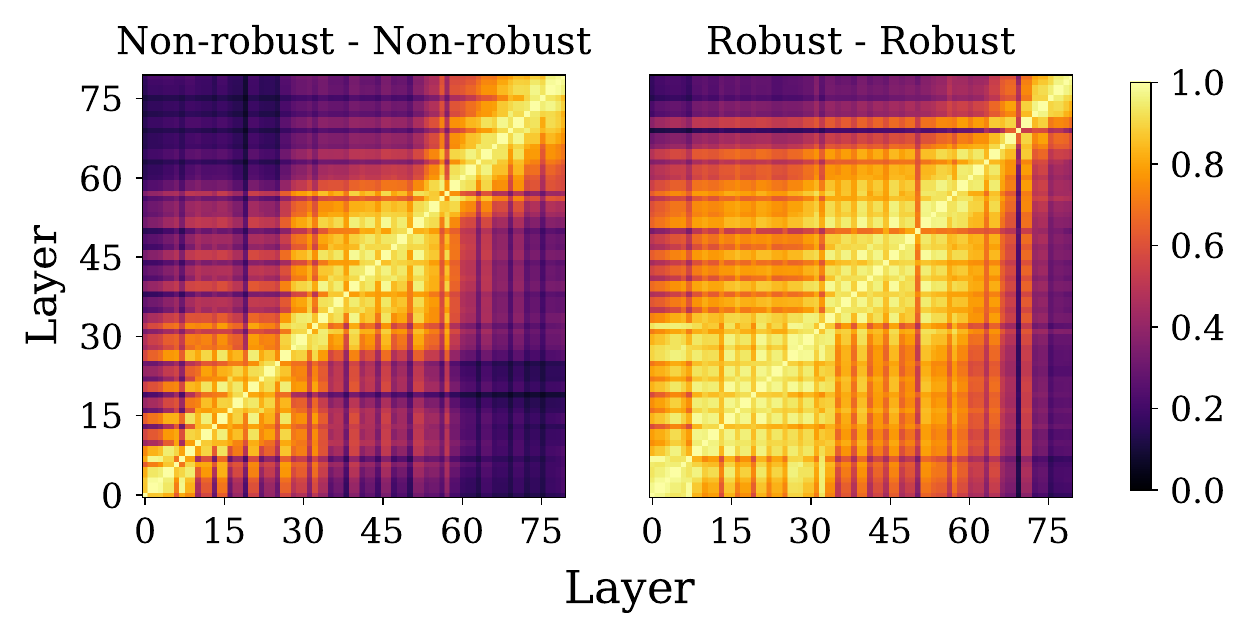}
		\caption{CIFAR-10}
	\end{subfigure}
	\hfill
	\begin{subfigure}[b]{0.48\textwidth}
		\centering
		\includegraphics[width=\linewidth]{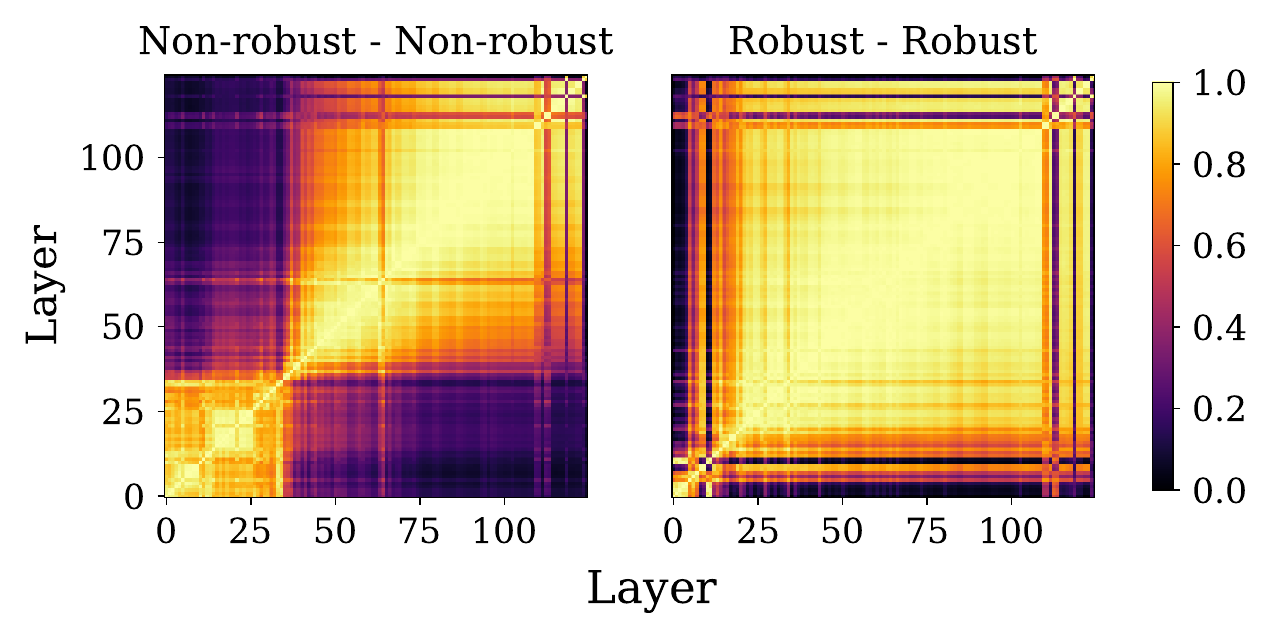}
		\caption{ImageNette}
	\end{subfigure}
	\begin{subfigure}[b]{0.48\textwidth}
		\centering
		\includegraphics[width=\linewidth]{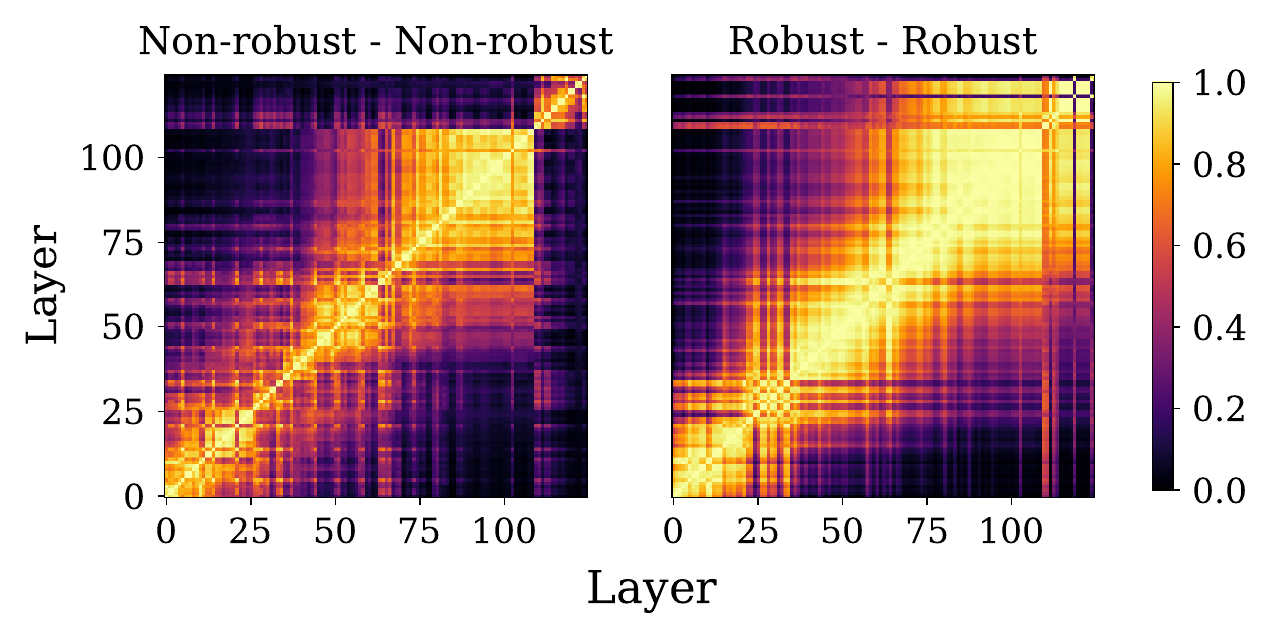}
		\caption{ImageWoof}
	\end{subfigure}
	\hfill
	\begin{subfigure}[b]{0.48\textwidth}
		\centering
		\includegraphics[width=\linewidth]{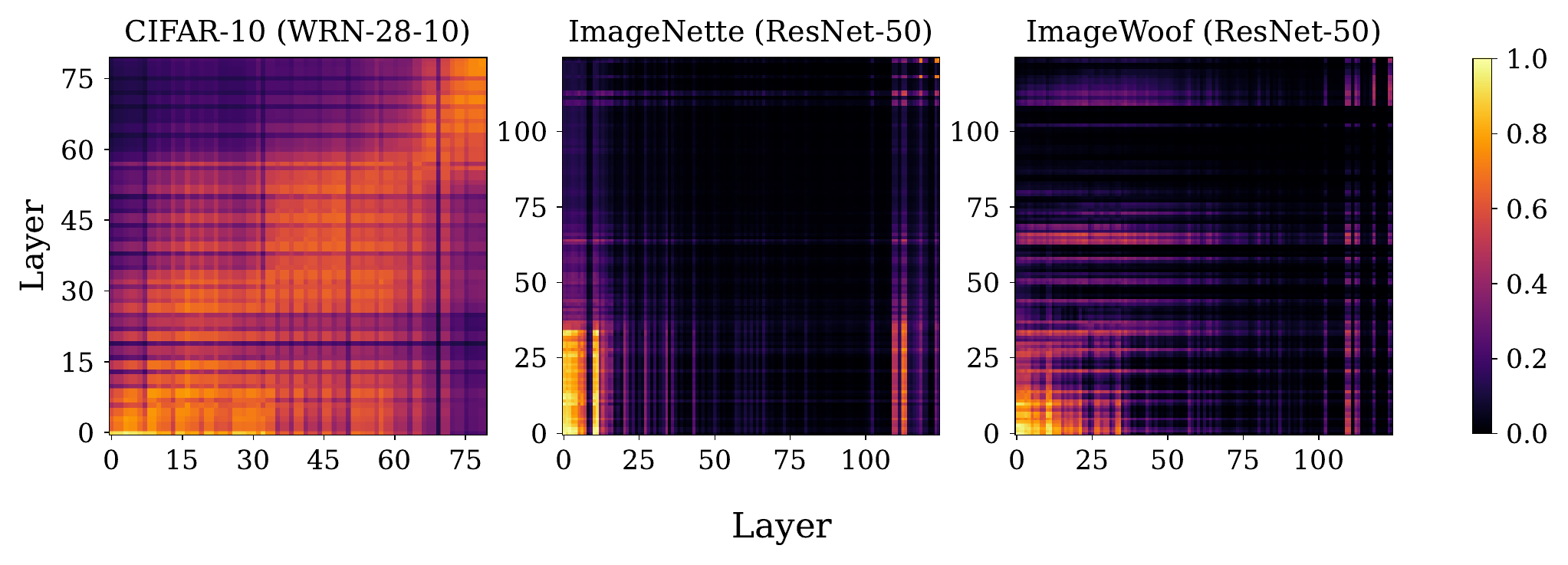}
		\caption{Similarity between representations of Non-robust and robust models.}
		\label{fig: cifar_wrn_base_comp_d}
	\end{subfigure}
	\caption{\textbf{Stark differences in the cross-layer similarity and block structure for robust networks.} We measure the similarity of benign representations across all pairs of layers in both non-robust and robust networks. \textit{(a, b, c):} Robust networks have an amplified degree of cross-layer similarity, even at long-range, which leads to vanishing of the well known block structure characteristic observed in non-robust networks \citep{nguyen2020wide}.\textit{(d)} We only observe a poor similarity between representations of benign and robust models, when experimental setup is identical for both. }
	\label{fig: cifar_wrn_base_comp}
	\vspace{-10pt}
\end{figure}

\section{How do representations from robust and non-robust networks differ?}\label{sec: robust_net_benign_reps}

Robust training is expected to be a harder training objective than non-robust training,
since it optimizes for both accuracy and robustness. The resulting trade-off is
also reflected in final performance, where the benign accuracy of robust networks
is much lower than non-robust networks~\citep{madry2017towards}. In this section,
we aim to understand the impact of robust training on internal representations,
in particular by comparing benign representations in non-robust and robust
networks. Further experiments and results from this section are included in
Appendix \ref{appsec: robust_sim}. We take the following two-fold approach:

\noindent \textbf{1) Comparing characteristics: Absence of block structure.} When comparing layerwise similarity in non-robust networks, numerous works have revealed a `block structure', with high similarity between the layers comprising a block, and low similarity outside it~\citep{kornblith2019similarity, nguyen2020wide, raghu2021vitRS} (Non-robust network plots in Figure~\ref{fig: cifar_wrn_base_comp}).We observe that when moving from benign to robust training, while keeping everything else constant, the \emph{block structure all but disappears}. The lack of a block structure in robustly trained networks is accompanied by the presence of \textit{long-range similarities} between layers, with layers having high similarity to those much farther away in the network (Figure \ref{fig: cifar_wrn_base_comp}).

\noindent \textbf{2) Direct comparison: Poor similarity between non-robust and robust network representations.} To further investigate the unique characteristic of robust networks, we conduct a cross-layer comparison of benign representations in non-robust and robust networks (Figure~\ref{fig: cifar_wrn_base_comp_d}). We find a very low degree of similarity across both network representations, which further shows that robust networks had indeed learned strikingly different representations than non-robust networks. 

We also show that both of these observations generalize across architectures, datasets and threat models. We now delving deeper in \textit{analyzing cross-layer characteristics of robust networks representations.}

\begin{figure*}[t]
	\centering
	\includegraphics[width=0.7\linewidth]{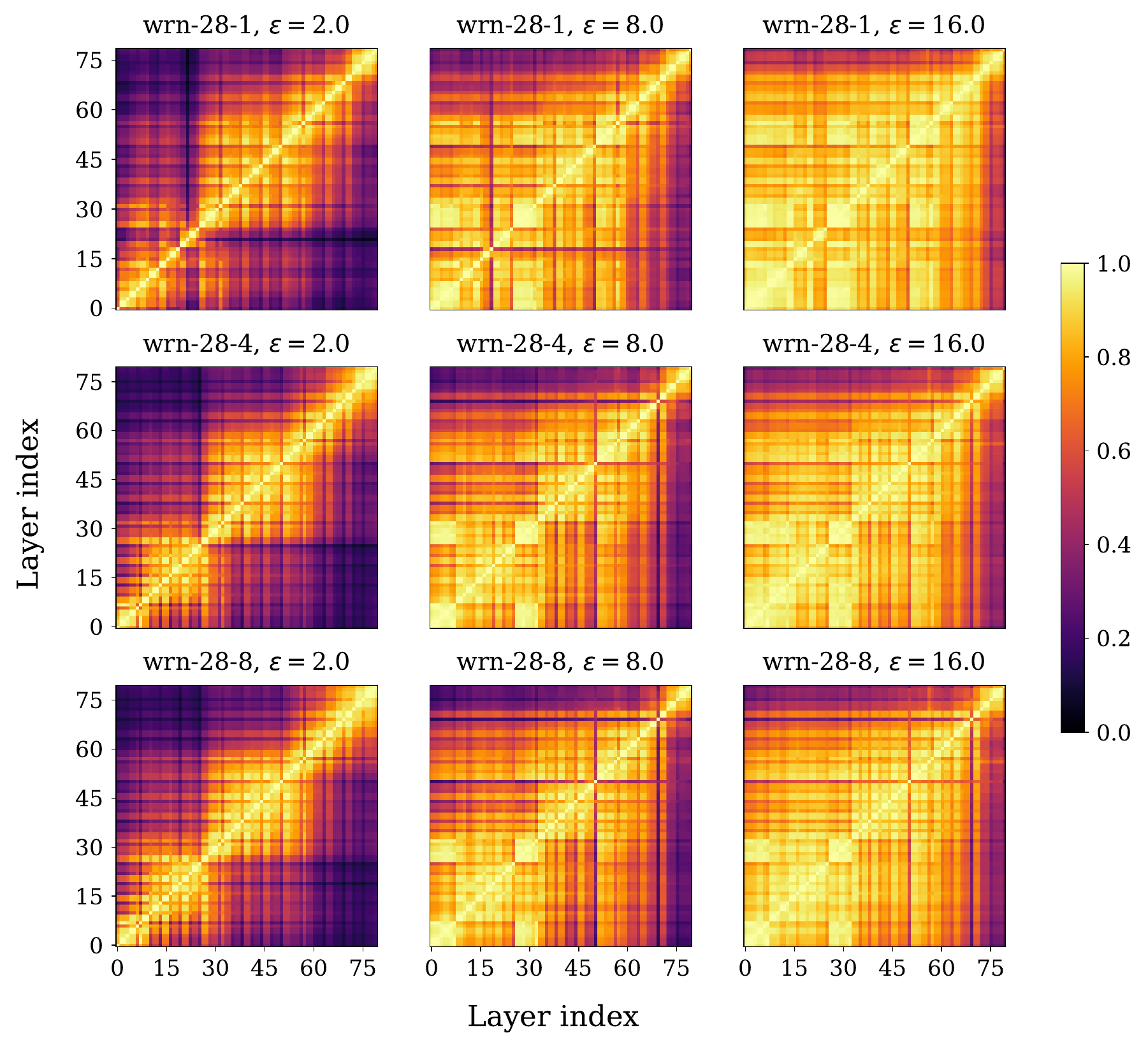}
	\caption{\textbf{Delving deeper into cross-layer similarity.} To better understand the unique cross-layer characteristics in robust networks (Figure~\ref{fig: cifar_wrn_base_comp}), we ablate along two main design choices: 1) Strength of adversarial attacks in robust training (by increasing perturbation budget ($\epsilon$)) 2) Changing network size (by increasing network width). We observe that the strength of adversarial perturbations predominantly leads to amplification of cross-layer similarity, i.e., higher degree of brightness in the plots. Increase network capacity, even by an order of magnitude, doesn't substantially change this phenomenon. We use Wide-ResNet architectures with CIFAR-10 dataset for this experiment. }
	\label{fig: eps_width_vary_cifar10}
	\vspace{-5pt}
\end{figure*}

\noindent \textbf{Impact of robust training strength.}
We investigate this disappearance of block structure in robustly trained
networks further by varying both the adversarial budget ($\epsilon$) used for
training and the width of the network (Figure \ref{fig:
	eps_width_vary_cifar10}). As the training $\epsilon$ increases, the block
structure becomes fainter, eventually disappearing entirely. The block structure
is also impacted by network width, with it being marginally more pronounced in
larger networks. However, this distinction is largely absent at larger values of
$\epsilon$. This establishes that increasing the value of $\epsilon$ is making
the task more complex, so maintaining the same model capacity leads to a
decrease in relative capacity (indicated by high similarity across
representations).


\noindent \textbf{Impact of architecture.} Previously \citet{nguyen2020wide} observed that increasing network width, thus capacity, leads to emergence of block-structure in non-robust networks. Thus we incrementally increase the width of  robust Wide-ResNet architecture on CIFAR-10 dataset and measure the cross-layer representation similarity of benign features (Figure~\ref{fig: eps_width_vary_cifar10}, expanded on in Appendix \ref{appsec: robust_sim}). Even after increasing network capacity by an order of magnitude, we don't see significant variation in the characteristics of robust network representations. These results suggest that while increasing width in robust networks doesn't lead a drastic shift in similarity of internal layer representations.

\section{Probing the impact of adversarial perturbations on representations}\label{sec:
perturbed_reps} 

Robust training aims to learn both high quality benign and robust representations. In the previous section we characterized the impact of robust training on benign representations. In this section, we focus on understanding the characteristics of adversarial representations in robust networks. Detailed results, as well as ablations on architecture and threat model, are available in Appendix \ref{appsec: adv_perturb}. We first analyze how adversarial perturbations distorts the internal representations as we go deeper into the network, i.e., compare adversarial and benign internal representations. Next we analyze whether the distortions introduced by adversarial perturbations can also provide insights into why such perturbations transfer between different networks. 



\begin{figure}[t]
	\centering
	\begin{subfigure}[b]{0.3\linewidth}
		\centering
		\includegraphics[width=\linewidth]{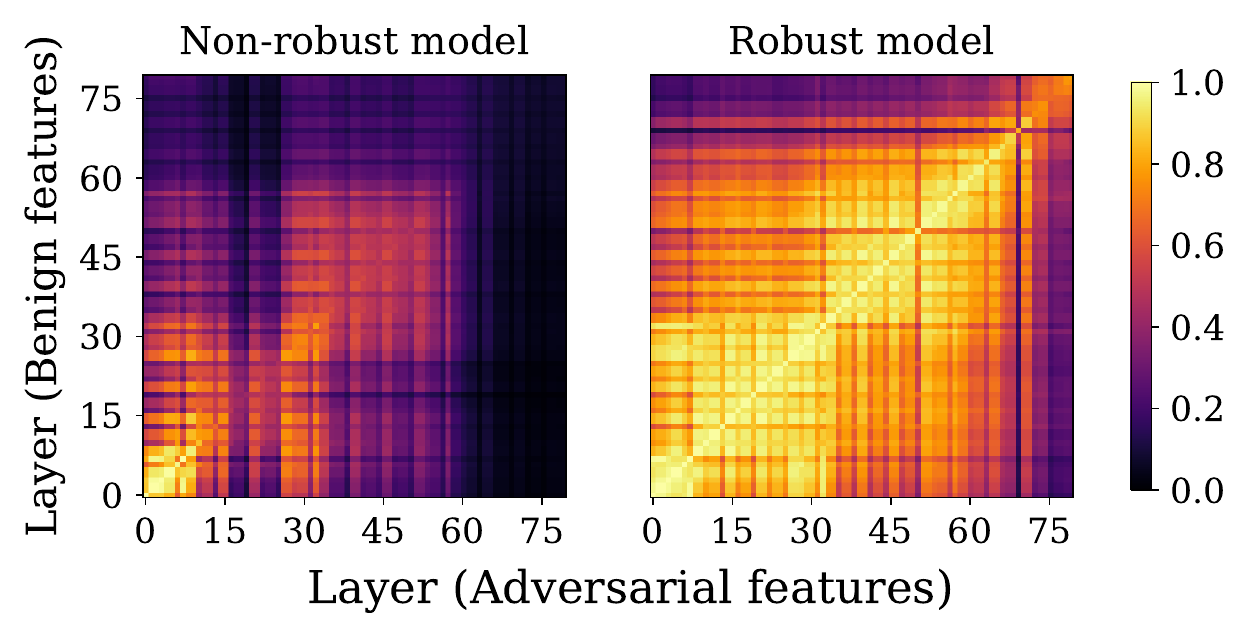}
		\caption{Cross-layer (CIFAR-10)}
		\label{subfig: adv_vs_benign}
	\end{subfigure}
	\hfill 
	\begin{subfigure}[b]{0.22\linewidth}
		\centering
		\includegraphics[width=\linewidth]{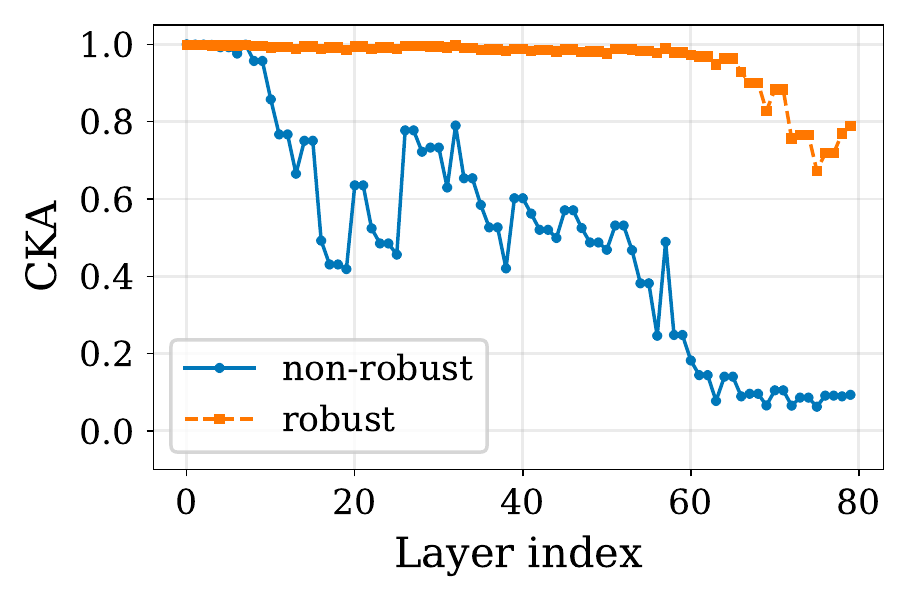}
		\caption{CIFAR-10}
		\label{subfig: divergence_benign_robust_cifar10}
	\end{subfigure}
	\hfill
	\begin{subfigure}[b]{0.22\linewidth}
		\centering
		\includegraphics[width=\linewidth]{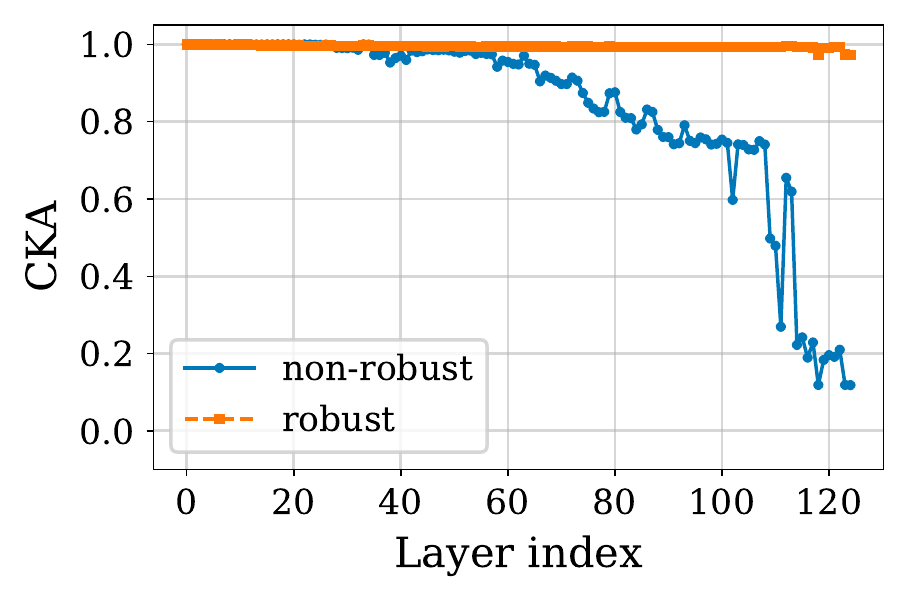}
		\caption{ImageNette}
	\end{subfigure}
	\hfill
	\begin{subfigure}[b]{0.22\linewidth}
		\centering
		\includegraphics[width=\linewidth]{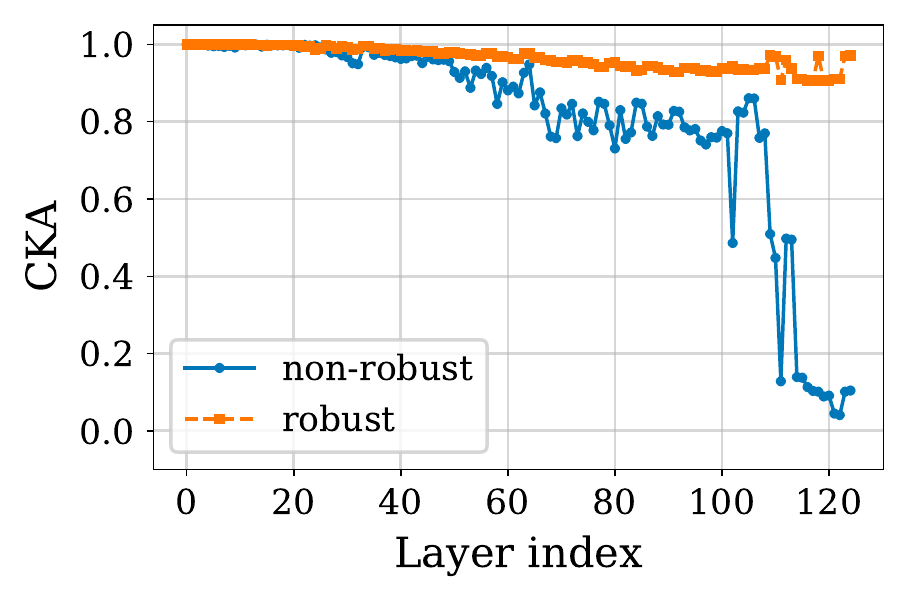}
		\caption{ImageWoof}
	\end{subfigure}
	\caption{\textbf{Divergence of adversarial representations from benign ones.} We compare how adversarial perturbations leads, which intends to heavily model output, distort the internal representations w.r.t. non-adversarial representations. In \textit{(a)} we analyze it across all layers, while in \textit{(b, c, d)} we study it across identical layers, but for different datasets. As expected, adversarial representations in non-robust network are highly dissimilar from benign ones near the end of network. Unexpectedly, such divergence is very small for a non-trivial fraction of early layers (up to one-third). Robust training successfully reduces the impact of adversarial perturbations, thus leading to similar benign and adversarial representations in robust networks.}
	\label{fig: adversarially_perturbed_feature_comparisons}
\end{figure}

\subsection{Impact of adversarial perturbations across layers in non-robust and robust networks}\label{subsec: layerwise_perturbed}
In this experiment, we measure the CKA similarity between benign image representations and corresponding adversarial images representations. 

\noindent \textbf{A fraction of early layers are largely unaffected.} Adversarial perturbation are crafted to corrupt final layer representations, where as expected, the similarity score decays to near zero by final layer (Figure~\ref{fig: adversarially_perturbed_feature_comparisons}). However, we also observe another surprising trend where the adversarial representations shares a high degree of similarity with a fraction (up to one-third for ImageNette and Imagewoof datasets) of early layers.  We find this observation holds across architectures and budgets, indicating that earlier layers representations, even in benign networks, are robust to adversarial perturbations. 
 
\noindent \textbf{Robust networks successfully prevent the distortion of internal representations.}  For robustly trained networks, benign and robust
representations maintain a high degree of similarity throughout the network,
with divergence only occurring in the last 10 layers or so. In our ablation on network size, we see that increasing the size of the network leads
to increased differentiation, implying sufficient capacity for increased
separability between benign and perturbed representations. 



\subsection{Understanding transferability through the lens of internal representation similarity}

\begin{figure}[!htb]
	\centering
	\begin{subfigure}[b]{0.24\linewidth}
		\centering
		\includegraphics[width=\linewidth]{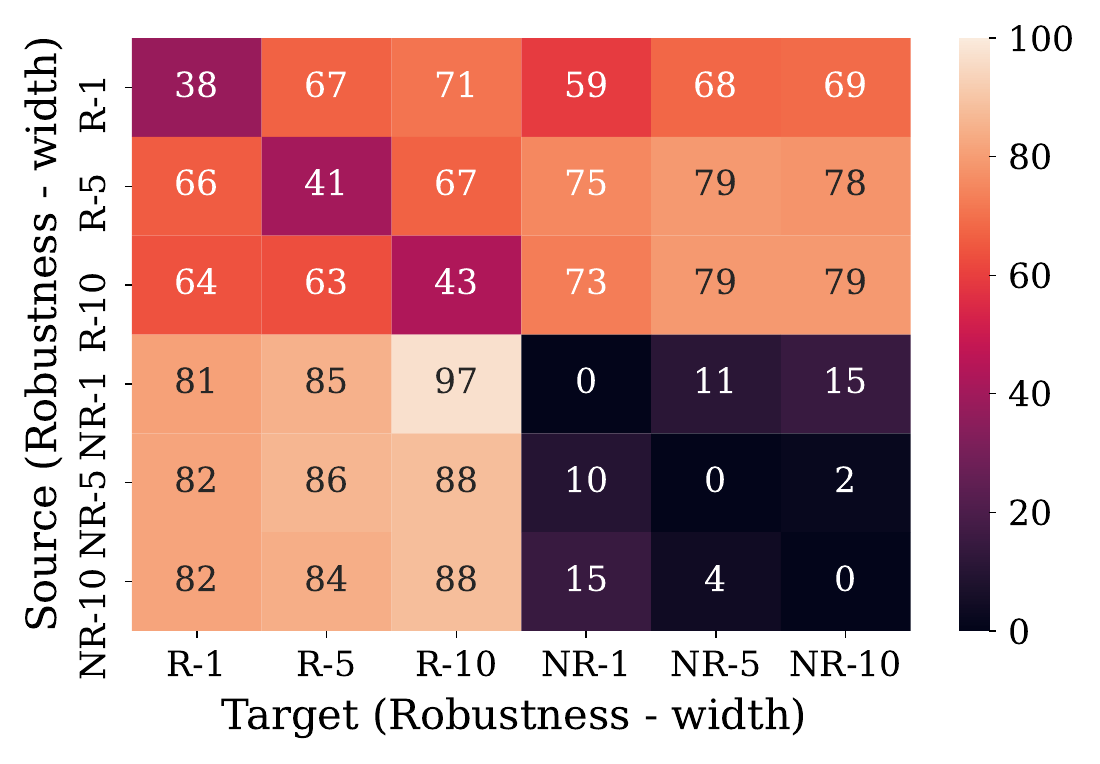}
		\vspace{-15pt}
		\caption{Transfer accuracy of adversarial examples.}
		\label{subfig: transfer_numbers}
	\end{subfigure}
	\hfill
	\begin{subfigure}[b]{0.24\linewidth}
		\centering
		\includegraphics[width=\linewidth]{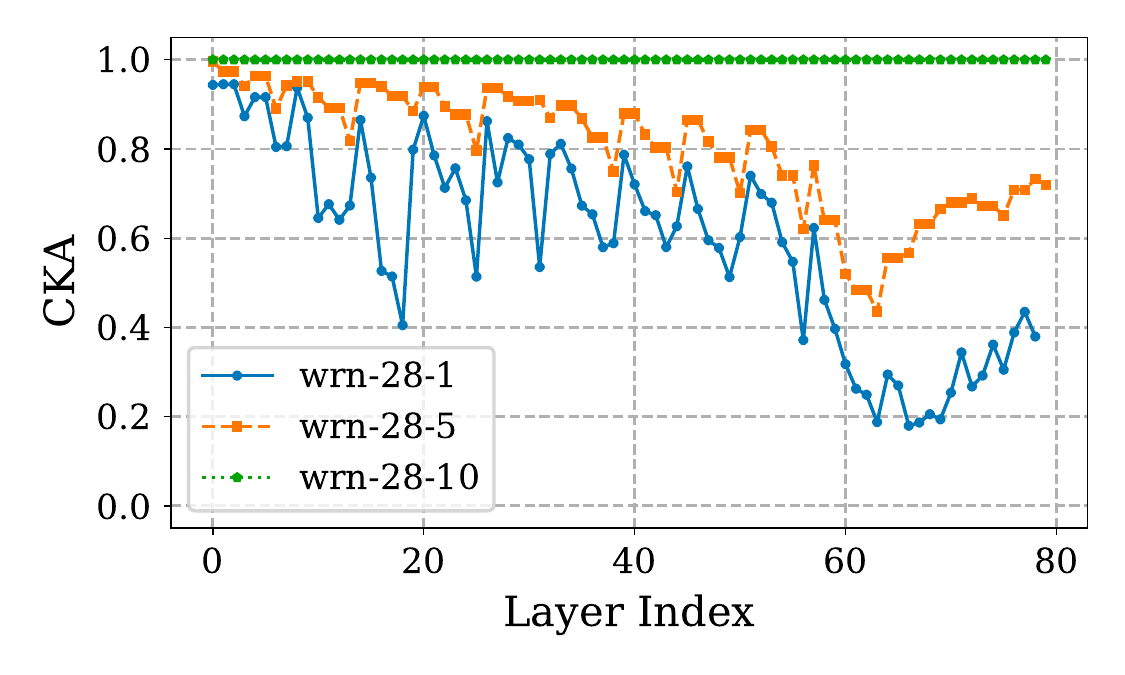}
		\caption{\textit{Non-robust} \wrnte{28}{10} vs. \textit{non-robust} Wide Resnets.}
		\label{subfig: benign_to_benign}
	\end{subfigure}
	\hfill
	\begin{subfigure}[b]{0.24\linewidth}
		\centering
		\includegraphics[width=\linewidth]{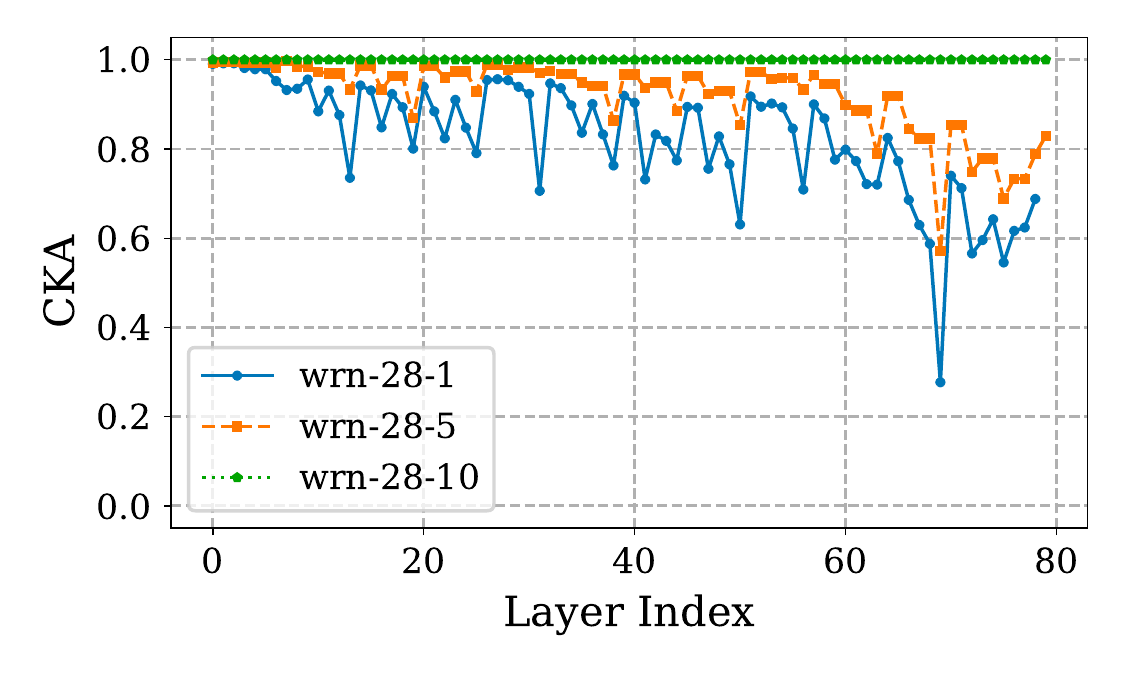}
		\caption{\textit{Robust} \wrnte{28}{10} and other \textit{robust}
		Wide Resnets.}
		\label{subfig: robust_to_robust}
	\end{subfigure}
	\hfill
	\begin{subfigure}[b]{0.24\linewidth}
		\centering
		\includegraphics[width=\linewidth]{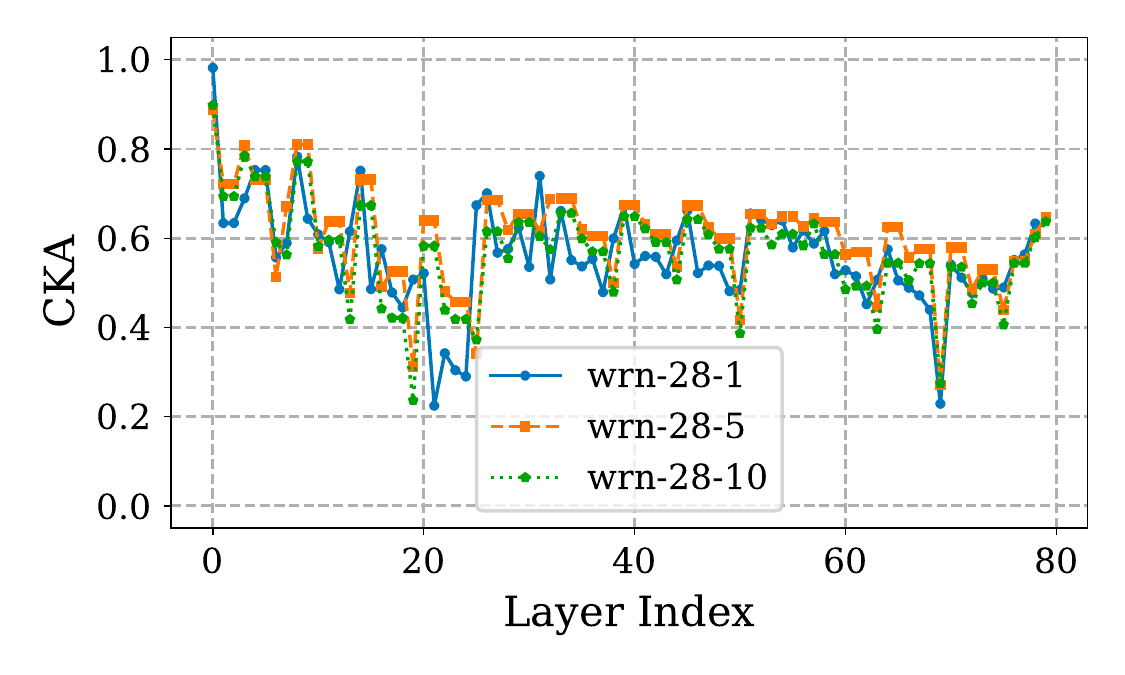}
		\caption{\textit{Robust} \wrnte{28}{10} and other \textit{non-robust}
		Wide Resnets}
		\label{subfig: robust_to_benign}
	\end{subfigure}

	\caption{\textbf{Analyzing cross-architecture representations to understand similarity.} \textit{(a)} Measuring the robust accuracy of adversarial examples transferred among robust (R) and non-robust (NR) Wide-ResNet networks of varying width (e.g., R-5 refers to Robust \wrnte{28}{5}). Low robust accuracy implies high transferability. \textit{(b, c, d)} We measure RS between source and target networks, where these representations are extracted using adversarial examples generated for the source network. When both networks are trained in the same fashion (benign or robust), the layer wise similarity is generally high at earlier layers but drops off towards the end. In contrast, it drops right after the first layer from robust to non-robust network, a set of networks which also have least transferability. 
	}
	\label{fig: transfer_comp}
	\vspace{-10pt}
\end{figure}

It is well-known that adversarial examples are \emph{transferable} from one
architecture to another, i.e., adversarial example generated on one network achieve the misclassification objective on others too. In Figure \ref{subfig: transfer_numbers}, we measure the transferability among non-robust and robust networks. Intriguingly, as previous also validate by \cite{croce2020robustbench}, transferability is higher among some networks (non-robust to non-robust) while much poorer for other (robust to non-robust). High transfer ability indicates that the target network has learned similar decision boundary. We aim to better understand this phenomenon by comparing the internal representations of source and target networks. We first generate adversarial example, and their corresponding internal layer representations, from a source network (\wrnte{28}{10} in this case). We compare it with internal layer representations of these adversarial examples on target network.  We consider transfer from non-robust to robust, robust to robust, and robust to non-robust networks \footnote{Robust networks are known to be highly resilient to adversarial examples generated from non-robust networks (see Figure~\ref{subfig: transfer_numbers}).} (Figure ~\ref{fig: transfer_comp}).

For transfer from non-robust to non-robust or robust-robust, where highest transferability occurs, we observe high degree of similarity between representations at early layers. It indicates that the transferred adversarial examples distort the internal representations in a similar fashion as source network. The most intriguing aspect in Figure \ref{subfig: transfer_numbers} is extremely poor transferability from robust to non-robust networks. We establish that this is due to significant differences in representations where, even across identical architectures, we observe a large drop in  CKA right after input layer.  It shows that adversarial perturbations  from robust nework distorts the internal representations of non-robust network in a strikingly different manner (even at very early layers), which likely renders them ineffective to cause an adversarial effect at the output of the network.

\section{Evolution of internal representation dynamics during robust training}\label{sec: time}
\begin{figure}[!htb]
	\centering
	\includegraphics[width=\linewidth]{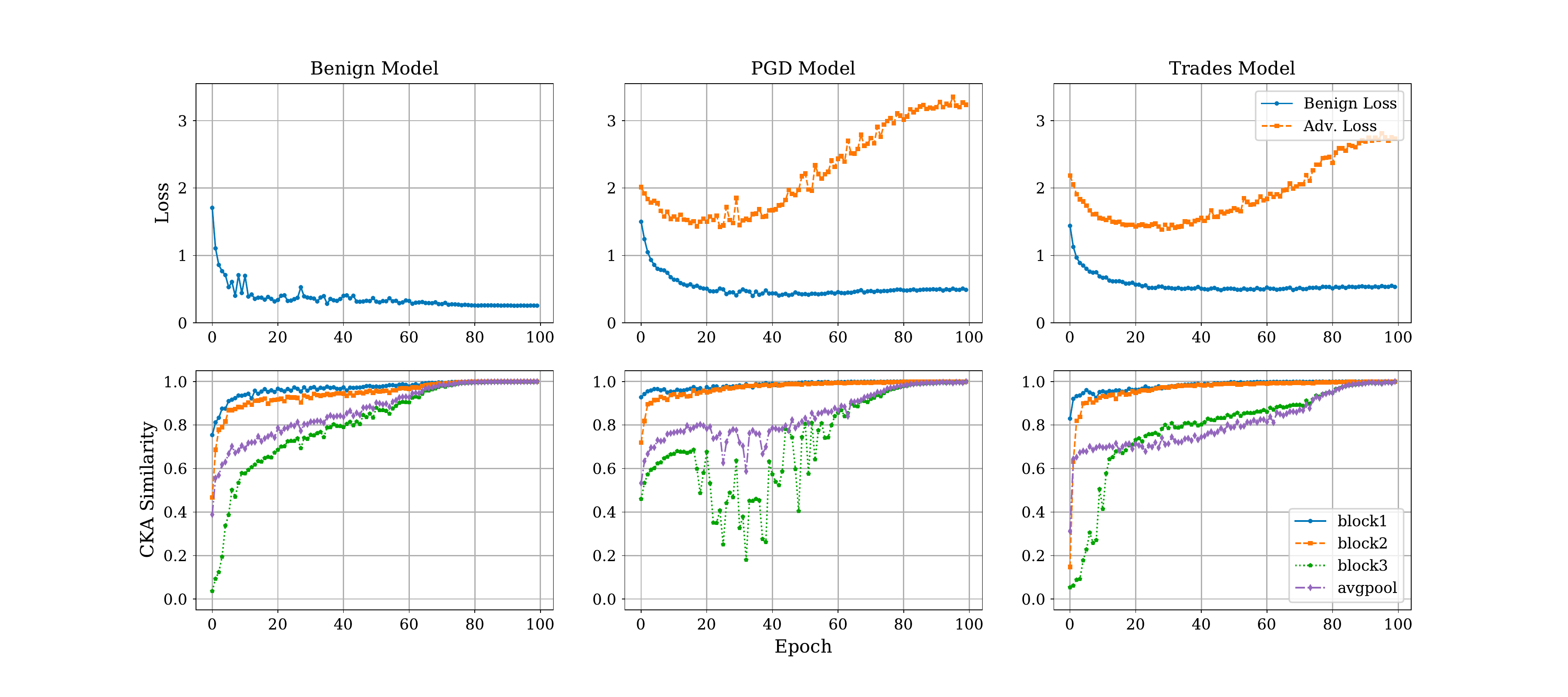}
	\caption{\textbf{Variation in convergence of different layers.} We measure the similarity of each epoch's representations with the fully trained network's representations across both non-robust and robust (both PGD-based and TRADES) training techniques. We observe that representations obtained after the first two blocks converge to their final state very early in the training process and have very little variation in both non-robust and robust network. We also find that later layers exhibit more complex behavior, with signs of overfitting when early stopping is not used, and a lack of stability from one epoch to the next.}
	\label{fig:wrn-28-10_evolve_linf_all}
\end{figure}


In this section, we track the evolution of representations learned by robust
training methods over the course of training epochs in the optimization process.
We study following two convergence properties: 1) Rate of convergence at
different layers, 2) Stability of different layers throughout learning. As in
earlier sections, we use CKA to measure similarity, but in this section we
\emph{compare the benign representations of the same layer across time}. We
present our results for Wide-ResNet architecture in
figure~\ref{fig:wrn-28-10_evolve_linf_all}, where we compare the representations
at each epoch to the final epoch (epoch 100). We further
investigate it across architectures, different layers, and attack budgets in
Appendix \ref{appsec: evolve}. 



\noindent \textbf{Early layers converge faster than later layers.} Benign
representations after the first two blocks of \wrnte{28}{10} are highly similar to
the representation obtained at the end of training, after just ten
epochs (Figure \ref{fig:wrn-28-10_evolve_linf_all}). This is true for all the
training methods considered. In Appendix \ref{appsec: evolve}, we show this largely holds true across architectures and smaller budgets. Our observation relates to the intuition that early layers learn simpler features~\citep{krizhevsky2012imagenet, yosinski2015understanding, zeiler2014visualizing}, thus are learned much faster than complex feature learned by later layers. Only with high budget ($\epsilon$ =  $\frac{16}{255}$), i.e., stronger attacks, we observe a deviation from this trend where second block convergence is slower. 

 \noindent \textbf{Overfitting  is predominantly visible in later layers.} From the cross-epoch
 heatmap of similarities between adversarial representations (Appendix \ref{appsec: evolve}), we observe that later layer representations
 only exhibit similarity off-diagonal at the beginning (until epoch 30) and end
 of training. This clearly shows that multiple local minima are found during
 training. This is true to a less pronounced degree for benign representations. In Fig. \ref{fig:wrn-28-10_evolve_linf_all}, we also observe the adversarial validation loss increasing after around 30
 epochs, while the training loss keeps decreasing, indicating overfitting. After
 this point in training, the CKA similarity between drops with respect to both
 the epoch 30 and final representation drops significantly in later layers,
 while the earlier layers remain unaffected. This strongly indicates that the
 later layers have overfit to the training data. 

\textbf{Adversarial training reduces stability of internal representations during training.} While subsequent epoch representations tend to be very similar to each other in non-robust training, there there can be very large jumps in similarity between one epoch and the next, indicating a less stable convergence. This effect further accentuates when training with a higher perturbation budget or when training a larger network (Appendix \ref{appsec: evolve}).

%

%


\section{How does the threat model impact representations?}\label{sec: threats}
Throughout the paper, we have focused on a single threat model: $\ell_{\infty}$
perturbations. In this section, we extend our analysis of robust representations
to understand what exactly is the link between the visual similarity of threat
models and the representations of models trained to be robust to them. To this
end, we consider \textit{four} additional threat models: $\ell_2$, JPEG, Gabor and Snow
\citep{kang2019testing} (visualized in Appendix \ref{appsec: extra_background}).
We also examine the generalizability of our earlier results across these threat
models. For space reasons, all associated figures from this section are in Appendix \ref{appsec: threat_models}.




\begin{figure}[t]
	\centering
	\includegraphics[width=\linewidth]{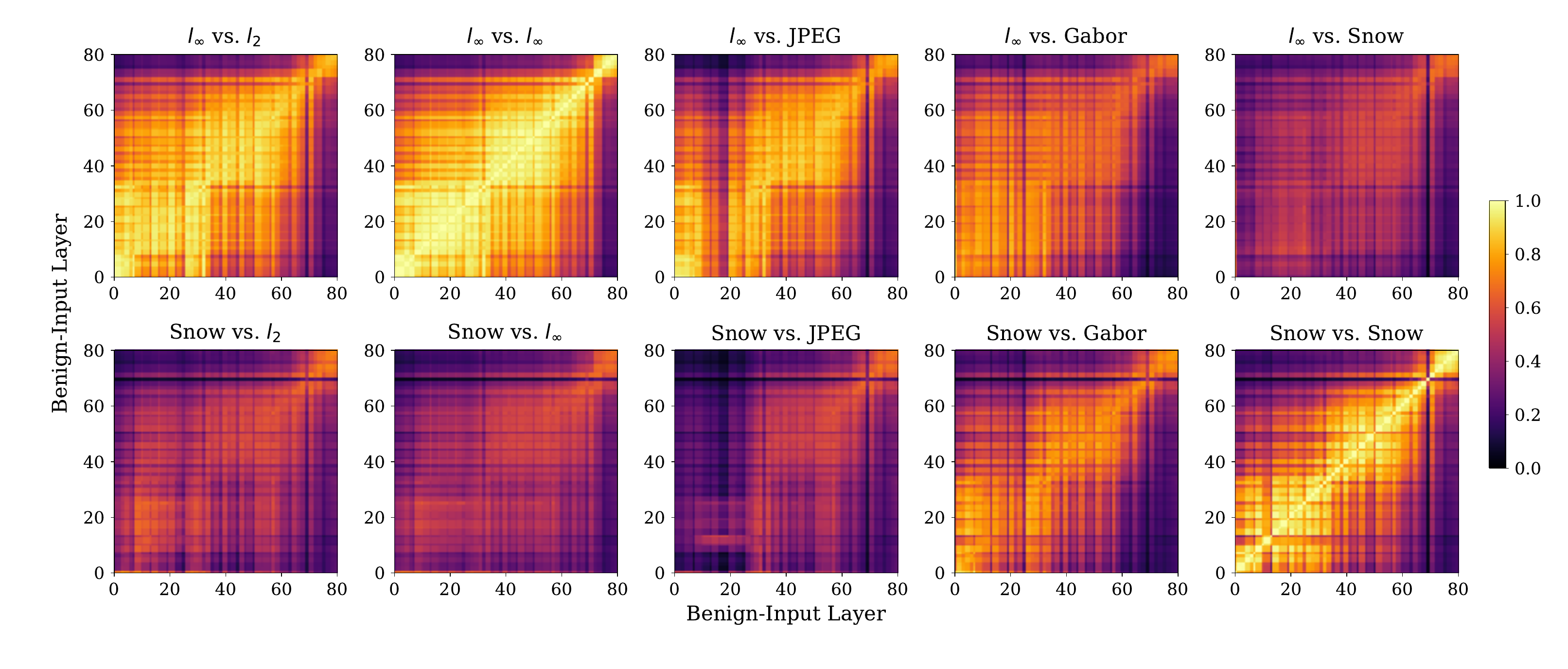}
	\caption{\textbf{Layer-wise similarity plots between different threat models}. The top row shows comparisons between all threat models and $l_\infty$, and the second row shows comparisons between all threat models and Snow. The $\ell_p$ threat models are similar to each other across layers. Among the other threat models, Snow and Gabor display the highest similarity, which is surprising due to large differences in their visual representations.}
	\label{fig:linf_snow}
\end{figure}

\noindent \textbf{Do visually similar attacks lead to similar robust representations?}
RS gives us a method by which we can compare how different
classes of adversarial perturbations affect learned representations. Using CKA,
we plotted the cross-layer similarity of robust WRN-28-10 networks adversarially
trained against five different threat models ($\ell_2$, $\ell_\infty$, JPEG, Gabor,
and Snow) (Figure \ref{fig:linf_snow}). These comparisons lend credence to
conventional knowledge about these threat models. The layerwise similarity plot between $l_2$ and $l_\infty$ robust
models is strikingly similar to CKA plots between robust models trained on the
same threat model. This mirrors theoretical results on correspondences between
$\ell_p$ threat models \citep{tramer2019adversarial}. We also observe a higher average similarity when comparing
an $\ell_\infty$ model with a JPEG model than with Snow or Gabor models. This makes
intuitive sense, given the $\ell_\infty$ constraint inherent in the JPEG attack.
When using CKA to compare against the Snow threat model, we observe that the
highest average similarity is achieved with Gabor. This represents a novel
insight into these threat classes, as correspondence between the two was not
previously known.


\noindent \textbf{Alignment of budgets across different attacks.}
CKA can also give insight into the varying effects of changing perturbation
strengths within different threat models. Expanding on our experiments above (all figures are in Appendix \ref{appsec: threat_models}), we
compared representations between different threat models while varying
perturbation strengths. Between $\ell_\infty$ and Gabor models, the layerwise similarity structure is much more
sensitive to changes in the strength used in $\ell_\infty$-robust training than in
Gabor-robust training. For more closely related attacks, like $\ell_\infty$ and
JPEG, we see similar degradations of the
structure when changing the strength of either attack. Furthermore, the
structure is largely preserved when the attack strengths are changed in tandem.
These results may be useful for developing techniques for training models to be
jointly-robust against different classes of attack.

\noindent \textbf{Layerwise structure across threat models.} We find that across all threat models except Snow, robust and benign activations exhibit high similarity (figures in Appendix \ref{appsec: threat_models}). Further, as we change budgets across the same threat model, we find that $\ell_p$ threat models have representations that evolve more gradually.

\section{Discussion}\label{sec: discussion}
In this paper, we demonstrated the ability of representation similarity metrics to enable a better understanding of robust neural networks. We identified several salient differences between internal representations of non-robust and robust models. We showed that robust models exhibit far higher inter-layer similarity and are prone to overfitting in deeper layers. These observations highlight that robust networks could be improved by encouraging greater differentiation between layers during training as well as targeted reduction of capacity in deeper layers. In addition, we showed why transferability is so low from adversarial examples from robust models to non-robust ones.  Our results on different threat models indicate a path forward for networks robust to multiple attackers as well by allowing for the determination of budgets and layers leading to the best alignment in features.

The key limitations of our study arise from the fundamental disjunct between aggregate properties and layer-wise representation similarity metrics. The latter serve mainly to highlight trends and differences between different models, but do not provide a direct method to improve upon aggregate properties like loss and accuracy. These improvements have to be inferred from a careful analysis of similarities, and future work needs to build upon the observations in this paper for actionable changes to training strategies and architectures.

\clearpage
\bibliographystyle{apalike}
\bibliography{multi_robust}

\begin{thebibliography}{}

\bibitem[Andriushchenko and Flammarion, 2020]{andriushchenko2020understanding}
Andriushchenko, M. and Flammarion, N. (2020).
\newblock Understanding and improving fast adversarial training.
\newblock {\em Advances in Neural Information Processing Systems},
  33:16048--16059.

\bibitem[Bansal et~al., 2021]{bansal2021revisiting}
Bansal, Y., Nakkiran, P., and Barak, B. (2021).
\newblock Revisiting model stitching to compare neural representations.
\newblock {\em Advances in Neural Information Processing Systems}, 34.

\bibitem[Beal et~al., 2022]{beal2022billion}
Beal, J., Wu, H.-Y., Park, D.~H., Zhai, A., and Kislyuk, D. (2022).
\newblock Billion-scale pretraining with vision transformers for multi-task
  visual representations.
\newblock In {\em Proceedings of the IEEE/CVF Winter Conference on Applications
  of Computer Vision}, pages 564--573.

\bibitem[Biggio et~al., 2014]{biggio2014security}
Biggio, B., Corona, I., Nelson, B., Rubinstein, B.~I., Maiorca, D., Fumera, G.,
  Giacinto, G., and Roli, F. (2014).
\newblock Security evaluation of support vector machines in adversarial
  environments.
\newblock In {\em Support Vector Machines Applications}, pages 105--153.
  Springer.

\bibitem[Brown et~al., 2020]{brown2020language}
Brown, T., Mann, B., Ryder, N., Subbiah, M., Kaplan, J.~D., Dhariwal, P.,
  Neelakantan, A., Shyam, P., Sastry, G., Askell, A., et~al. (2020).
\newblock Language models are few-shot learners.
\newblock {\em Advances in neural information processing systems},
  33:1877--1901.

\bibitem[Conneau et~al., 2019]{conneau2019unsupervised}
Conneau, A., Khandelwal, K., Goyal, N., Chaudhary, V., Wenzek, G., Guzm{\'a}n,
  F., Grave, E., Ott, M., Zettlemoyer, L., and Stoyanov, V. (2019).
\newblock Unsupervised cross-lingual representation learning at scale.
\newblock {\em arXiv preprint arXiv:1911.02116}.

\bibitem[Croce et~al., 2020]{croce2020robustbench}
Croce, F., Andriushchenko, M., Sehwag, V., Debenedetti, E., Flammarion, N.,
  Chiang, M., Mittal, P., and Hein, M. (2020).
\newblock Robustbench: a standardized adversarial robustness benchmark.
\newblock {\em arXiv preprint arXiv:2010.09670}.

\bibitem[Croce and Hein, 2021]{croce_adversarial_2021}
Croce, F. and Hein, M. (2021).
\newblock Adversarial robustness against multiple \$l\_p\$-threat models at the
  price of one and how to quickly fine-tune robust models to another threat
  model.
\newblock {\em arXiv:2105.12508 [cs]}.

\bibitem[Csisz{\'a}rik et~al., 2021]{csiszarik2021similarity}
Csisz{\'a}rik, A., K{\H{o}}r{\"o}si-Szab{\'o}, P., Matszangosz, {\'A}., Papp,
  G., and Varga, D. (2021).
\newblock Similarity and matching of neural network representations.
\newblock {\em Advances in Neural Information Processing Systems}, 34.

\bibitem[Deng et~al., 2009]{deng2009imagenet}
Deng, J., Dong, W., Socher, R., Li, L.-J., Li, K., and Fei-Fei, L. (2009).
\newblock Imagenet: A large-scale hierarchical image database.
\newblock In {\em 2009 IEEE conference on computer vision and pattern
  recognition}, pages 248--255. Ieee.

\bibitem[Deng et~al., 2019]{deng2019arcface}
Deng, J., Guo, J., Xue, N., and Zafeiriou, S. (2019).
\newblock Arcface: Additive angular margin loss for deep face recognition.
\newblock In {\em Proceedings of the IEEE/CVF conference on computer vision and
  pattern recognition}, pages 4690--4699.

\bibitem[Devlin et~al., 2018]{devlin2018bert}
Devlin, J., Chang, M.-W., Lee, K., and Toutanova, K. (2018).
\newblock Bert: Pre-training of deep bidirectional transformers for language
  understanding.
\newblock {\em arXiv preprint arXiv:1810.04805}.

\bibitem[Ding et~al., 2021]{ding2021grounding}
Ding, F., Denain, J.-S., and Steinhardt, J. (2021).
\newblock Grounding representation similarity through statistical testing.
\newblock {\em Advances in Neural Information Processing Systems}, 34.

\bibitem[Dosovitskiy et~al., 2021]{dosovitskiy2021an}
Dosovitskiy, A., Beyer, L., Kolesnikov, A., Weissenborn, D., Zhai, X.,
  Unterthiner, T., Dehghani, M., Minderer, M., Heigold, G., Gelly, S.,
  Uszkoreit, J., and Houlsby, N. (2021).
\newblock An image is worth 16x16 words: Transformers for image recognition at
  scale.
\newblock In {\em International Conference on Learning Representations}.

\bibitem[Dwivedi et~al., 2020]{dwivedi2020duality}
Dwivedi, K., Huang, J., Cichy, R.~M., and Roig, G. (2020).
\newblock Duality diagram similarity: a generic framework for initialization
  selection in task transfer learning.
\newblock In {\em European Conference on Computer Vision}, pages 497--513.
  Springer.

\bibitem[Gao et~al., 2019]{gao2019convergence}
Gao, R., Cai, T., Li, H., Hsieh, C.-J., Wang, L., and Lee, J.~D. (2019).
\newblock Convergence of adversarial training in overparametrized neural
  networks.
\newblock {\em Advances in Neural Information Processing Systems}, 32.

\bibitem[Gavrikov and Keuper, 2022]{gavrikov2022adversarial}
Gavrikov, P. and Keuper, J. (2022).
\newblock Adversarial robustness through the lens of convolutional filters.
\newblock {\em arXiv preprint arXiv:2204.02481}.

\bibitem[Gilboa and Gur-Ari, 2019]{gilboa2019wider}
Gilboa, D. and Gur-Ari, G. (2019).
\newblock Wider networks learn better features.
\newblock {\em arXiv preprint arXiv:1909.11572}.

\bibitem[Gowal et~al., 2020]{gowal2020uncovering}
Gowal, S., Qin, C., Uesato, J., Mann, T., and Kohli, P. (2020).
\newblock Uncovering the limits of adversarial training against norm-bounded
  adversarial examples.
\newblock {\em arXiv preprint arXiv:2010.03593}.

\bibitem[Grigg et~al., 2021]{grigg2021self}
Grigg, T.~G., Busbridge, D., Ramapuram, J., and Webb, R. (2021).
\newblock Do self-supervised and supervised methods learn similar visual
  representations?
\newblock {\em arXiv preprint arXiv:2110.00528}.

\bibitem[Hardoon et~al., 2004]{hardoon2004canonical}
Hardoon, D.~R., Szedmak, S., and Shawe-Taylor, J. (2004).
\newblock Canonical correlation analysis: An overview with application to
  learning methods.
\newblock {\em Neural computation}.

\bibitem[Hinton et~al., 2012]{hinton2012deep}
Hinton, G., Deng, L., Yu, D., Dahl, G.~E., Mohamed, A.-r., Jaitly, N., Senior,
  A., Vanhoucke, V., Nguyen, P., Sainath, T.~N., et~al. (2012).
\newblock Deep neural networks for acoustic modeling in speech recognition: The
  shared views of four research groups.
\newblock {\em IEEE Signal processing magazine}, 29(6):82--97.

\bibitem[Howard, 2020]{imagenette}
Howard, J. (2020).
\newblock imagenette.
\newblock \url{https://github.com/fastai/imagenette}.

\bibitem[Huang et~al., 2021]{huang2021exploring}
Huang, H., Wang, Y., Erfani, S., Gu, Q., Bailey, J., and Ma, X. (2021).
\newblock Exploring architectural ingredients of adversarially robust deep
  neural networks.
\newblock {\em Advances in Neural Information Processing Systems}, 34.

\bibitem[Ilyas et~al., 2019]{ilyas2019adversarial}
Ilyas, A., Santurkar, S., Tsipras, D., Engstrom, L., Tran, B., and Madry, A.
  (2019).
\newblock Adversarial examples are not bugs, they are features.
\newblock {\em Advances in neural information processing systems}, 32.

\bibitem[Kang et~al., 2019]{kang2019testing}
Kang, D., Sun, Y., Hendrycks, D., Brown, T., and Steinhardt, J. (2019).
\newblock Testing robustness against unforeseen adversaries.
\newblock {\em arXiv preprint arXiv:1908.08016}.

\bibitem[Kornblith et~al., 2021]{kornblith2021better}
Kornblith, S., Chen, T., Lee, H., and Norouzi, M. (2021).
\newblock Why do better loss functions lead to less transferable features?
\newblock {\em Advances in Neural Information Processing Systems}, 34.

\bibitem[Kornblith et~al., 2019]{kornblith2019similarity}
Kornblith, S., Norouzi, M., Lee, H., and Hinton, G. (2019).
\newblock Similarity of neural network representations revisited.
\newblock In {\em International Conference on Machine Learning}, pages
  3519--3529. PMLR.

\bibitem[Krizhevsky et~al., 2009]{krizhevsky2009learning}
Krizhevsky, A., Hinton, G., et~al. (2009).
\newblock Learning multiple layers of features from tiny images.

\bibitem[Krizhevsky et~al., 2012]{krizhevsky2012imagenet}
Krizhevsky, A., Sutskever, I., and Hinton, G.~E. (2012).
\newblock Imagenet classification with deep convolutional neural networks.
\newblock {\em Advances in neural information processing systems}, 25.

\bibitem[Madry et~al., 2018]{madry2017towards}
Madry, A., Makelov, A., Schmidt, L., Tsipras, D., and Vladu, A. (2018).
\newblock Towards deep learning models resistant to adversarial attacks.
\newblock In {\em ICLR}.

\bibitem[Mahajan et~al., 2018]{mahajan2018exploring}
Mahajan, D., Girshick, R., Ramanathan, V., He, K., Paluri, M., Li, Y.,
  Bharambe, A., and Van Der~Maaten, L. (2018).
\newblock Exploring the limits of weakly supervised pretraining.
\newblock In {\em Proceedings of the European conference on computer vision
  (ECCV)}, pages 181--196.

\bibitem[Morcos et~al., 2018]{morcos2018insights}
Morcos, A., Raghu, M., and Bengio, S. (2018).
\newblock Insights on representational similarity in neural networks with
  canonical correlation.
\newblock {\em Advances in Neural Information Processing Systems}, 31.

\bibitem[Nakkiran, 2019]{nakkiran2019adversarial}
Nakkiran, P. (2019).
\newblock Adversarial robustness may be at odds with simplicity.
\newblock {\em arXiv preprint arXiv:1901.00532}.

\bibitem[Neyshabur et~al., 2020]{neyshabur2020being}
Neyshabur, B., Sedghi, H., and Zhang, C. (2020).
\newblock What is being transferred in transfer learning?
\newblock {\em Advances in neural information processing systems}, 33:512--523.

\bibitem[Nguyen and Hein, 2018]{nguyen2018optimization}
Nguyen, Q. and Hein, M. (2018).
\newblock Optimization landscape and expressivity of deep cnns.
\newblock In {\em International conference on machine learning}, pages
  3730--3739. PMLR.

\bibitem[Nguyen et~al., 2020]{nguyen2020wide}
Nguyen, T., Raghu, M., and Kornblith, S. (2020).
\newblock Do wide and deep networks learn the same things? uncovering how
  neural network representations vary with width and depth.
\newblock {\em arXiv preprint arXiv:2010.15327}.

\bibitem[Raghu et~al., 2020]{Raghu2020Rapid}
Raghu, A., Raghu, M., Bengio, S., and Vinyals, O. (2020).
\newblock Rapid learning or feature reuse? towards understanding the
  effectiveness of maml.
\newblock In {\em International Conference on Learning Representations}.

\bibitem[Raghu et~al., 2017]{raghu2017svcca}
Raghu, M., Gilmer, J., Yosinski, J., and Sohl-Dickstein, J. (2017).
\newblock Svcca: Singular vector canonical correlation analysis for deep
  learning dynamics and interpretability.
\newblock {\em Advances in neural information processing systems}, 30.

\bibitem[Raghu et~al., 2021]{raghu2021vitRS}
Raghu, M., Unterthiner, T., Kornblith, S., Zhang, C., and Dosovitskiy, A.
  (2021).
\newblock Do vision transformers see like convolutional neural networks?
\newblock {\em Advances in Neural Information Processing Systems}, 34.

\bibitem[Raghunathan et~al., 2018]{raghunathan2018certified}
Raghunathan, A., Steinhardt, J., and Liang, P. (2018).
\newblock Certified defenses against adversarial examples.
\newblock In {\em International Conference on Learning Representations}.

\bibitem[Selvaraju et~al., 2017]{selvaraju2017grad}
Selvaraju, R.~R., Cogswell, M., Das, A., Vedantam, R., Parikh, D., and Batra,
  D. (2017).
\newblock Grad-cam: Visual explanations from deep networks via gradient-based
  localization.
\newblock In {\em Proceedings of the IEEE international conference on computer
  vision}, pages 618--626.

\bibitem[Shen et~al., 2018]{shen2018natural}
Shen, J., Pang, R., Weiss, R.~J., Schuster, M., Jaitly, N., Yang, Z., Chen, Z.,
  Zhang, Y., Wang, Y., Skerrv-Ryan, R., et~al. (2018).
\newblock Natural tts synthesis by conditioning wavenet on mel spectrogram
  predictions.
\newblock In {\em 2018 IEEE international conference on acoustics, speech and
  signal processing (ICASSP)}, pages 4779--4783. IEEE.

\bibitem[Simonyan et~al., 2013]{simonyan2013deep}
Simonyan, K., Vedaldi, A., and Zisserman, A. (2013).
\newblock Deep inside convolutional networks: Visualising image classification
  models and saliency maps.
\newblock {\em arXiv preprint arXiv:1312.6034}.

\bibitem[Sundararajan et~al., 2017]{sundararajan2017axiomatic}
Sundararajan, M., Taly, A., and Yan, Q. (2017).
\newblock Axiomatic attribution for deep networks.
\newblock In {\em International conference on machine learning}, pages
  3319--3328. PMLR.

\bibitem[Szegedy et~al., 2013]{szegedy2013intriguing}
Szegedy, C., Zaremba, W., Sutskever, I., Bruna, J., Erhan, D., Goodfellow, I.,
  and Fergus, R. (2013).
\newblock Intriguing properties of neural networks.
\newblock {\em arXiv preprint arXiv:1312.6199}.

\bibitem[Tram{\`e}r and Boneh, 2019]{tramer2019adversarial}
Tram{\`e}r, F. and Boneh, D. (2019).
\newblock Adversarial training and robustness for multiple perturbations.
\newblock In {\em Advances in Neural Information Processing Systems}, pages
  5866--5876.

\bibitem[Tsipras et~al., 2018]{tsipras2018robustness}
Tsipras, D., Santurkar, S., Engstrom, L., Turner, A., and Madry, A. (2018).
\newblock Robustness may be at odds with accuracy.
\newblock {\em arXiv preprint arXiv:1805.12152}.

\bibitem[Vaswani et~al., 2017]{vaswani2017attention}
Vaswani, A., Shazeer, N., Parmar, N., Uszkoreit, J., Jones, L., Gomez, A.~N.,
  Kaiser, {\L}., and Polosukhin, I. (2017).
\newblock Attention is all you need.
\newblock {\em Advances in neural information processing systems}, 30.

\bibitem[Vuli{\'c} et~al., 2020]{vulic-etal-2020-probing}
Vuli{\'c}, I., Ponti, E.~M., Litschko, R., Glava{\v{s}}, G., and Korhonen, A.
  (2020).
\newblock Probing pretrained language models for lexical semantics.
\newblock In {\em Proceedings of the 2020 Conference on Empirical Methods in
  Natural Language Processing (EMNLP)}, pages 7222--7240, Online. Association
  for Computational Linguistics.

\bibitem[Xie and Yuille, 2020]{xie_intriguing_2019}
Xie, C. and Yuille, A. (2020).
\newblock Intriguing properties of adversarial training at scale.
\newblock In {\em ICLR}.
\newblock arXiv: 1906.03787.

\bibitem[Yang et~al., 2021]{yang2021taxonomizing}
Yang, Y., Hodgkinson, L., Theisen, R., Zou, J., Gonzalez, J.~E., Ramchandran,
  K., and Mahoney, M.~W. (2021).
\newblock Taxonomizing local versus global structure in neural network loss
  landscapes.
\newblock {\em Advances in Neural Information Processing Systems}, 34.

\bibitem[Yosinski et~al., 2015]{yosinski2015understanding}
Yosinski, J., Clune, J., Nguyen, A., Fuchs, T., and Lipson, H. (2015).
\newblock Understanding neural networks through deep visualization.
\newblock {\em arXiv preprint arXiv:1506.06579}.

\bibitem[Zagoruyko and Komodakis, 2016]{zagoruyko2016wide}
Zagoruyko, S. and Komodakis, N. (2016).
\newblock Wide residual networks.
\newblock {\em arXiv preprint arXiv:1605.07146}.

\bibitem[Zeiler and Fergus, 2014]{zeiler2014visualizing}
Zeiler, M.~D. and Fergus, R. (2014).
\newblock Visualizing and understanding convolutional networks.
\newblock In {\em European conference on computer vision}, pages 818--833.
  Springer.

\bibitem[Zhang et~al., 2019]{zhang2019theoretically}
Zhang, H., Yu, Y., Jiao, J., Xing, E., El~Ghaoui, L., and Jordan, M. (2019).
\newblock Theoretically principled trade-off between robustness and accuracy.
\newblock In {\em International conference on machine learning}, pages
  7472--7482. PMLR.

\bibitem[Zhou et~al., 2020]{zhou2020graph}
Zhou, J., Cui, G., Hu, S., Zhang, Z., Yang, C., Liu, Z., Wang, L., Li, C., and
  Sun, M. (2020).
\newblock Graph neural networks: A review of methods and applications.
\newblock {\em AI Open}, 1:57--81.

\end{thebibliography}

\newpage

\appendix



In this Appendix, we aim to further validate the discoveries from the main body
of the paper by conducting additional experiments. Our main focus is to ablate
across architectures, representation similarity metrics and threat models to
understand the generalizability of the results presented in the main body. For
specific experiments, we also vary other parameters such as datasets as
necessary.

The Appendix is organized as follows:
\begin{enumerate}
	\item Further background and related work (Section \ref{appsec: extra_background})
  	\item Exploring benign representations from robust
  	networks (Section \ref{appsec: robust_sim})
	\item Analysis of adversarially perturbed representations (Section
	\ref{appsec: adv_perturb})
	\item Impact of training parameters on layer-wise convergence (Section
	\ref{appsec: evolve})
	\item Alignment of different threat models (Section \ref{appsec:
	threat_models})
\end{enumerate}




\section{Additional background and setup details}\label{appsec: extra_background}
\subsection{Representation similarity metrics}

\noindent\textbf{Canonical Correlation Analysis (CCA):} Given activation matrices $X \in \R^{n \times p_1},Y \in \R^{n \times p_2}$,
canonical correlation analysis involves finding the orthogonal bases $w^i_X,
w^i_Y$ such that, after the matrices are projected onto their respective bases,
correlation between the matrices is maximized \citep{kornblith2019similarity}. The relevant summary statistic
for CCA is $\bar{\rho}_{CCA}$, which is defined to be the mean of the canonical
correlation coefficients $\rho_i$, for $1 \leq i \leq p_1$. $\rho_i$ can be
computed by
\begin{align}
	\begin{split}
	\rho_i = &\max_{w^i_X, w^i_Y} \text{corr}(Xw^i_X, Yw^i_Y)\\
	s.t.\quad &\forall_{j<i} Xw^i_X \perp Xw^j_X\\
	&\forall_{j<i} Yw^i_Y \perp Yw^j_Y
	\end{split}
\end{align} 

In order to improve the robustness of CCA, \textbf{Singular Vector Canonical
Correlation Analysis (SVCCA)} has been proposed by \citet{raghu2017svcca}. Rather
than compute CCA on $X$ and $Y$, SVCCA first computes the singular vector
decomposition of $X$ and $Y$ to find representations $X'$ and $Y'$ whose
principle components explain a fixed proportion of the variance of $X$ and $Y$.
Standard CCA is then performed on $X'$ and $Y'$.

\noindent\textbf{Orthogonal Procrustes:} The solution to the Orthogonal
Procrustes problem is the left rotation of $X$ that minimizes the Frobenius norm
between the rotated $X$ and $Y$. Formally, the problem is defined as
\begin{align}
	\min_R \| Y - RX \|^2_F, \: s.t. \: R^\intercal R = I
\end{align}
The Orthogonal Procrustes distance between $X$ and $Y$ is defined as
\begin{align}
	d_{\text{Proc}}(X,Y) = \| X \|^2_F +   \| Y \|^2_F - 2 \| X^\intercal Y \|_*,
\end{align}
where $\| \cdot \|_*$ is the nuclear norm \citep{ding2021grounding}. When $X$ and $Y$ are normalized, this
distance metric lies in the range $[0,2]$. For our purposes, we compute the
Orthogonal Procrustes similarity between normalized matrices $X$ and $Y$,
defined to be
\begin{align}
	S_\text{Proc}(X,Y) = 2 - d_\text{Proc}(X,Y)
\end{align}

\subsection{Threat models}\label{appsubsec: threat_models_back}
\begin{figure}
	\centering
	\includegraphics[width=\linewidth]{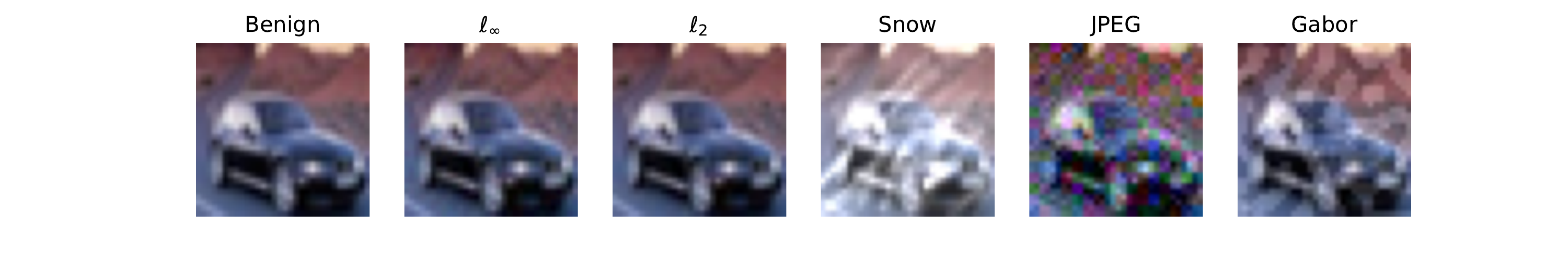}
	\caption{Adversarial examples generated by the threat models used in this paper. \vikash{Lets add an imagenet example too. Its true that we don't test these attacks on imagenet, but large resolution images will better highlight the nature of attack.}}
	\label{fig:adv_examples}
\end{figure}

Most of our robust models were trained and evaluated on $\ell_\infty$ adversarial images. To supplement our results, we also ran experiments using conventional $\ell_2$ attacks, as well as three adversarial attacks introduced by \citet{kang2019testing}: JPEG, Snow, and Gabor. The JPEG attack adds $\ell_p$ bounded adversarial noise to the JPEG-encoded space of images, rather than the pixel space of images. The Snow attack adds adversarially optimized visual occlusions to images that simulate snowfall. The Gabor attack optimizes the parameters of a Gabor kernel that is used to noise the image. Visual examples of these attacks are shown in Figure \ref{fig:adv_examples}. Unless otherwise noted, for our CIFAR-10 experiments we run the $\ell_2$ attack with $\epsilon = 1$, the JPEG attack with $\epsilon = 25$, the Snow attack with $\epsilon = 2$, and the Gabor attack with $\epsilon = 75$. Table \ref{tab:threat_model_acc} has the training and validation accuracies for models robustly trained using these threat models.

\begin{table}
	\centering
\begin{tabular}{lrrrr}
	\toprule
	&  \multicolumn{2}{c}{Benign Images} & \multicolumn{2}{c}{Adversarial Images} \\
	\cmidrule(lr){2-3}
	\cmidrule(lr){4-5}
	Threat Model   &   Train Accuracy &   Val. Accuracy &   Train Accuracy &   Val. Accuracy \\
	\midrule
	Base           &                  100.00 &                  94.92 &           -            &           -          \\
	$\ell_\infty$ ($\epsilon = \frac{8}{255}$)  &                  100.00 &                  86.36 &                 99.03 &               46.34 \\
	$\ell_2$ ($\epsilon = 1$)       &                  100.00 &                  86.49 &                 99.40 &               45.76 \\
	JPEG ($\epsilon = 25$)           &                  100.00 &                  84.65 &                 91.98 &               45.52 \\
	Gabor ($\epsilon = 75$)         &                  100.00 &                  93.12 &                 99.83 &               92.47 \\
	Snow ($\epsilon = 2$)           &                  100.00 &                  90.23 &                 95.72 &               67.35 \\
	\bottomrule
\end{tabular}
\caption{Training and validation accuracies of the \wrnte{28}{10} models used in this paper's experiments. The threat model represents the attack that each model was trained to be robust against, and adversarial images were generated using the same threat model and attack strength as was used in training.}
\label{tab:threat_model_acc}
\end{table}

\begin{table}
	\centering
	\begin{tabular}{rrrrrrr}
		\toprule
		&  \multicolumn{2}{c}{Non-Robust Model} & \multicolumn{4}{c}{$\ell_\infty$-Robust Model} \\
		\cmidrule(lr){2-3}
		\cmidrule(lr){4-7}
		& \multicolumn{2}{c}{Benign Images} & \multicolumn{2}{c}{Benign Images} & \multicolumn{2}{c}{Adversarial Images}\\
		\cmidrule(lr){2-3}
		\cmidrule(lr){4-5}
		\cmidrule(lr){6-7}
		Width &   Train Acc. &   Val. Acc. &   Train Acc. &   Val. Acc. &   Train Acc. &   Val. Acc. \\
		\midrule
		1 &            99.68 &           91.23 &            89.21 &           82.39 &            50.90 &           42.92 \\
		2 &            99.97 &           93.44 &            97.73 &           84.91 &            66.91 &           44.02 \\
		3 &            99.99 &           94.18 &            99.72 &           85.53 &            80.78 &           43.74 \\
		4 &           100.00 &           94.33 &            99.98 &           85.46 &            89.64 &           44.13 \\
		5 &           100.00 &           94.66 &           100.00 &           85.97 &            93.42 &           45.38 \\
		6 &           100.00 &           94.86 &           100.00 &           86.40 &            95.79 &           45.70 \\
		7 &           100.00 &           94.59 &           100.00 &           86.87 &            97.38 &           46.79 \\
		8 &           100.00 &           94.50 &           100.00 &           86.90 &            98.24 &           47.15 \\
		9 &           100.00 &           94.98 &           100.00 &           87.39 &            98.68 &           48.11 \\
		10 &           100.00 &           94.92 &           100.00 &           87.59 &            98.90 &           48.89 \\
		\bottomrule
	\end{tabular}
	\caption{Training and validation accuracies of the \wrnte{28}{N} models used in this paper's experiments, where N is the width factor. Robust models were trained on the $\ell_\infty$ threat model with $\epsilon = \frac{8}{255}$.}
\end{table}

\subsection{Related Work}\label{appsec: rel_work} 

\noindent \textbf{Tools to
probe neural networks.} These can be sorted broadly into two categories: i) \textit{sample-based} and ii)
\textit{distribution-based} tools. Sample-based tools include techniques like input saliency maps \citep{simonyan2013deep}, Grad-CAM \citep{selvaraju2017grad}, integrated gradients \citep{sundararajan2017axiomatic} etc. which are focused on interpreting the output of neural networks for specific inputs. On the other hand, loss surface visualizations \citep{yosinski2015understanding,nguyen2018optimization,yang2021taxonomizing} and representation similarity metrics \citep{dwivedi2020duality,kornblith2019similarity,bansal2021revisiting,raghu2017svcca,csiszarik2021similarity} consider aggregate network properties derived from a set of samples.

\noindent \textbf{Visualizing and validating network properties:}
\citet{gilboa2019wider} use activation atlases to argue that the learned
representations of wider networks are `more informative' than shallower ones. To
validate this assertion, they fine-tune a linear layer for a new task and find
the wider representations perform better. \citet{raghu2021vitRS} study how transformers and CNNs learn features differently, \citet{grigg2021self} analyze the differences between supervised and self-supservised representations and \citet{kornblith2021better} consider the link between loss functions and feature transfer. \citet{gavrikov2022adversarial} use convolutional filters to understand the impact of robust training while \citet{ilyas2019adversarial} visualize what input features are needed for robust training.

\section{How are benign representations from robust
networks different?}\label{appsec: robust_sim}
In this section, we provide additional details on how the benign representations from robustly trained networks differ from those obtained from non-robust networks (see \S3 of the main paper). We ablate across RS metrics (\ref{appsec:
other_rs}), datasets (\ref{appsec: other_datasets}), robust training methods (\ref{appsec: training_method}) and architectures (\ref{appsec: cross_archs}). We also delve deeper into the link between RS metrics and classification accuracy in \ref{appsec:class_cond_cka}.

\begin{figure}[!htb]
	\centering
	\includegraphics[width=0.9\linewidth]{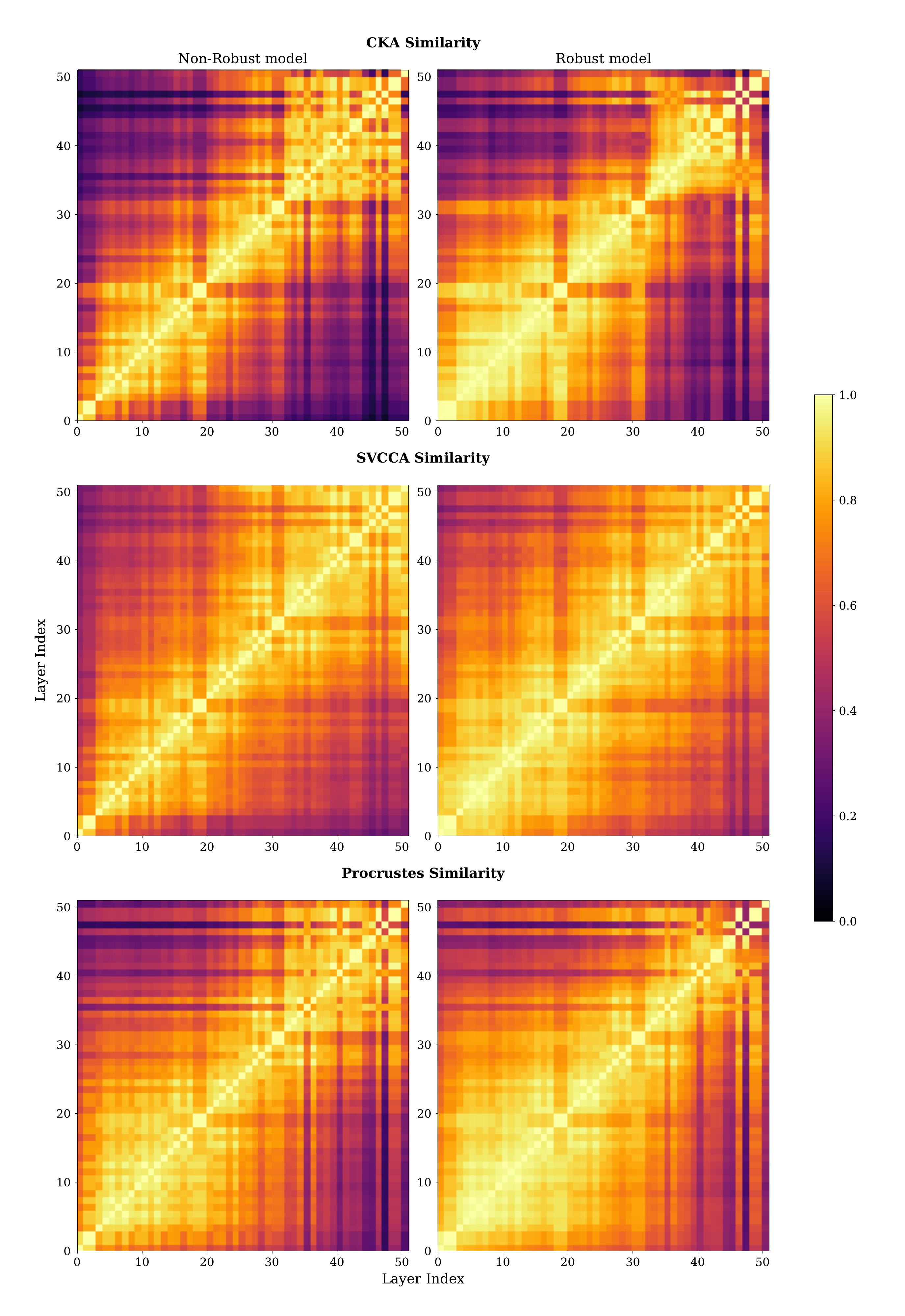}
	\caption{\textbf{Layerwise similarity plots of non-robust and robust ResNet18 models}. 3 different RS metrics are used: CKA, SVCCA, and Orthogonal Procrustes similarity over benign representations.}
	\label{fig:metric_compare}
\end{figure}

\subsection{Using other representation similarity metrics} \label{appsec:
other_rs} 

In Figure \ref{fig:metric_compare}, we plot the layerwise similarity between the
layers for benign representations from both a non-robust and robust model. Due
to computational constraints when computing SVCCA and Orthogonal Procrustes,
comparisons are shown using ResNet18 models, instead of the Wide ResNet models
used throughout much of the rest of the paper. We note that the trend we observe
of increased cross-layer similarity for robustly trained networks appears regardless of the RS metric used. Further, CKA leads to the most visually distinct structure. The broad takeaways from the different RS metrics are aligned enough that we focus on CKA for the remainder of the results in this Appendix.

\subsection{Using other datasets} \label{appsec: other_datasets}

\begin{figure}[!htb]
	\centering
	\begin{subfigure}[b]{0.45\linewidth}
		\centering
		\includegraphics[width=\linewidth]{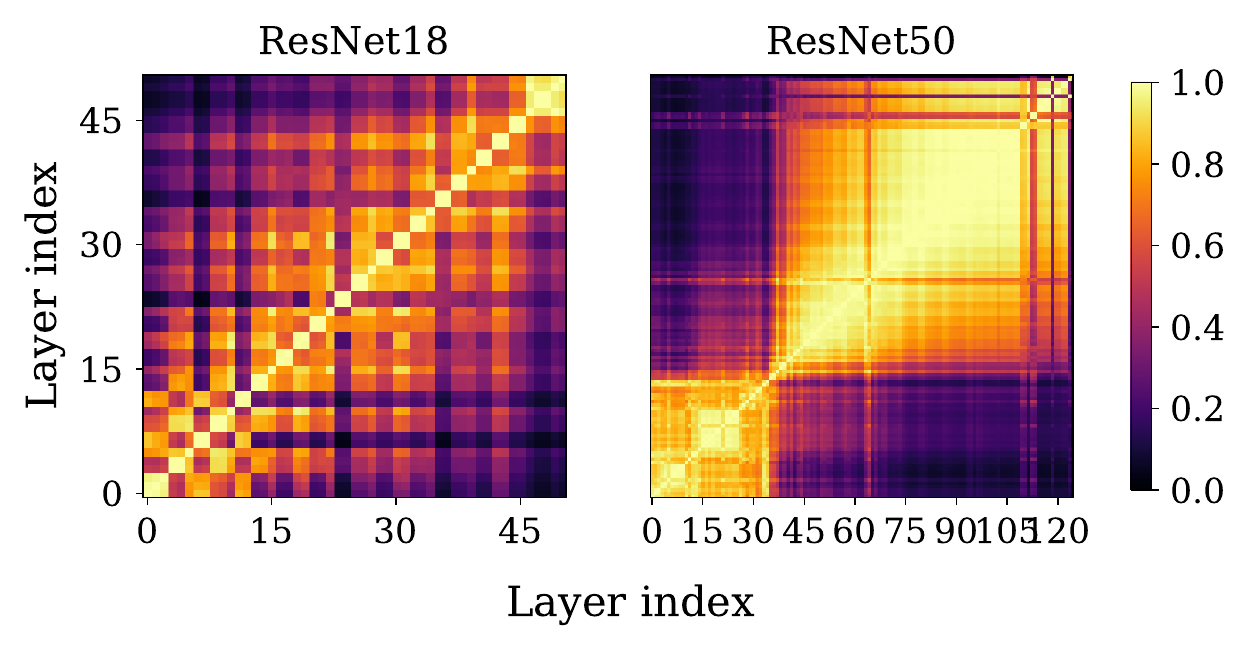}
		\caption{ImageNette (Easy)}
	\end{subfigure}
	\begin{subfigure}[b]{0.45\linewidth}
		\centering
		\includegraphics[width=\linewidth]{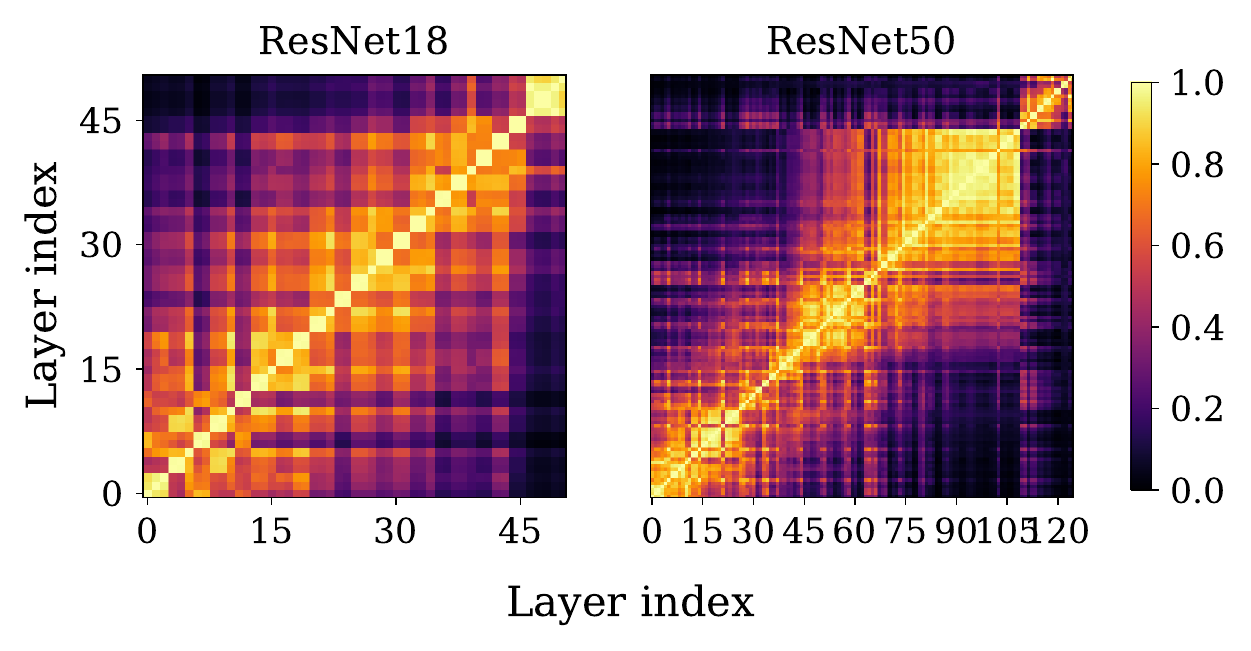}
		\caption{ImageWoof (Hard)}
	\end{subfigure}
	\caption{Cross layer cka for \textit{non-robust} network (using benign features)}
	\label{fig:non_robust_imagenet}
\end{figure}

\begin{figure}[!htb]
	\centering
	\begin{subfigure}[b]{0.45\linewidth}
		\centering
		\includegraphics[width=\linewidth]{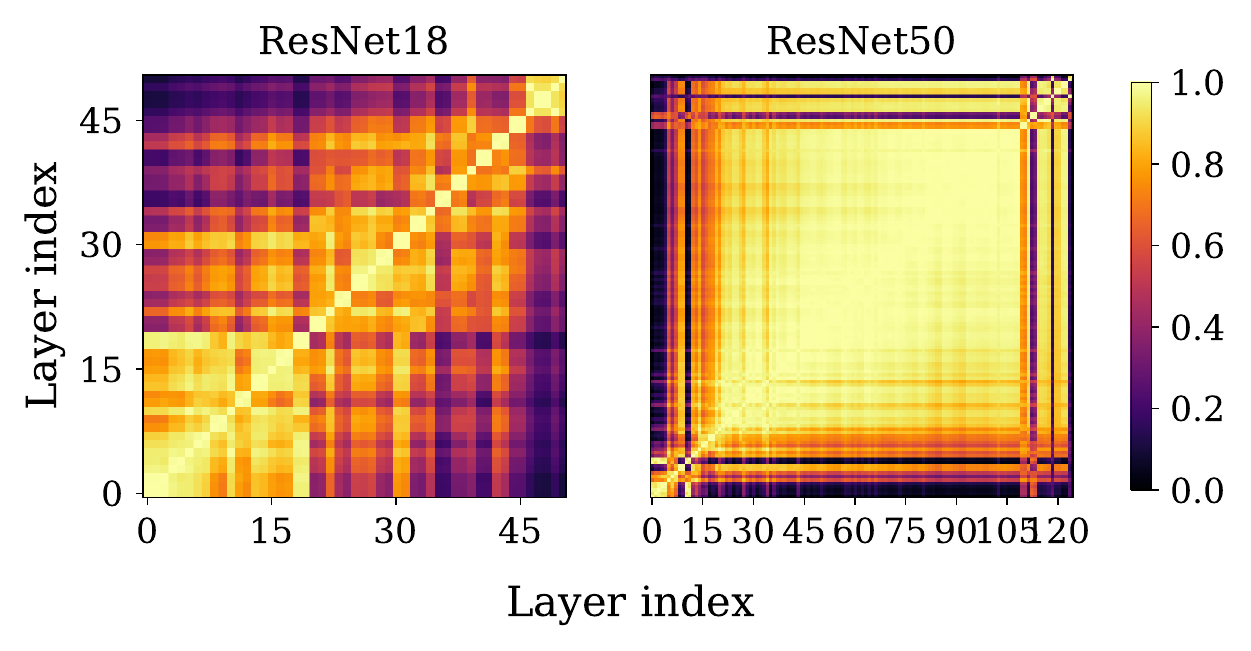}
		\caption{ImageNette (Easy)}
	\end{subfigure}
	\begin{subfigure}[b]{0.45\linewidth}
		\centering
		\includegraphics[width=\linewidth]{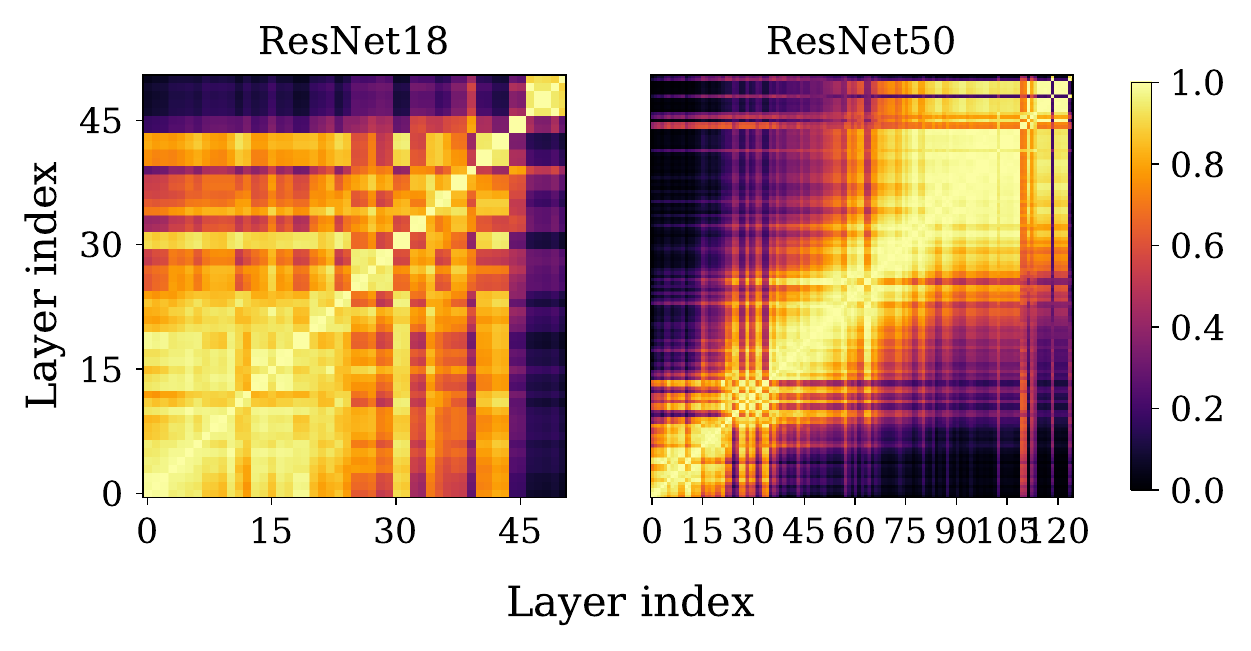}
		\caption{ImageWoof (Hard)}
	\end{subfigure}
	\caption{Cross layer cka for \textit{robust} network (using benign features)}
	\label{fig:robust_imagenet}
\end{figure}

In Figures \ref{fig:non_robust_imagenet} and \ref{fig:robust_imagenet}, we plot the layerwise similarities using CKA for ResNets trained on the ImageNette and Imagewoof datasets, both of which are subsets of the Imagenet dataset. These expand upon the results from Figure 2 of the main body, and show the stark differences between the representations obtained from non-robust and robust networks. The cross-layer similarities are far more pronounced for robust networks across datasets and architectures.

\subsection{Comparing across robust training methods} \label{appsec: training_method}
\begin{figure}[!htb]
	\centering
	\includegraphics[width=0.8\linewidth]{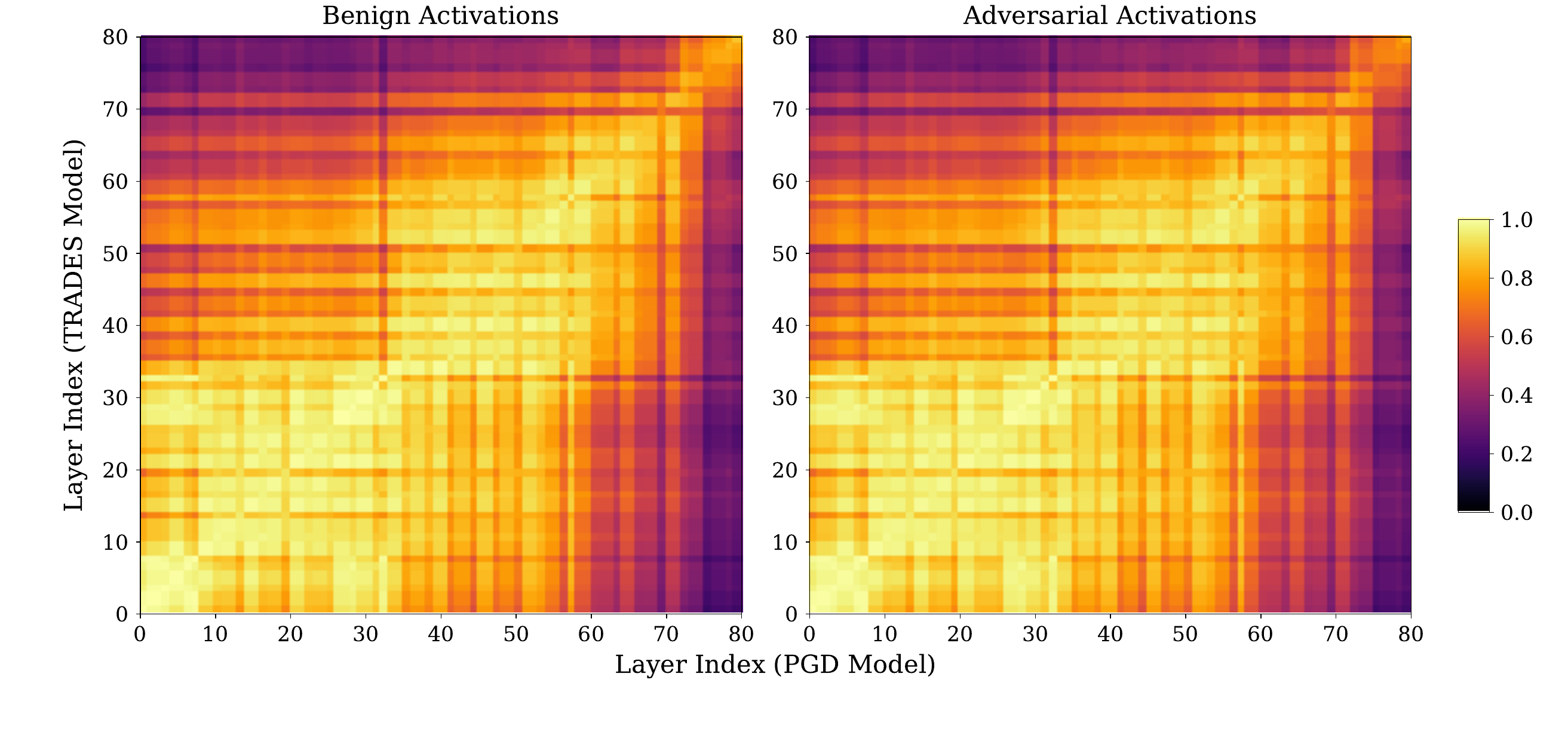}
	\caption{\textbf{Adversarial training and TRADES produce similar final representations.} Layerwise similarity between benign activations of a TRADES-trained
		\wrnte{28}{5} model and a PGD-trained \wrnte{28}{5} model.  The structure of this plot is similar to that of Figure 2a in the main paper.}
	\label{fig:trades_pgd}
\end{figure}

In Fig. \ref{fig:trades_pgd}, we compare the benign representations from models robustly trained using two different robust training techniques: PGD-based adversarial training and TRADES. Both methods lead to very similar robust representations, and exhibit the same long-range similarity across layers.

\subsection{Comparing across architectures} \label{appsec: cross_archs}


\begin{figure*}[!htb]
	\centering
	\begin{subfigure}[b]{0.7\linewidth}
		\centering
		\includegraphics[width=\linewidth]{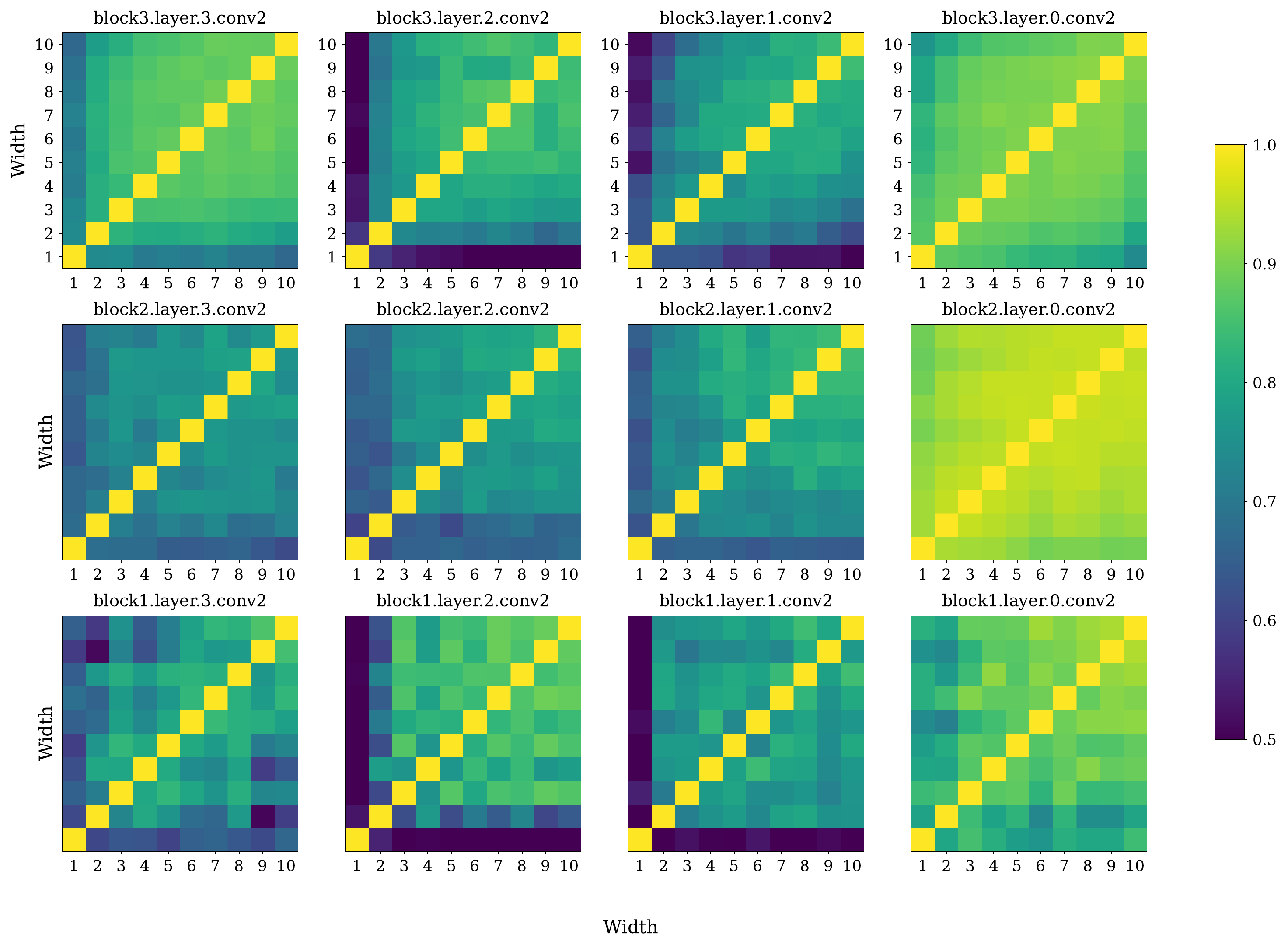}
		\caption{\textit{Non-robust} networks. Features extracted over
		\textit{benign} images.}
	\end{subfigure}
	\begin{subfigure}[b]{0.7\linewidth}
		\centering
		\includegraphics[width=\linewidth]{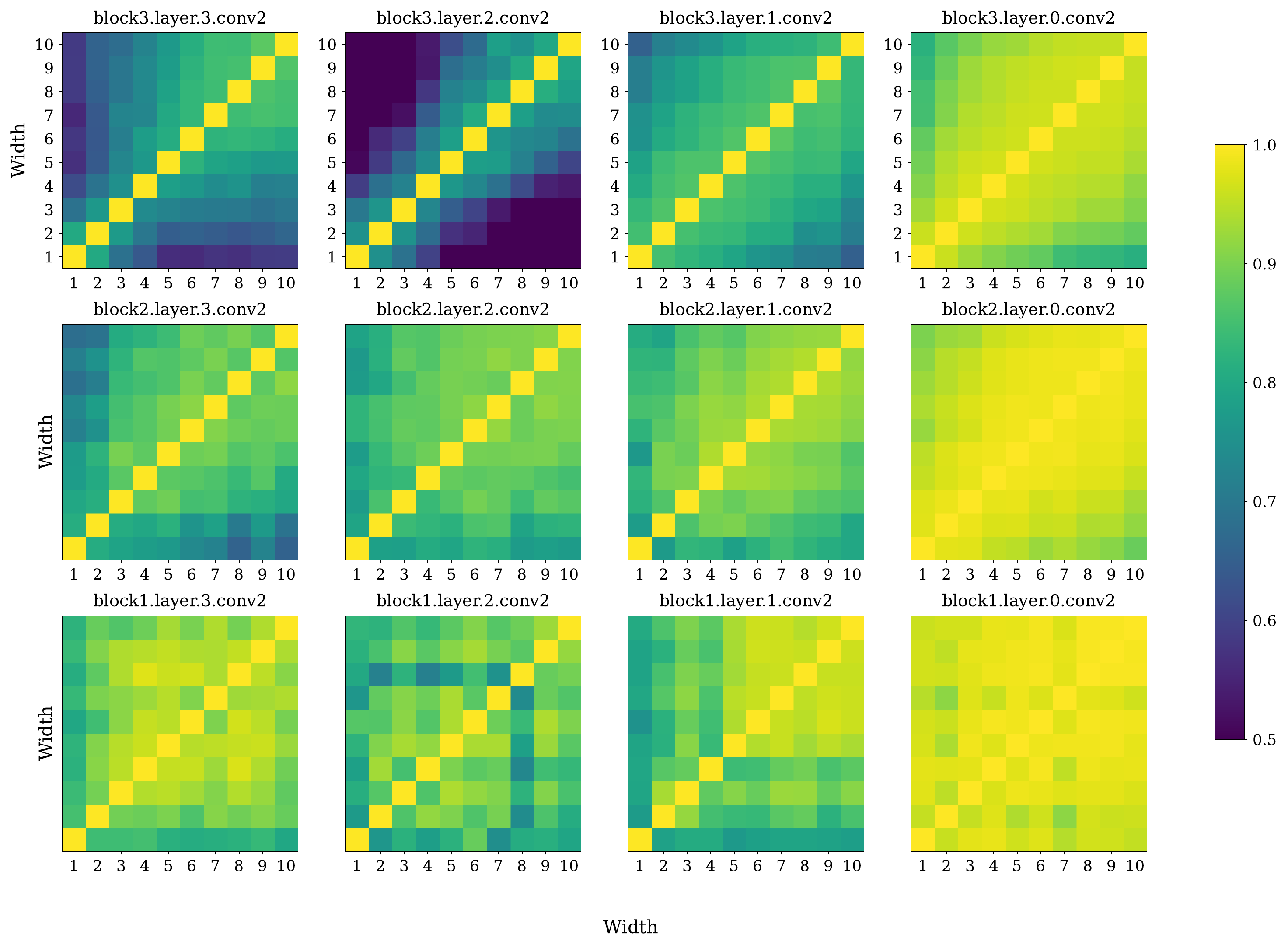}
		\caption{\textit{Robust} networks. Features extracted over
		\textit{benign} images.}
	\end{subfigure}
	\caption{\textbf{Cross-model CKA between Wide ResNet models of different widths}. Data is presented across 12 convolutional layers, starting from the last convolutional layer, i.e., block3.layer3.conv2 and moving backward in the network. Robust networks have a high degree of similarity across widths in earlier layers but large variation in deeper layers, which is the opposite trend to that observed for non-robust networks.}
	\label{fig: cross_width_cka}
	\vspace{-10pt}
\end{figure*}

 Recent work \citep{huang2021exploring}
has studied the influence of depth and width on adversarial robustness. We
extend this investigation by comparing the benign representations obtained from
Wide-ResNet models trained with different widths in Figure~\ref{fig:
	cross_width_cka} with both non-robust and robust training. A few key observations
stand out: i) early layers in different blocks are very similar across different
widths for robust training, as compared to benign training, ii) increasing width
only leads to large variations in learned representations in the last block for
robust training, and iii) benign networks of different widths are similar deeper
in the network, and different early on, with the opposite trend for robust
networks. These results, taken together suggest that increased width early on in
the network may not be particularly useful, but deeper in the network, leads to
divergent representations, which can be investigated further to synthesize
robust architectures.

\subsection{Delving deeper into CKA} \label{appsec:class_cond_cka}

\noindent \textbf{Class-aware representation similarity.} Standard RS metrics don't consider data labels, thus similarity scores doesn't
provide insights in how individual class data impacts it. As an example, we consider Linear CKA (Equation 3 from the main body), where the similarity is dependent on the dot-product of vectorized Gram  matrices ($\mK, \mL$). Two factors contribute to the similarity of gram matrices: 1) Intra-class similarity ($C_1$): $\sum_{\substack{{i, j} ,  y_i  = y_j}} K_{ij} L_{ij}$, 2) Inter-class similarity ($C_2$): $\sum_{\substack{i, j},  y_i \neq y_j} K_{ij} L_{ij}$. When measured between identical features, linear CKA is $1$, i.e., $C_1/(C_1 + C_2) + C_2 / (C_1 + C_2)$. Under a near uniform distribution of class labels, only a small fraction of entries in gram matrices contribute to intra-class similarity ($C_1$), as there are a lot more cross terms in the CKA computation. However, when experimentally measuring the influence of both terms, i.e., we find that the contribution of intra-class CKA is very high, and it increases as we go deeper into the network. This is likely because separability of feature improves deeper in the network with similar class features being clustered together, which naturally increases the magnitude of $C_1$ in the similarity score. 

We hypothesized that robust networks would have lower intra-class similarity as compared to non-robust networks, which would explain their lower classification accuracy on benign data. While this holds for the CIFAR-10 and ImageNette, for Imagewoof, robust networks have a much higher intra-class contribution. We leave a detailed investigation for future work.

\begin{figure}[!htb]
	\centering
	\includegraphics[width=0.9\linewidth]{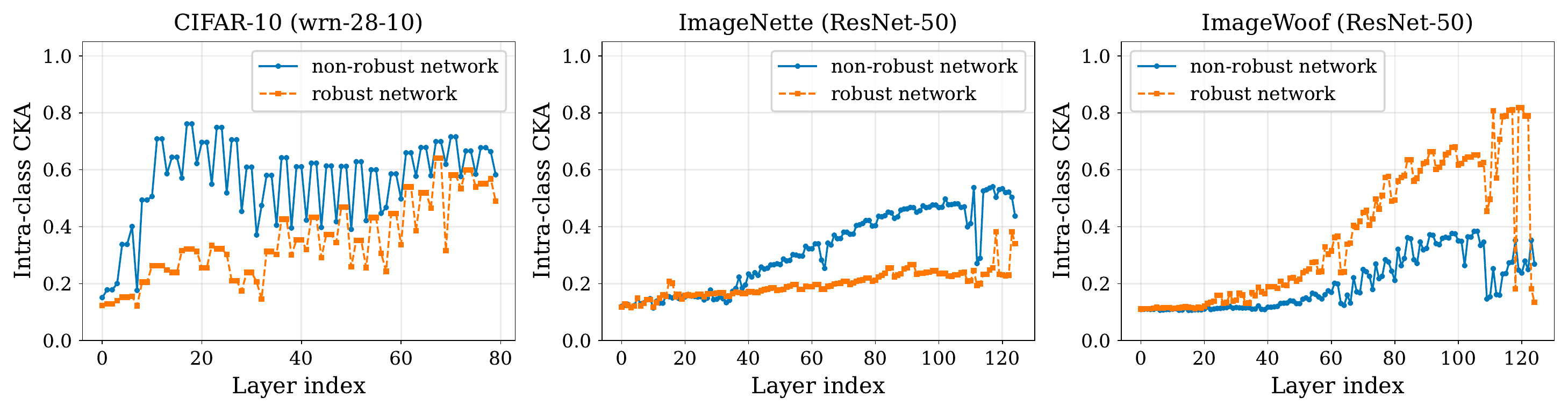}
	\caption{\textbf{Intra-class similarity.} We find that the contribution of intra-class similarity, measured by intra-class CKA, increases to non-trivial fraction, as we go deeper into the network.}
\end{figure}



\section{Adversarially Perturbed Representations}\label{appsec: adv_perturb}
In this section, we add further results to back up the insights about the impact of adversarially perturbed data on resulting representations from \S 4 of the main paper.

\subsection{Comparing benign and perturbed representations}
\begin{figure}[!htb]
	\centering
	\begin{subfigure}[b]{0.95\linewidth}
		\centering
		\includegraphics[width=\linewidth]{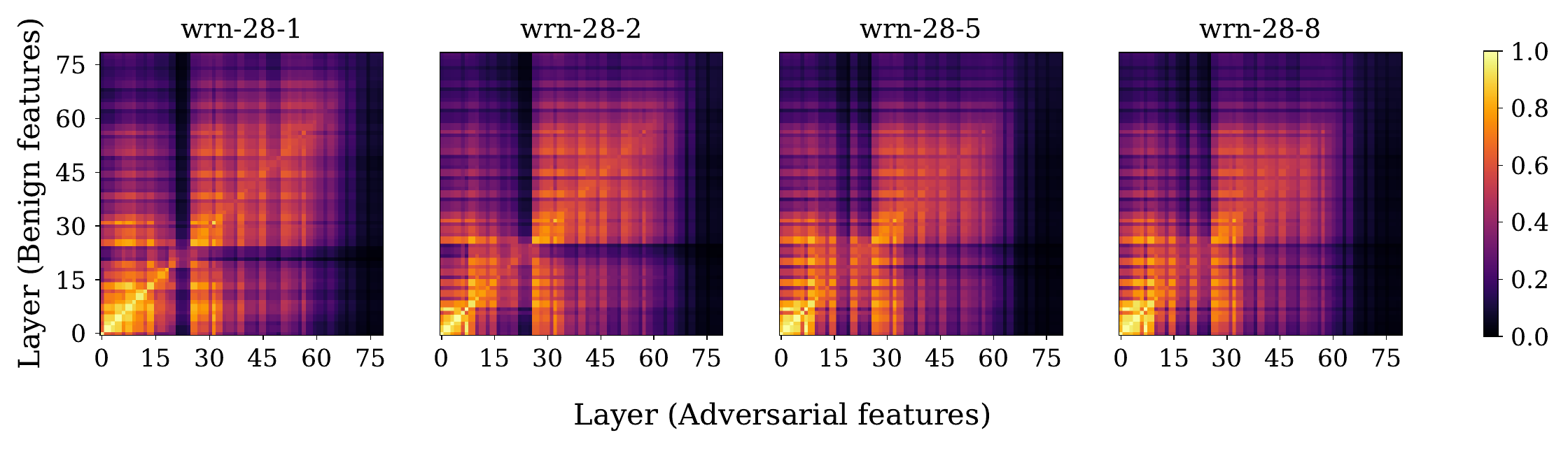}
	\end{subfigure}
	\begin{subfigure}[b]{0.95\linewidth}
		\centering
		\includegraphics[width=\linewidth]{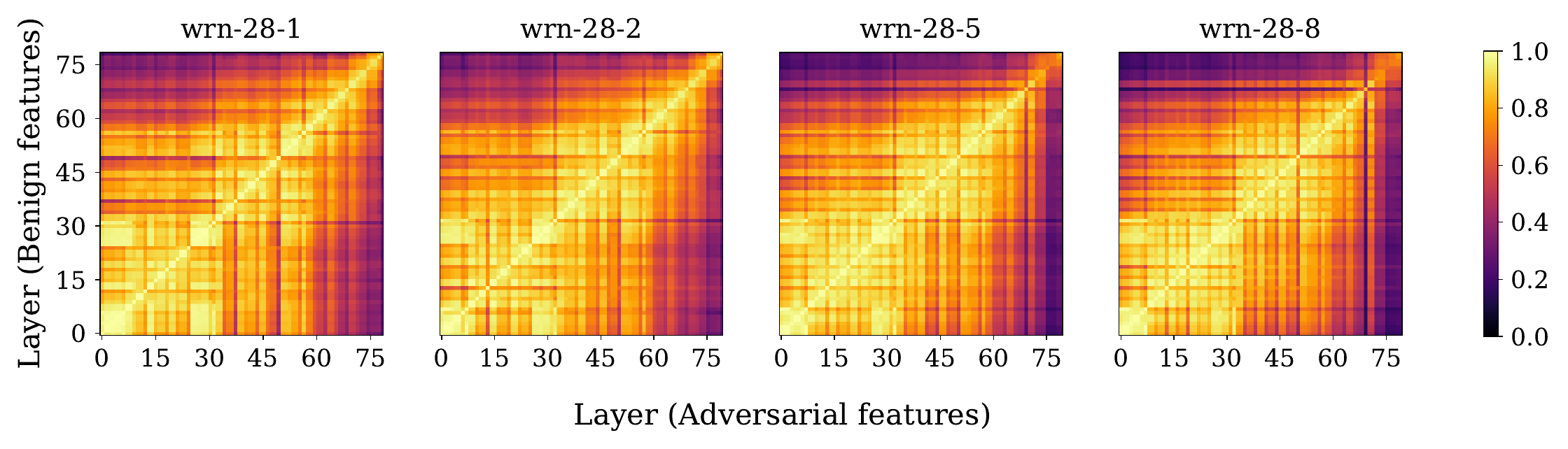}
	\end{subfigure}
	\caption{\textbf{Adversarial examples cause representation drift in later layers.} In the similarity plots of non-robust network representations (\emph{top row}), adversarial and benign features display a similar structure to plots comparing just benign representations of non-robust models only in the early layers. In later layers, adversarial representations are very dissimilar from the benign representation of all layers. The similarity plots of robust networks (\emph{bottom row}) are quite similar to plots comparing just benign representations of robust models, suggesting that robust networks learn similar representations for benign and adversarial data.}
	\label{fig: perturbed_vs_benign_all}
\end{figure}

\label{appsec:
ben_robust_comp}
In Figure \ref{fig: perturbed_vs_benign_all}, we can see, across, architectures, benign and perturbed representations differ at all layers except the first few for non-robust networks. In addition, for robustly trained networks, increasing width increases the divergence between benign and perturbed representations. This may indicate the presence of increased capacity to learn different representations for benign and adversarial inputs.

Across different threat models (Figure \ref{fig:benign_robust_compare_cross_threats}), we find robustly trained networks learn aligned benign and perturbed representations. This indicates that, in general, for most threat models, robust training is effective at ensuring that perturbed representations do not differ too significantly from benign ones, unlike for non-robust networks.

\begin{figure}[!htb]
	\centering
	\begin{subfigure}[b]{0.95\linewidth}
		\centering
		\includegraphics[width=\linewidth]{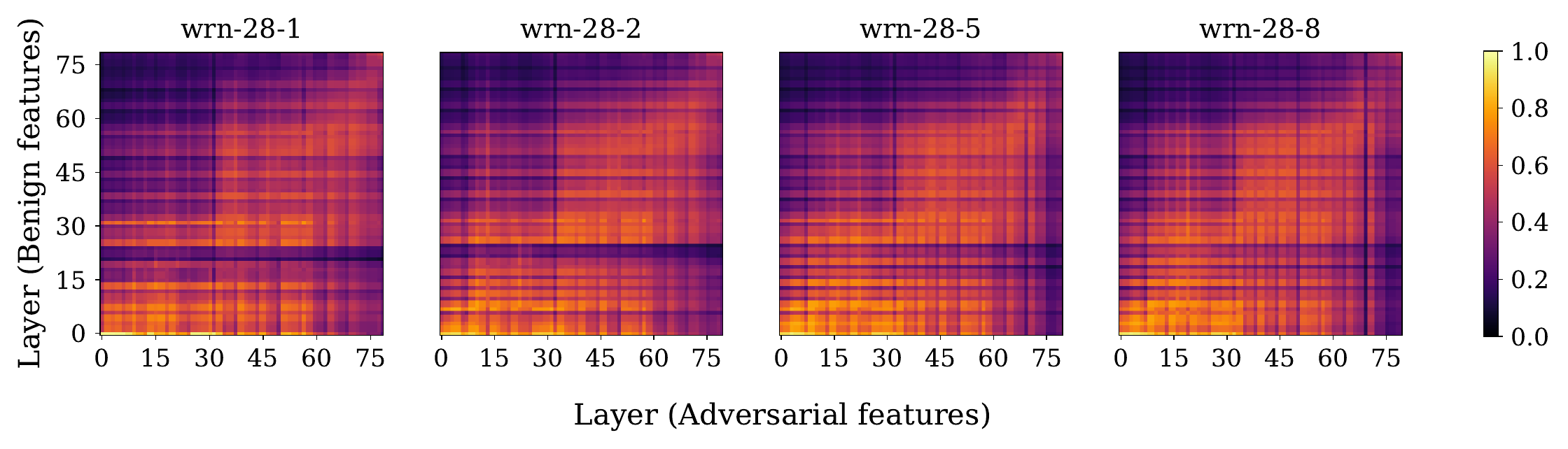}
	\end{subfigure}
	\begin{subfigure}[b]{0.95\linewidth}
		\centering
		\includegraphics[width=\linewidth]{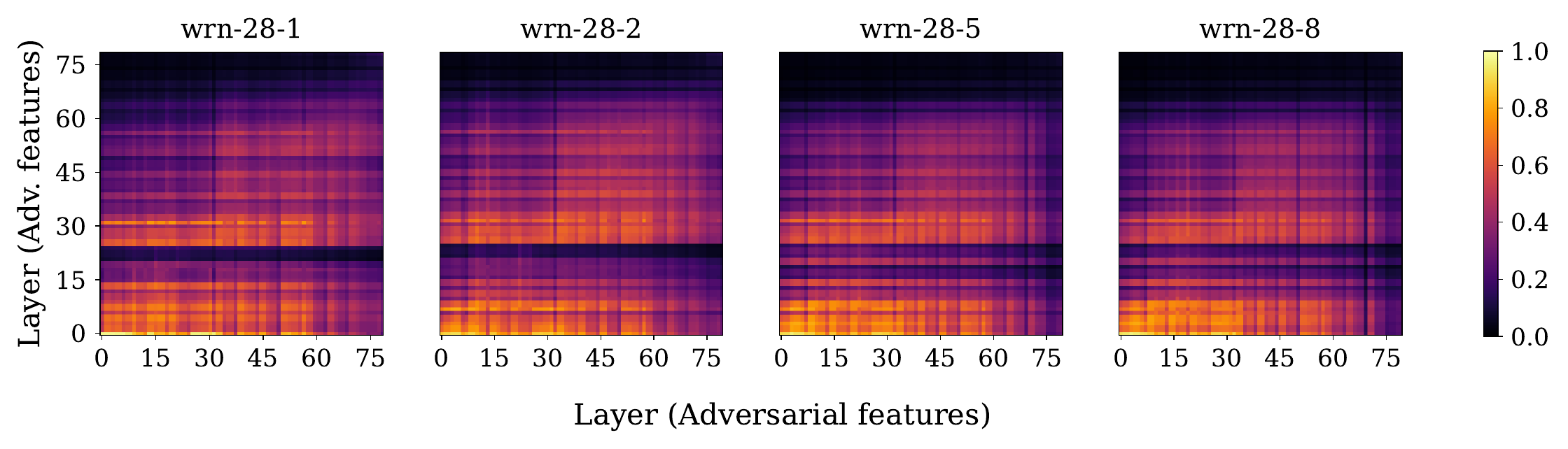}
	\end{subfigure}
	\caption{\textbf{Non-robust and robust models differ substantially across layers for both benign and perturbed inputs.} Cross layer CKA between a non-robust model (Y-axis) and robust model (X-axis) using perturbed and non-perturbed features.}
\end{figure}

\begin{figure}[!htb]
	\centering
	\includegraphics[width=\linewidth]{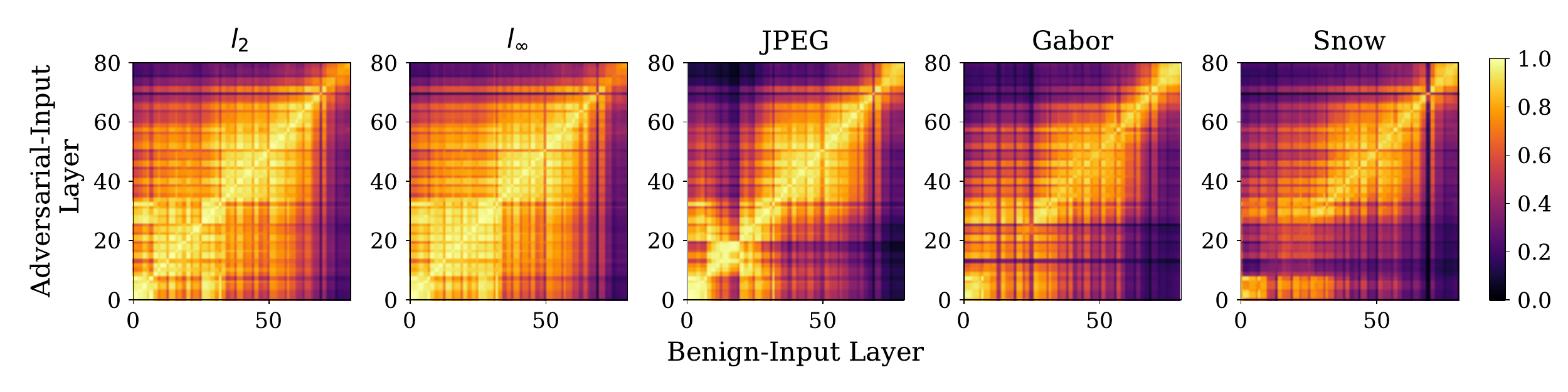}
	\caption{\textbf{Robustly trained models across threat models have aligned benign and perturbed representations.} Except the `Snow' threat model, all the others have identical benign and perturbed representations across layers.}
	\label{fig:benign_robust_compare_cross_threats}
\end{figure}

\begin{figure}[!htb]
	\centering
	\begin{subfigure}[b]{0.95\linewidth}
		\centering
		\includegraphics[width=\linewidth]{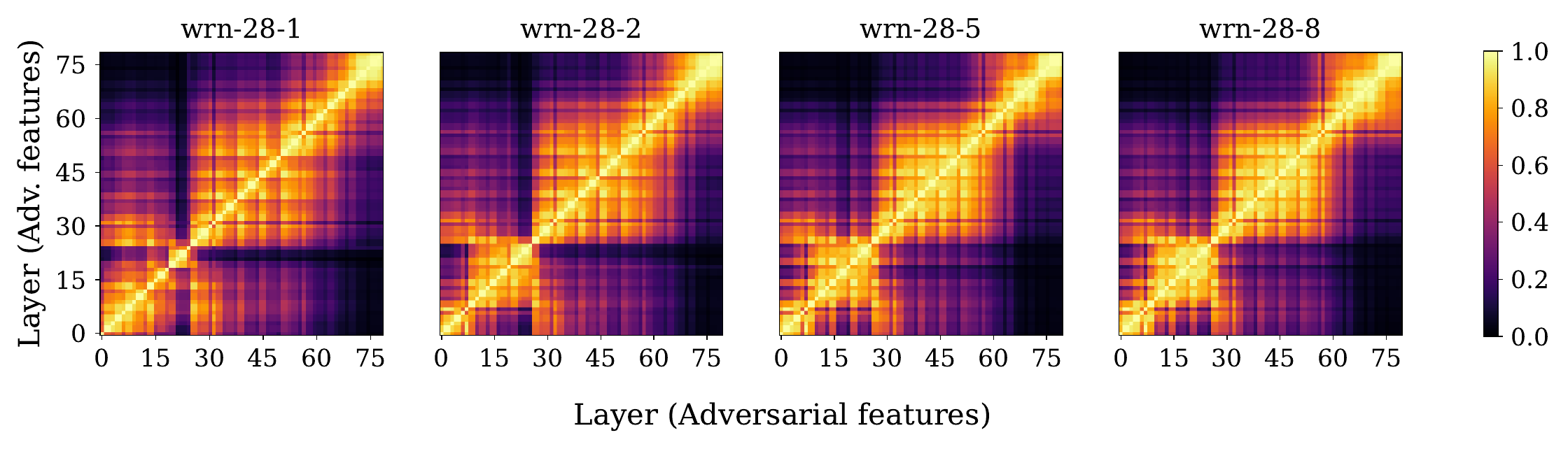}
	\end{subfigure}
	\begin{subfigure}[b]{0.95\linewidth}
		\centering
		\includegraphics[width=\linewidth]{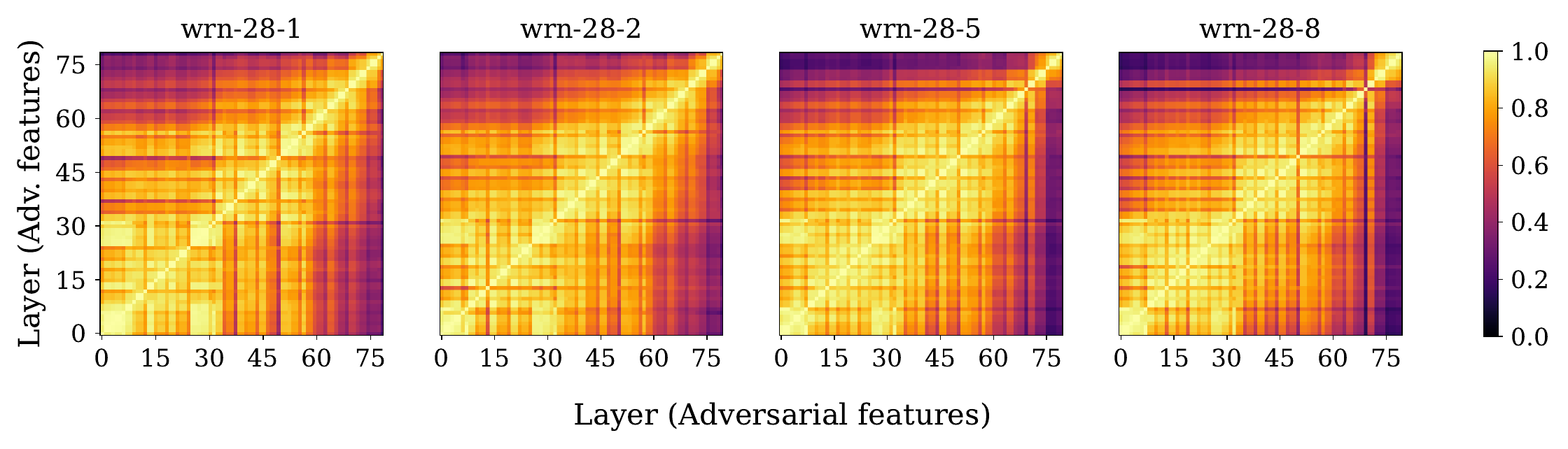}
	\end{subfigure}
	\caption{\textbf{Adversarial examples have different impacts on the representations of benign and robust models.} In benign models (top row), the layerwise similarity between adversarial activations displays a similar structure to that of benign activations, except the layers with dissimilar benign activations appear to have even more dissimilar robust activations. This implies significant changes in adversarial activations are taking place in between the self similar layer "blocks". This pattern is largely absent from robust models (bottom row), which display similar patterns for both benign and adversarial inputs.}
	\label{fig: perturbed_vs_perturbed_all}
\end{figure}

\begin{figure}[!htb]
	\centering
	\begin{subfigure}[b]{0.95\linewidth}
		\centering
		\includegraphics[width=\linewidth]{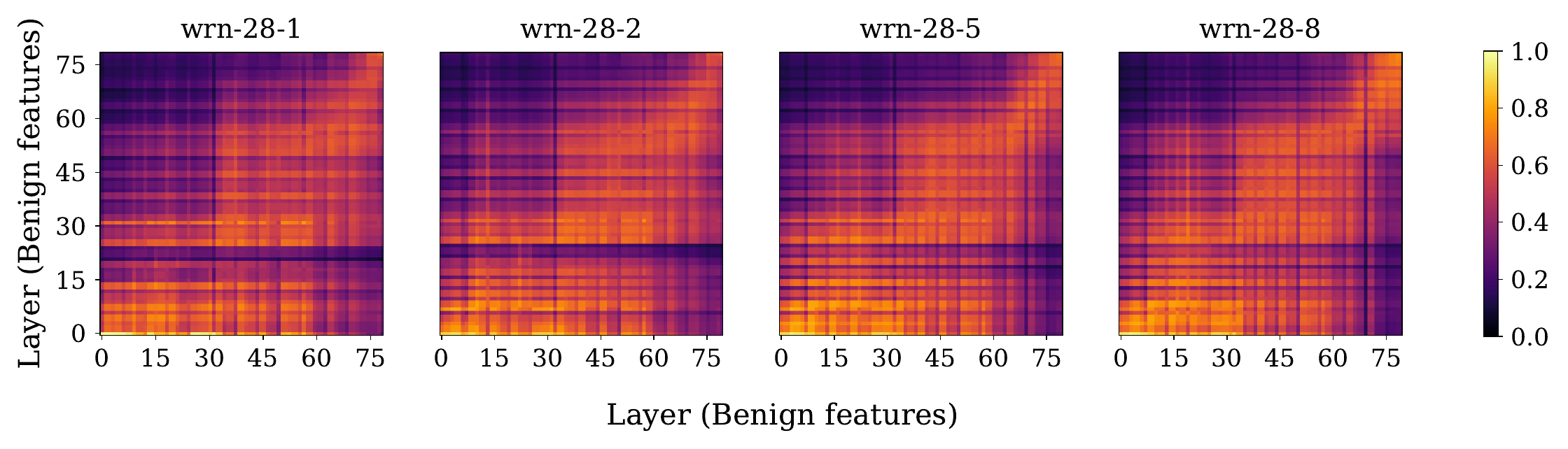}
	\end{subfigure}
	\begin{subfigure}[b]{0.95\linewidth}
		\centering
		\includegraphics[width=\linewidth]{images/consolidated_vikash/crosslayer_benign_vs_adv_net_adv_feat.pdf}
	\end{subfigure}
	\caption{ \textbf{Perturbed representations for robust and non-robust networks differ across all layers.} Cross layer CKA between a non-robust model (y-axis) and robust model (x-axis) using perturbed features.}
	\label{fig: perturbed_non_robust_vs_robust}
\end{figure}

\subsection{Adversarially perturbed
representations of non-robust networks}  

When analyzing the architectural impacts on learned representations using benign data, \citet{nguyen2020wide} note
the emergence of a block structure with increased width and depth. We find that a much stronger block structure emerges even for low-width
networks when using adversarial inputs (Fig. \ref{fig: perturbed_vs_perturbed_all}). This implies that perturbations force
representations within a residual block to have higher similarity than outside
the block. The block structure has been linked to the capacity of models, with
\citet{nguyen2020wide} noting that the `block structure arises in models that are
heavily overparametrized relative to the training dataset'. In this light,
adversarial examples have a peculiar impact on the representations of benign
models, since the block structure is more clearly visible for adversarial
inputs, but for models which do not classify the inputs correctly. This
indicates that local similarity is being exhibited and is a different phenomenon
compared to block structure emergence for well-trained models.

Figure \ref{fig: perturbed_non_robust_vs_robust} shows the stark difference between perturbed features for robust and non-robust networks. This is similar to the lack of similarity for benign inputs to these networks.

\section{Evolution over Training} \label{appsec: evolve}

In this section, we study the similarity between the activations produced by models after 
each epoch of training. We present our results in two formats, time series and heatmaps. For ease of
understanding, the time series corresponds to a single row of the heatmap, where
the representation being compared to is held constant.

\subsection{Impact of Model Capacity}
In Figure \ref{fig: cross-epoch-cka}, we plot the pairwise similarity between the representations of a single layer after each epoch of training. Natural training behaves as we would expect it to: a layer’s representations at a given epoch are most similar to those from nearby epochs, and similarity degrades as the distance between epochs increases. Higher average similarity between later epochs implies convergence to a final learned representation. This pattern is also observed in the robust training of low-capacity networks, but as capacity increases starkly different dynamics emerge. In high-capacity robust networks, we observe similarity start to decrease after about 30 epochs of training, only to increase again later on. This pattern is observed for both benign and adversarial activations.

\subsection{Adversarial Training and Overfitting}
\begin{figure*}[!htb]
	\centering
	\begin{subfigure}[b]{0.95\linewidth}
		\centering
		\includegraphics[width=\linewidth]{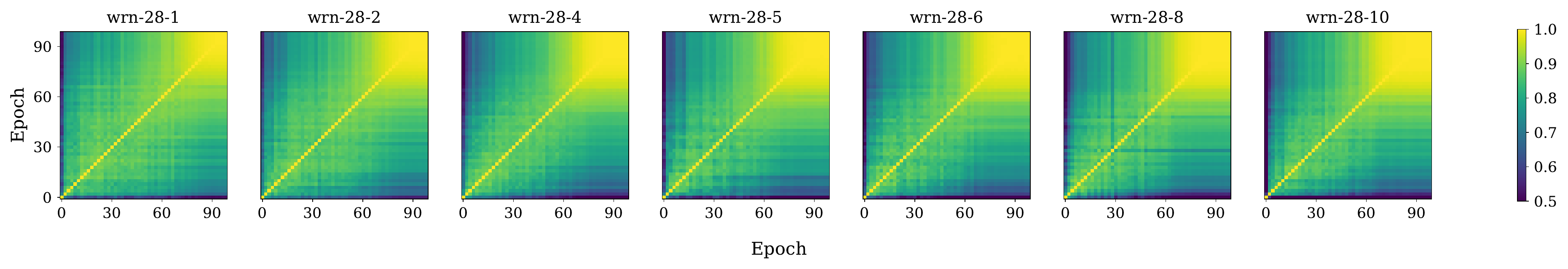}
		\caption{\textit{Non-robust} networks. Features extracted over
			\textit{benign} images.}
	\end{subfigure}
	\begin{subfigure}[b]{0.95\linewidth}
		\centering
		\includegraphics[width=\linewidth]{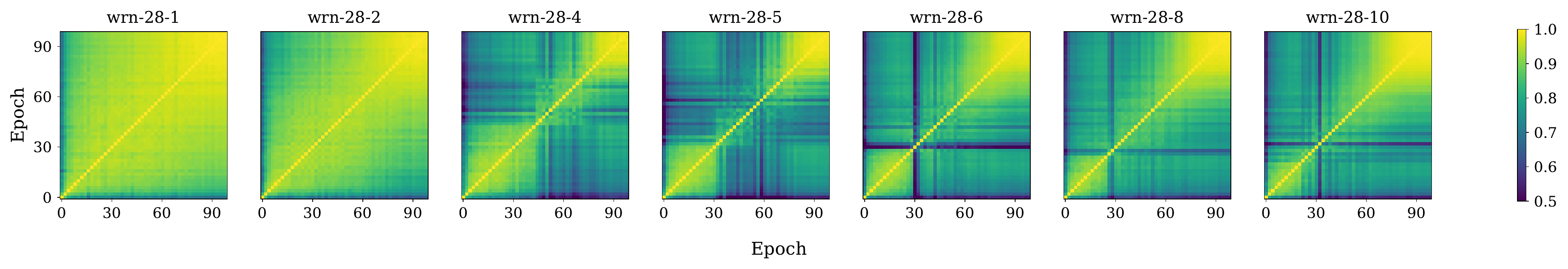}
		\caption{\textit{Robust} networks. Features extracted over
			\textit{benign} images.}
	\end{subfigure}
	\begin{subfigure}[b]{0.95\linewidth}
		\centering
		\includegraphics[width=\linewidth]{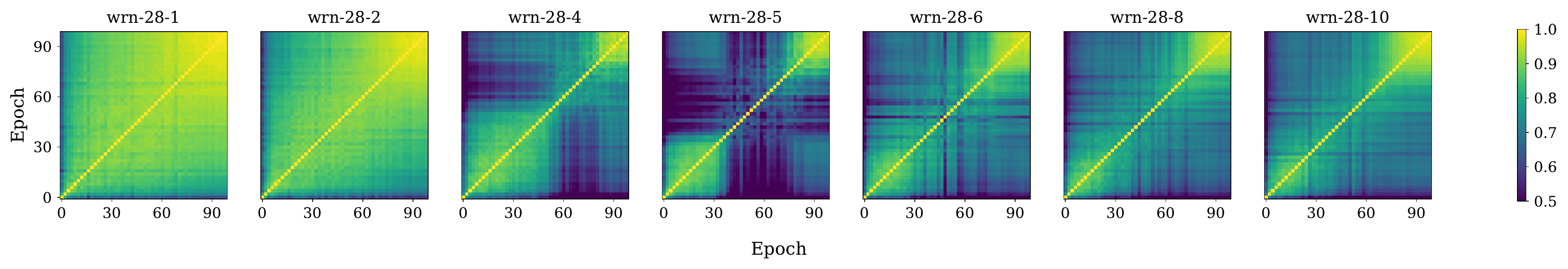}
		\caption{\textit{Robust} networks. Features extracted over
			\textit{adversarial} images.}
	\end{subfigure}
	\caption{\textbf{Stability of robust training is impacted greatly by model capacity.} Each plot shows the CKA similarity between the activations of the final layer of a Wide ResNet feature extractor at each epoch of its 100-epoch training. While the stability of non-robust training is largely unaffected by capacity, the robust training of models above a certain capacity causes representations to be learned in the middle of training that are starkly different from the initial and final representations. This effect is even more pronounced when comparing adversarial representations. }
	\label{fig: cross-epoch-cka}
\end{figure*}

In Figure \ref{fig: training_evolution}, we plot the similarity to the final learned representations of the representations extracted from each of the three Wide ResNet blocks (block1, block2, block3) and the final average pooling layer before classification (avgpool) after each epoch of training. We also plot the validation loss of each model for comparison. We find that for models trained at $\epsilon = \frac{8}{255}$, instability in the later layers is not observed until the model begins to overfit on adversarial examples. At high perturbation strengths, training instability is observed at earlier layers and earlier in training. Training of the lowest capacity model is stable at all perturbation strengths, a trend also observed in Figure \ref{fig: cross-epoch-cka}.

\subsection{Different Training Methods}	
\begin{figure*}
	\centering
	\begin{subfigure}{0.8\linewidth}
		\centering
		\includegraphics[width=\linewidth]{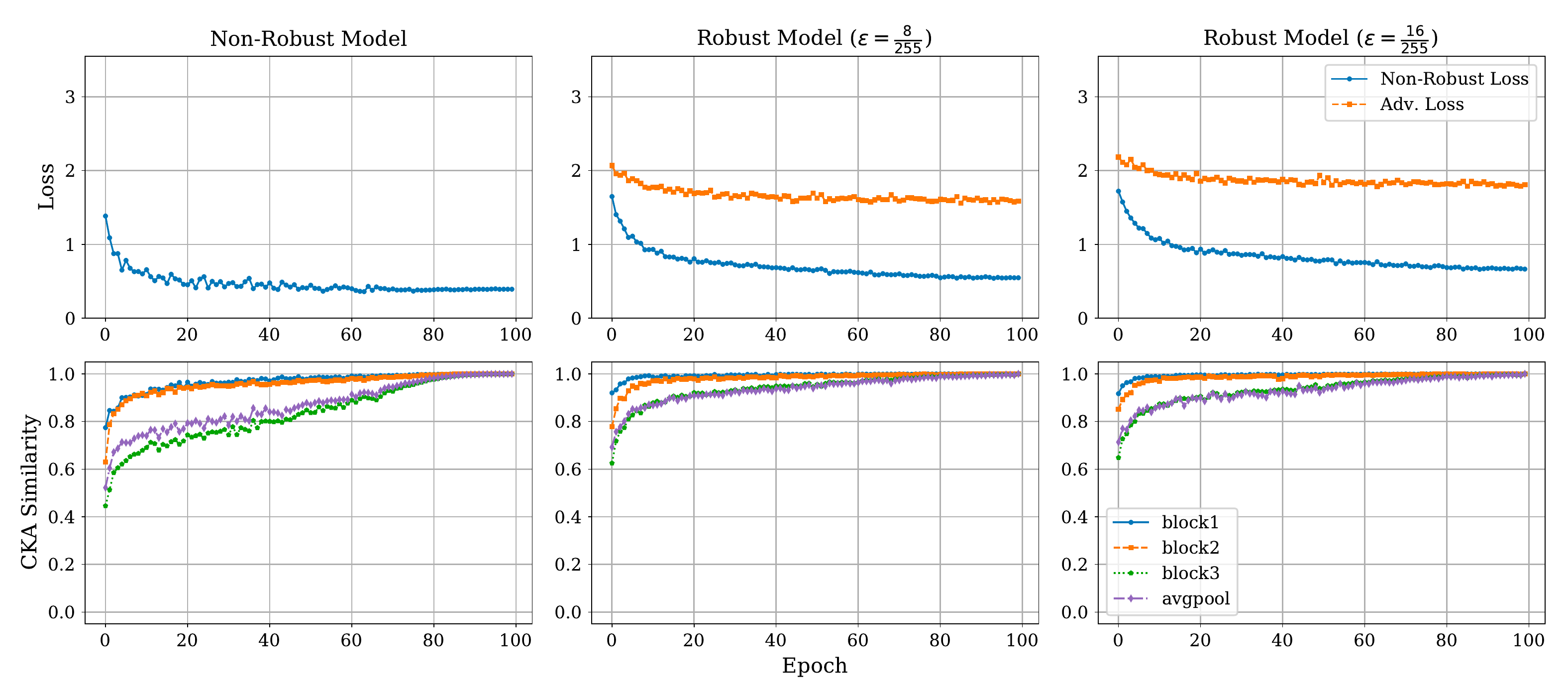}
		\caption{\wrnte{28}{1}}
		\label{subfig: wrn-28-1_evolve_linf}
	\end{subfigure}
	\begin{subfigure}{0.8\linewidth}
		\centering
		\includegraphics[width=\linewidth]{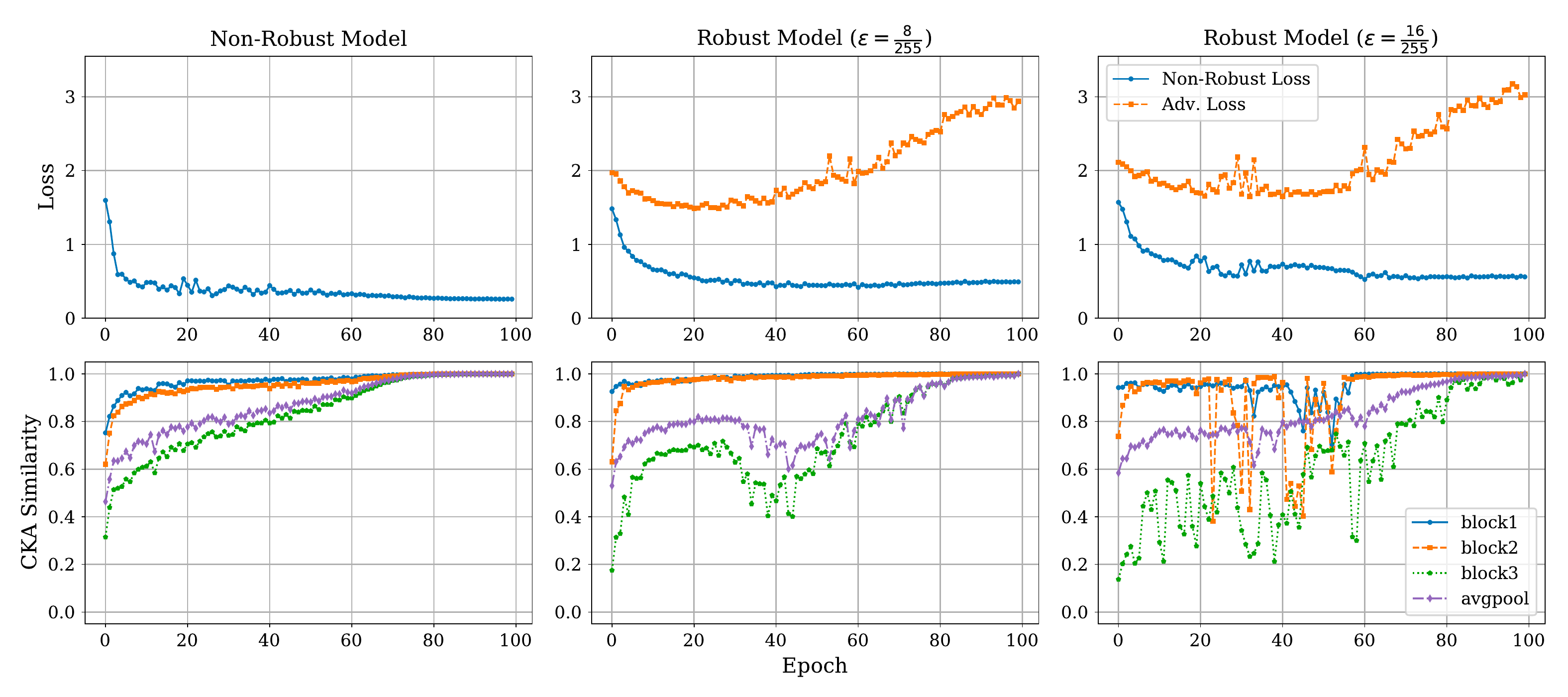}
		\caption{\wrnte{28}{5}}
		\label{subfig: wrn-28-5_evolve_linf}
	\end{subfigure}
	\begin{subfigure}{0.8\linewidth}
		\centering
		\includegraphics[width=\linewidth]{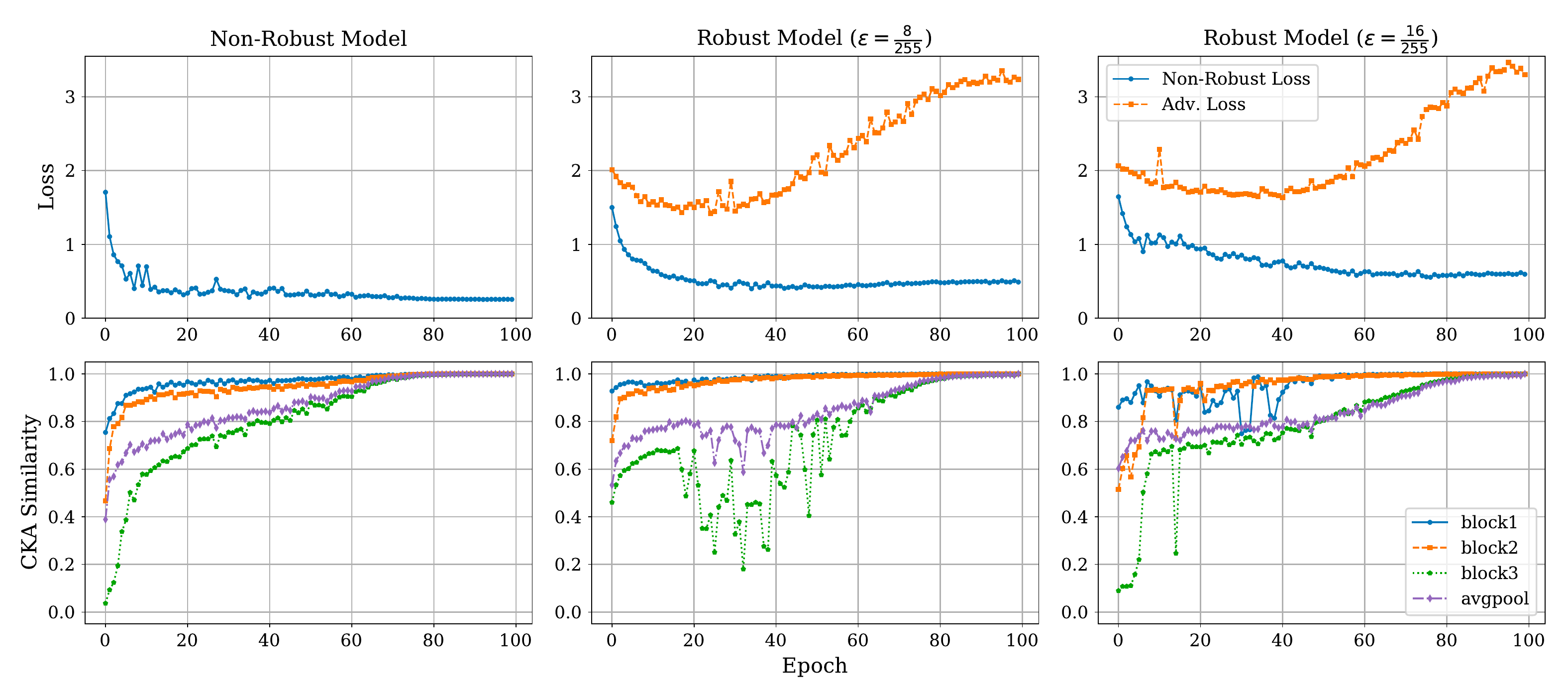}
		\caption{\wrnte{28}{10}}
		\label{subfig: wrn-28-10_evolve_linf}
	\end{subfigure}
	\caption{\textbf{Overfitting and training instability in robust learning.} Each plot shows how benign representations of selected layers of Wide ResNet models compare to their final learned representations after each epoch of training. In larger robust models trained at $\epsilon = \frac{8}{255}$, training instability begins after the point of overfitting on adversarial data. However, when the perturbation strength is increased, there are large changes in the similarities of representations of subsequent epochs starting very early in training. High perturbation strengths also induce instability at earlier layers, which is not observed at lower strengths or in non-robust training.}
	\label{fig: training_evolution}
\end{figure*}

We also sought to test whether more principled robust training methods, such as those proposed by \citet{zhang2019theoretically}, would affect the patterns in similarity we observed in this section. In Figure 6, we compare benign training, PGD training, and TRADES training in a similar manner to Figure \ref{fig: training_evolution}. The patterns we observed with PGD training disappeared when training with TRADES, highlighting the sensitivity of CKA to meaningful changes in training dynamics.

\section{Threat models}\label{appsec: threat_models}
In this section, we further investigate the impact of threat model on learned representations (see \S7 in the main body). 

\subsection{Impact of budget within a threat model}
\begin{figure*}[!htb]
	\centering
	\includegraphics[width=0.8\linewidth]{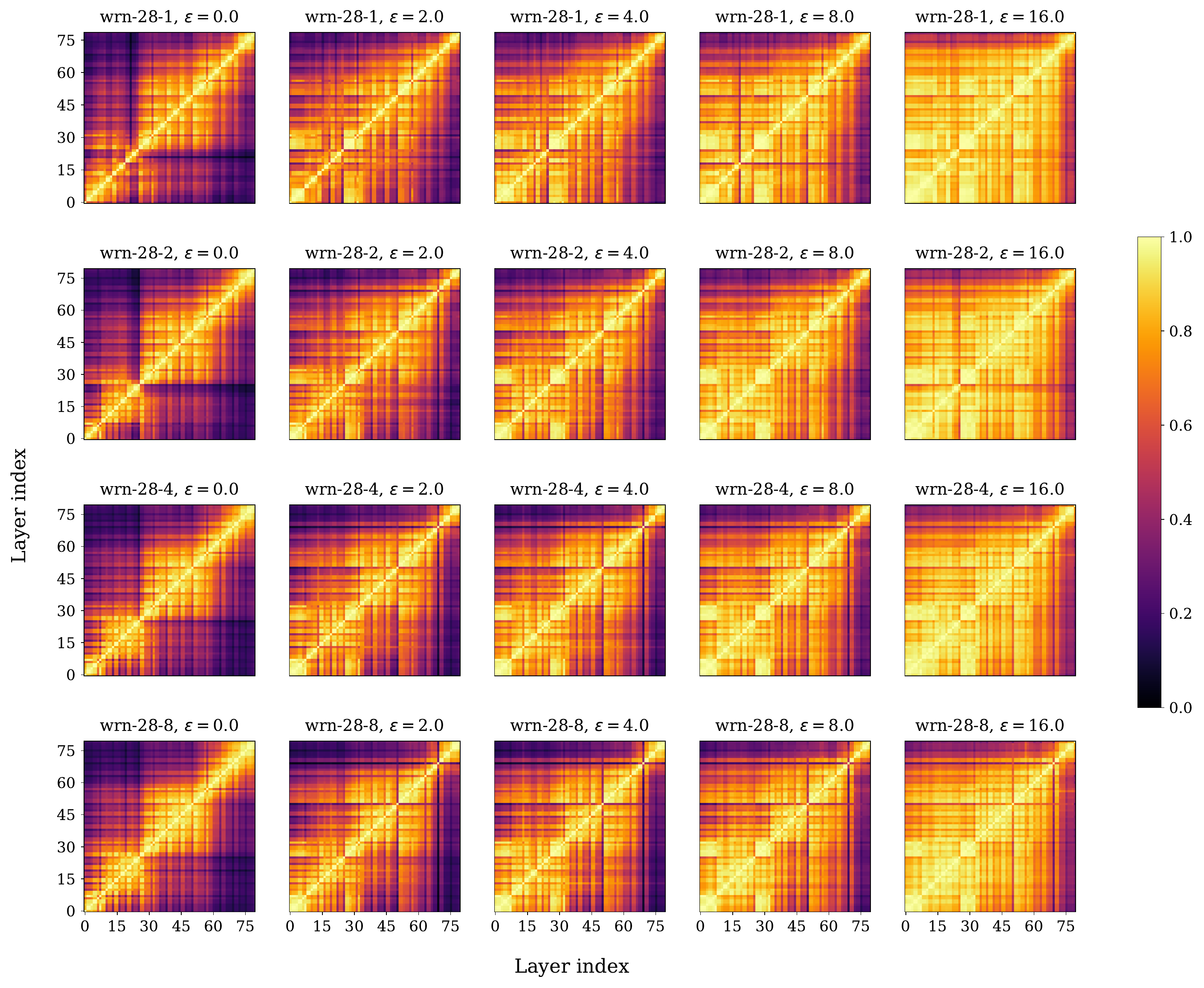}
	\caption{\textbf{Delving deeper into cross-layer similarity.} This expands upon Figure 3 from the main body of the paper, again showing that increasing the perturbation budget leads to an increase in cross-layer similarity, as does reducing the width. Both factors thus affect the relative capacity of the model.}
	\label{fig: eps_width_vary_cifar10_full}
\end{figure*}

\begin{figure*}
	\begin{subfigure}{.5\textwidth}
	  \centering
	  \includegraphics[width=\linewidth]{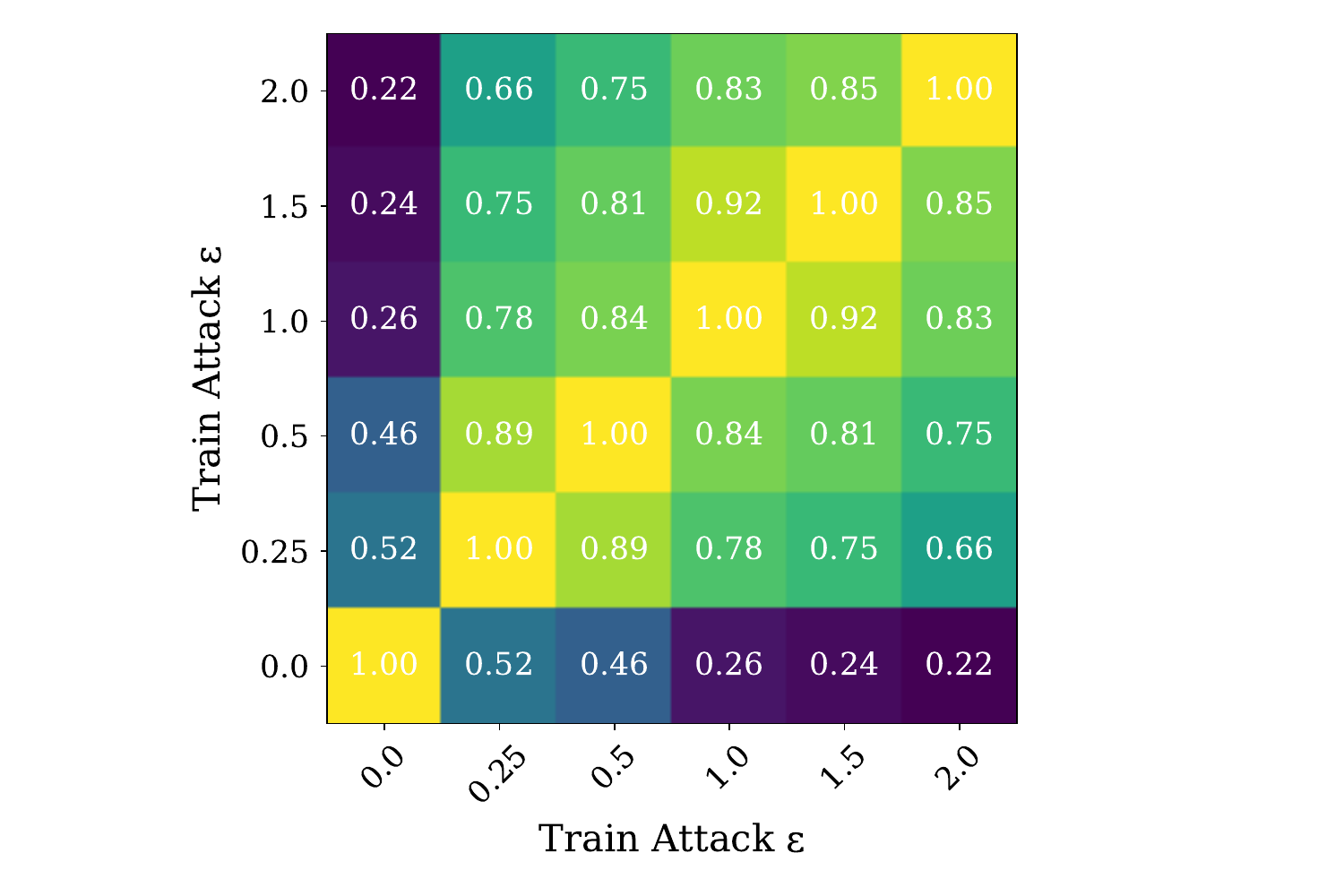}
	  \caption{$\ell_2$}
	  \label{fig:l2}
	\end{subfigure}%
	\begin{subfigure}{.5\textwidth}
	  \centering
	  \includegraphics[width=\linewidth]{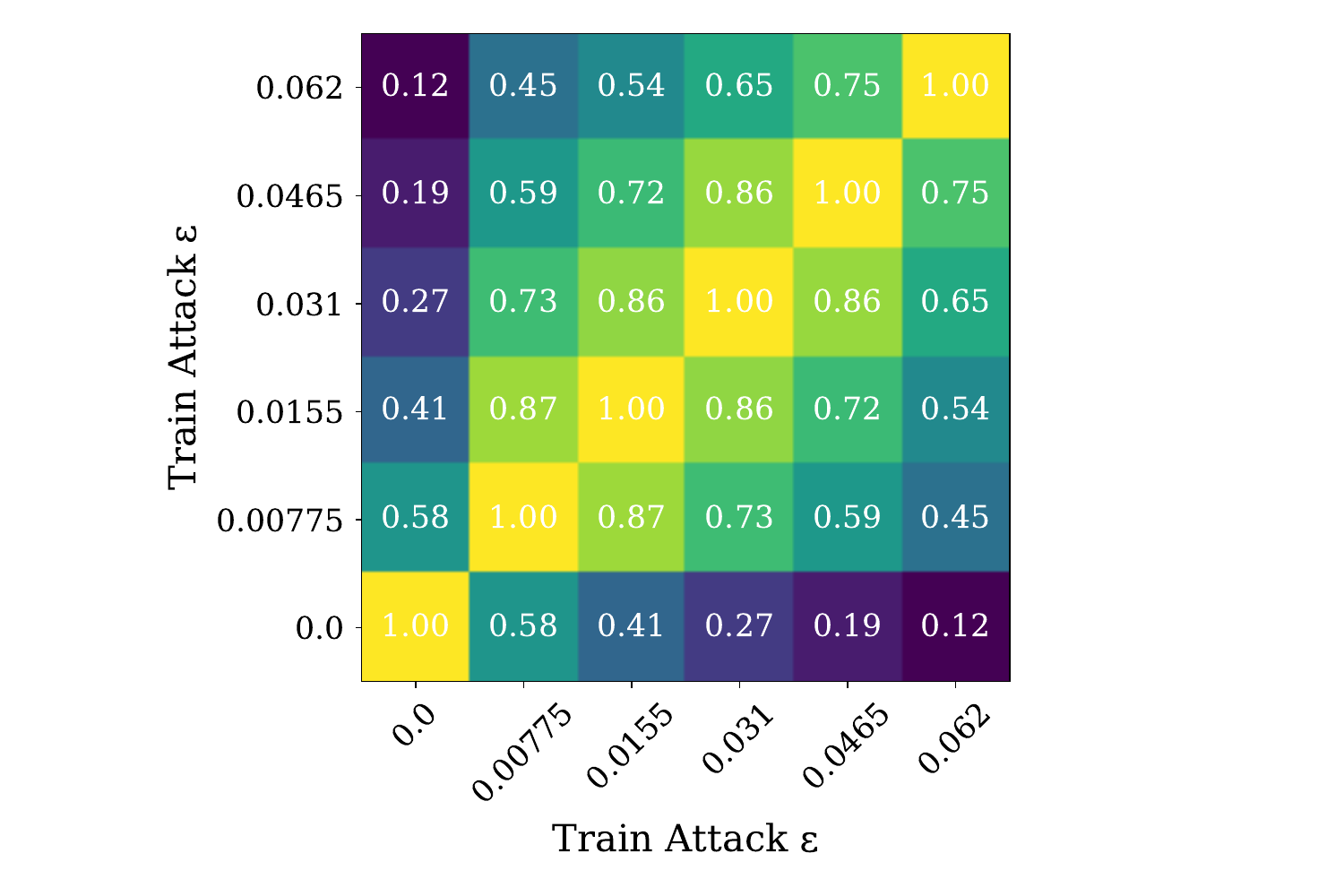}
	  \caption{$\ell_{\infty}$}
	  \label{fig:linf}
	\end{subfigure}
	\begin{subfigure}{.5\textwidth}
	  \centering
	  \includegraphics[width=\linewidth]{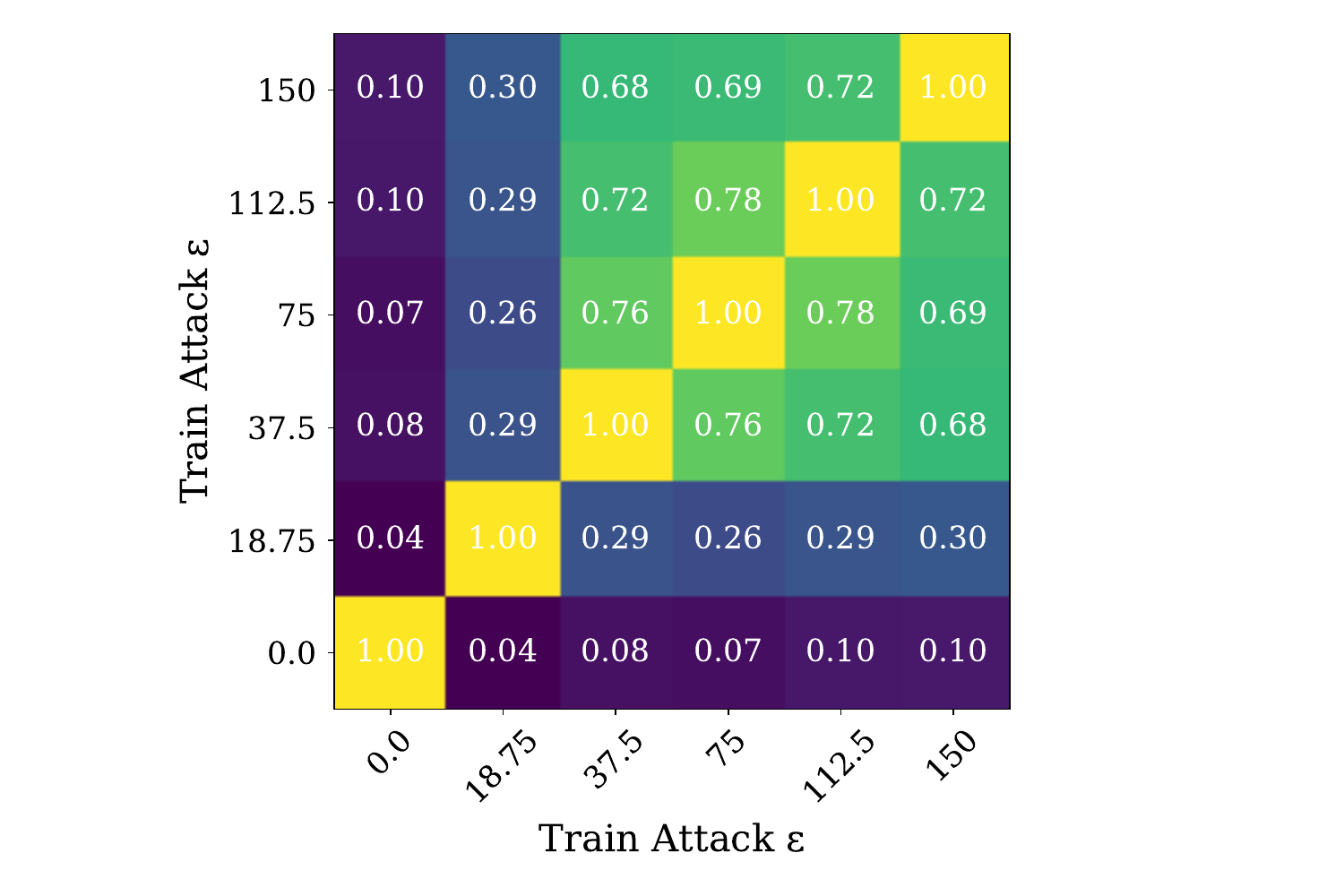}
	  \caption{Gabor}
	  \label{fig:gabor}
	\end{subfigure}
	\begin{subfigure}{.5\textwidth}
	  \centering
	  \includegraphics[width=\linewidth]{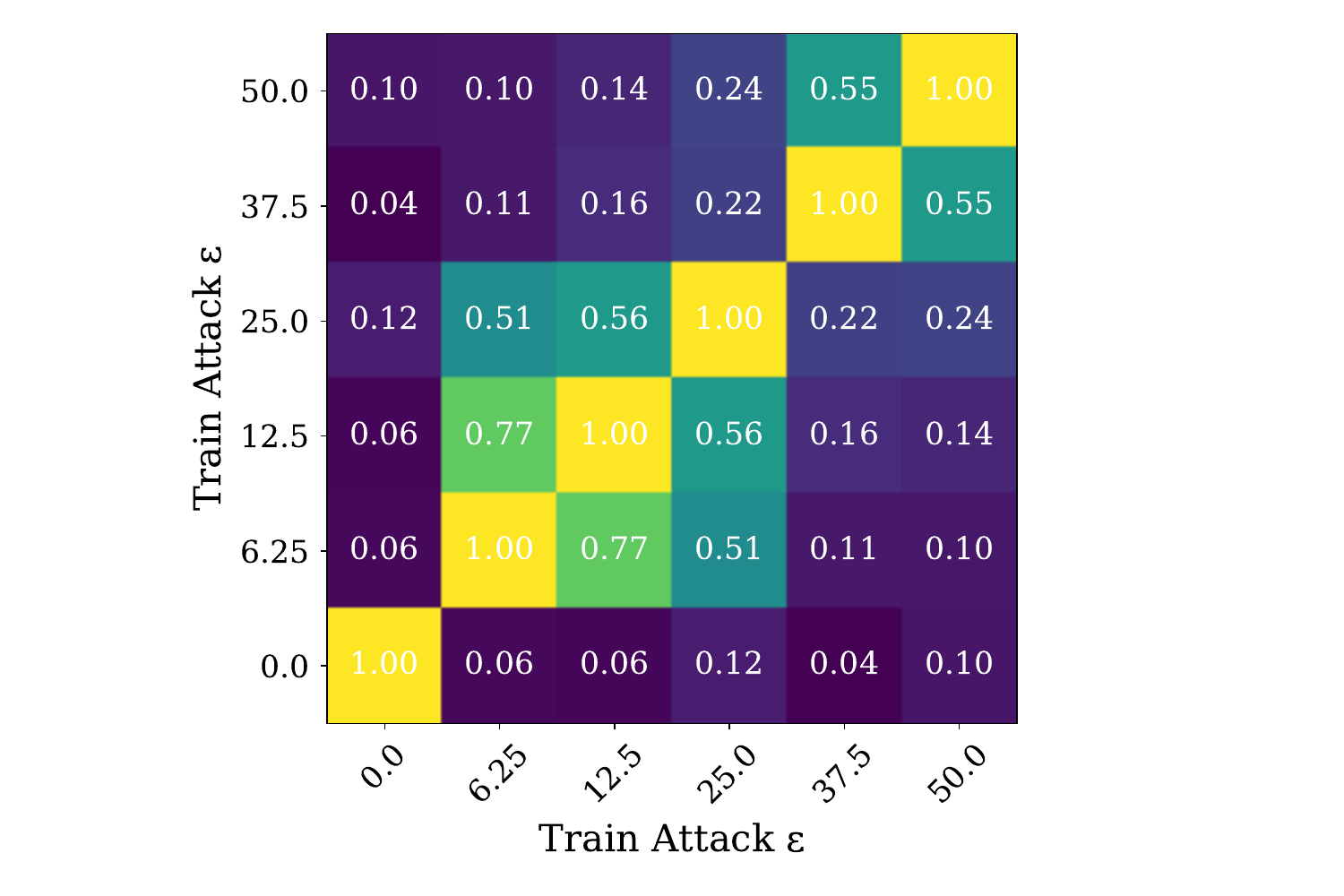}
	  \caption{JPEG}
	  \label{fig:jpeg}
	\end{subfigure}
	\begin{subfigure}{\textwidth}
	  \centering
	  \includegraphics[width=.5\linewidth]{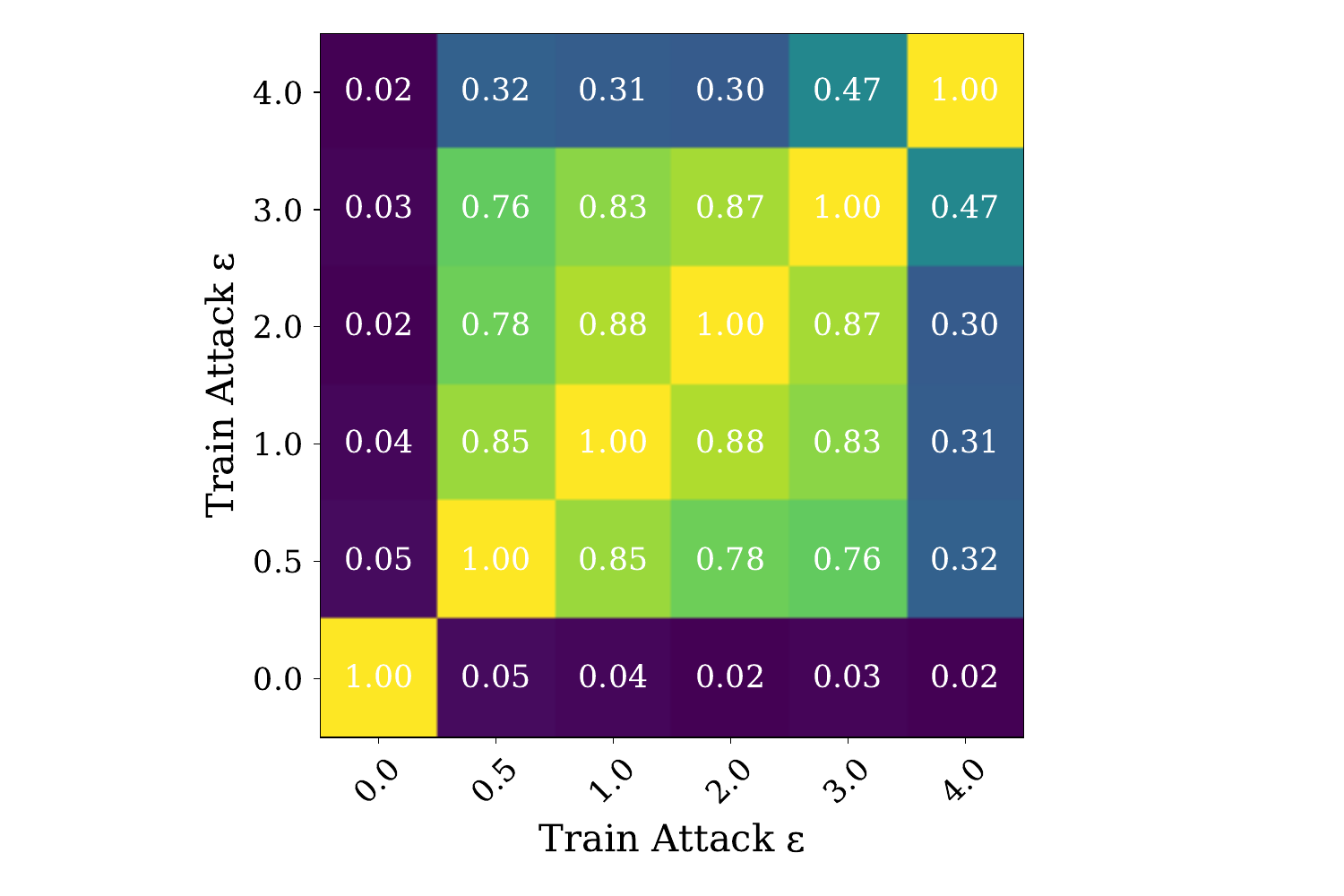}
	  \caption{Snow}
	  \label{fig:snow}
	\end{subfigure}
	\caption{\textbf{Adversarial training induces very different representations, even at low attack strengths.} Each plot displays the CKA similarity between activations of the final layer of \wrnte{28}{10}~feature extractors adversarially trained on different strengths of each threat model. In each plot, even the representations of the lowest strength robust model are quite dissimilar from those of the benign model, with only the $\ell_p$ bounded models having that similarity be relatively close to the lowest similarities between robust models.}
	\label{fig: within_attack_cross_strength}
	\end{figure*}

In Figure \ref{fig: eps_width_vary_cifar10_full}, we expand upon Figure 3 from the main body of the paper. We show that increasing the budget and reducing width both lead to a decrease in relative capacity of the model, inducing higher redundancy of learned representations. In Figure \ref{fig: within_attack_cross_strength}, we show that even within a single attack, changing the budget leads to drastic changes in the representations at deeper layers.

\subsection{Cross-threat model similarity}
In Figure \ref{fig:threat_model_grid}, we present a pairwise comparison of benign representations extracted from robust models trained on each of our threat models. This expands upon Figure 7 from the main body and demonstrates that while visual similarity leads to similar robust representations, the converse is not necessarily true. Snow and Gabor have similar representations in spite of being extremely visually different threat models (Fig. \ref{fig:adv_examples}), demonstrating the value of using RS metrics to evaluate the viability of joint robust training.

\subsection{Aligning representations across threat models}

\begin{figure}[!htb]
	\centering
	\includegraphics[width=\linewidth]{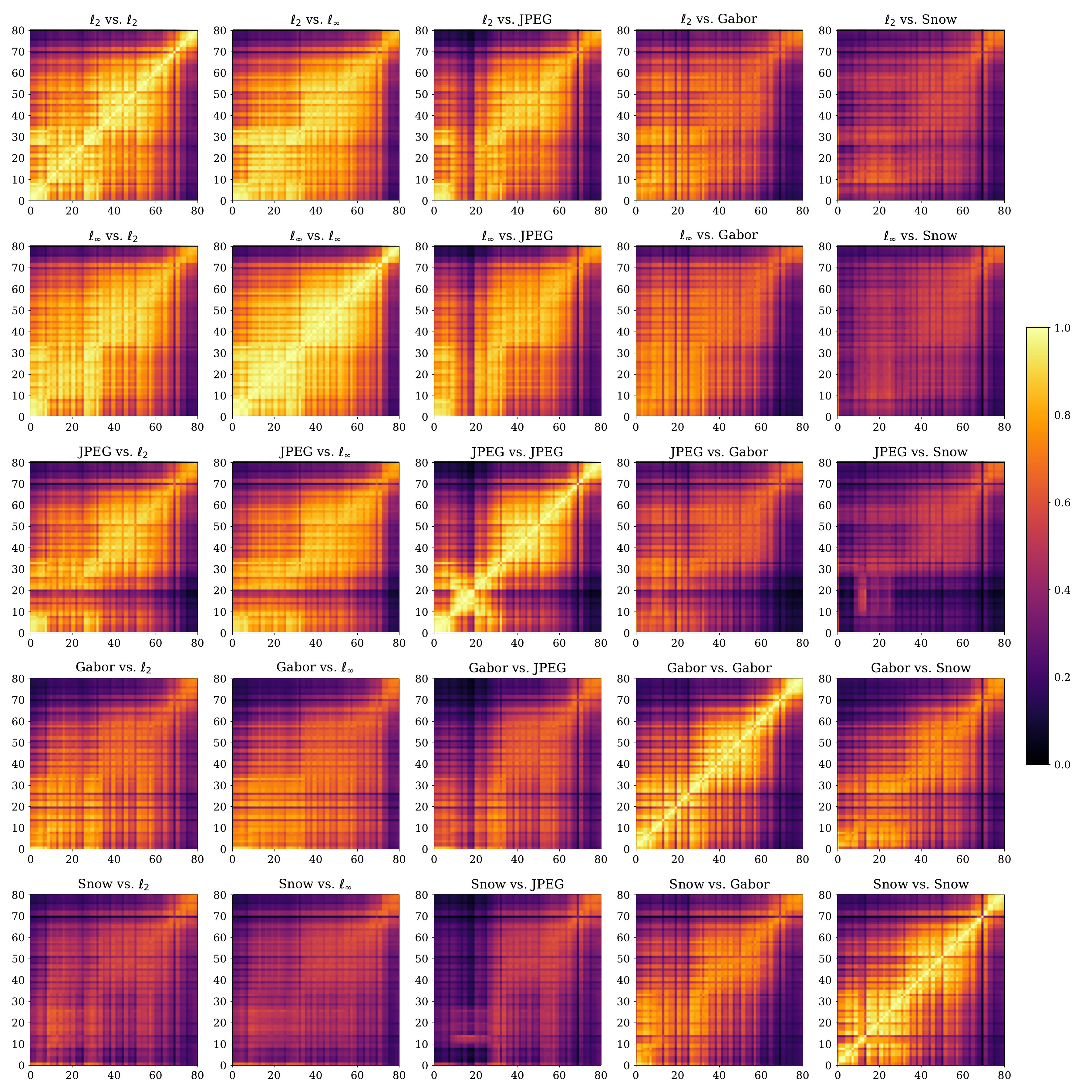}
	\caption{\textbf{Exploring relationships between threat models.} Each plot displays the layerwise similarity 
		between benign activations of robust WRN-28-10 models trained against different threat models. As suggested
		by prior research \citep{croce_adversarial_2021}, $\ell_p$ robust networks are shown to be highly similar. JPEG, which is $\ell_p$ 
		bounded in the compression space, is also noticeably more similar to the $\ell_p$ bounded attacks than to
		the others. Interestingly, Snow and Gabor trained models are noticeably more similar to each other than
		to the other models, implying an unknown similarity between the classes of attack.}
	\label{fig:threat_model_grid}
\end{figure}

\begin{figure}[!htb]
	\centering
	\includegraphics[width=\linewidth]{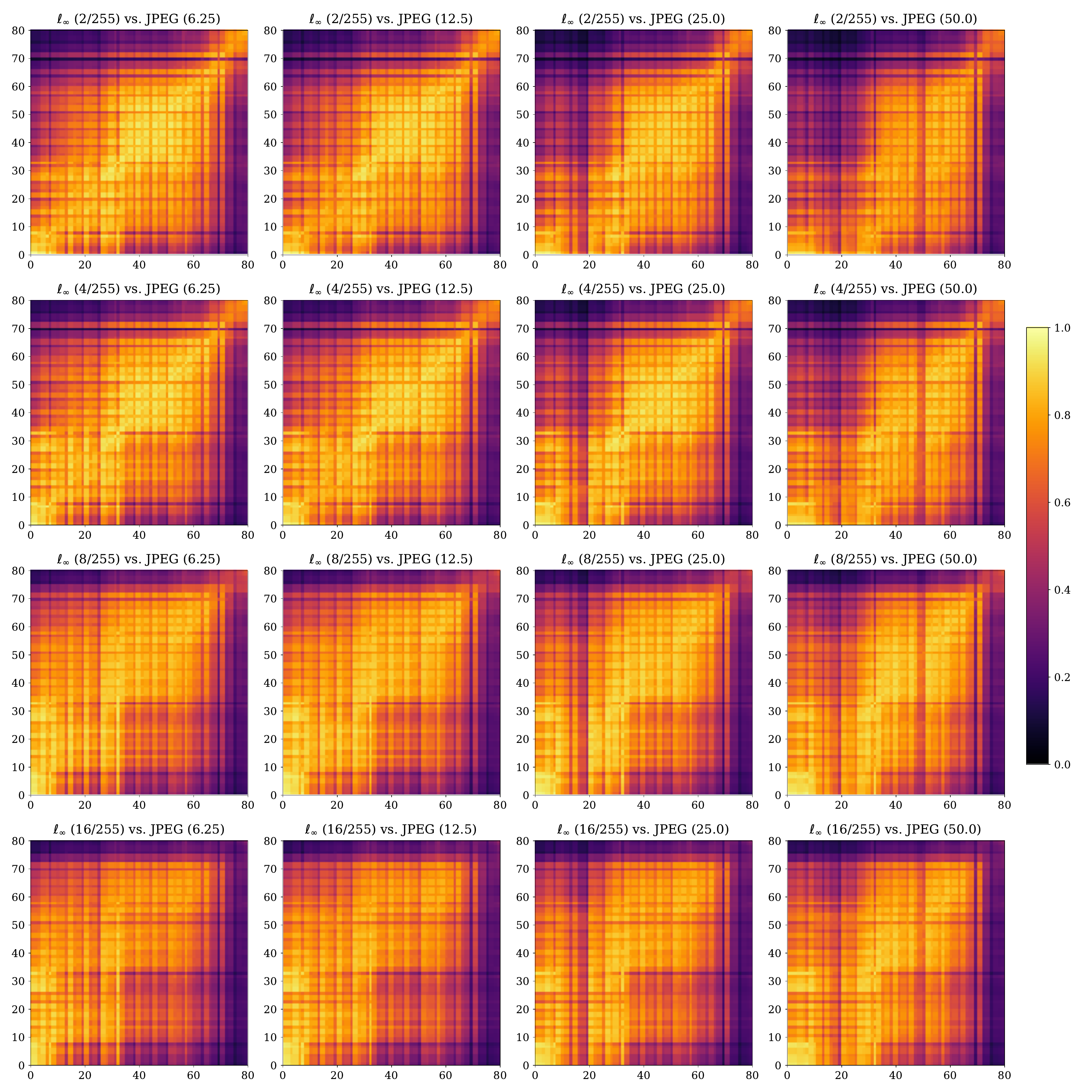}
	\caption{Comparing representations of $l_{\infty}$ and jpeg robust models at
	different attack strengths.}
	\label{fig:linf_jpeg_align}
\end{figure}

\begin{figure}[!htb]
	\centering
	\includegraphics[width=\linewidth]{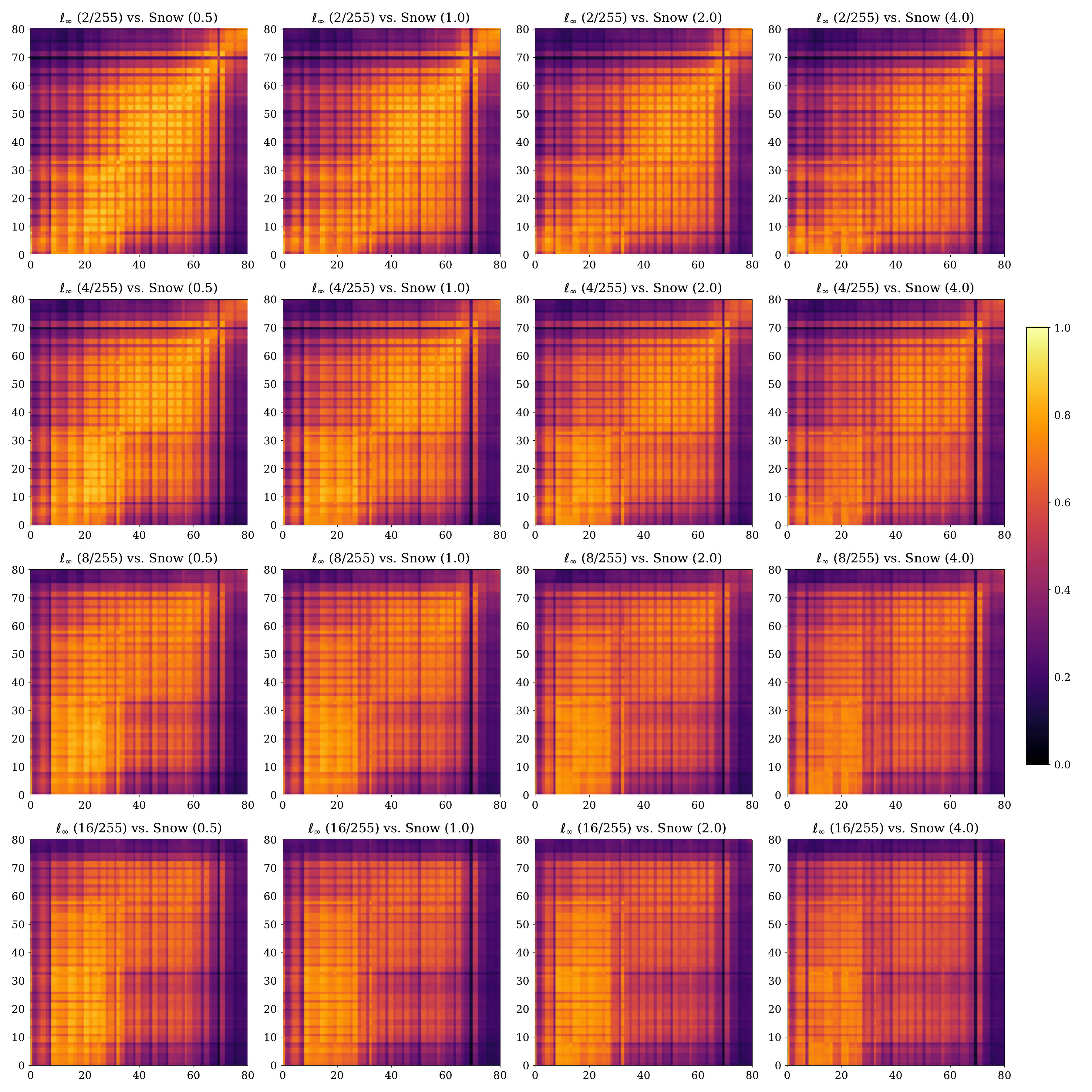}
	\caption{Comparing representations of $l_{\infty}$ and snow robust models at
	different attack strengths.}
	\label{fig:linf_snow_align}
\end{figure}

\begin{figure}[!htb]
	\centering
	\includegraphics[width=\linewidth]{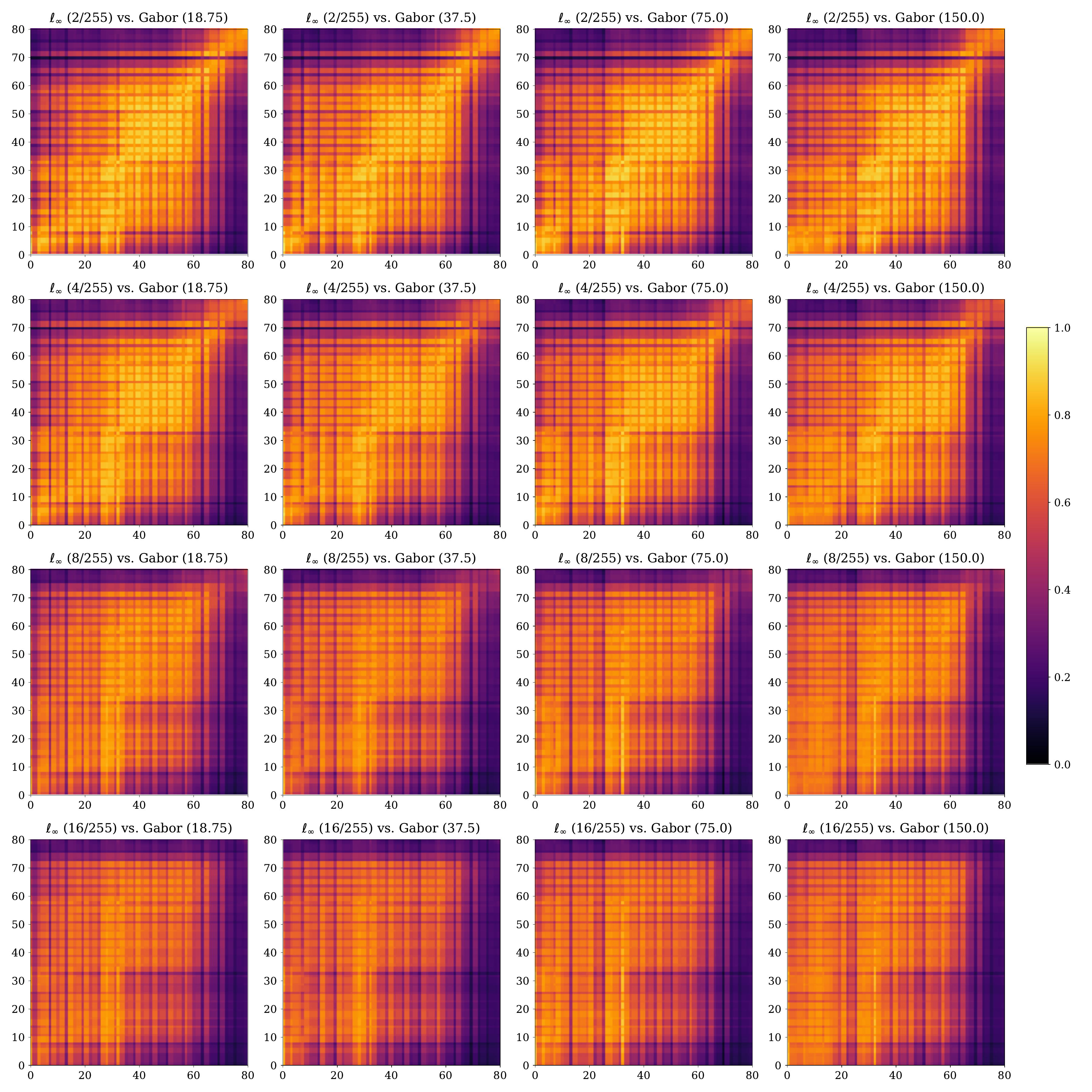}
	\caption{Comparing representations of $l_{\infty}$ and gabor robust models
	at different attack strengths.}
	\label{fig:linf_gabor_align}
\end{figure}

\begin{figure}[!htb]
	\centering
	\includegraphics[width=\linewidth]{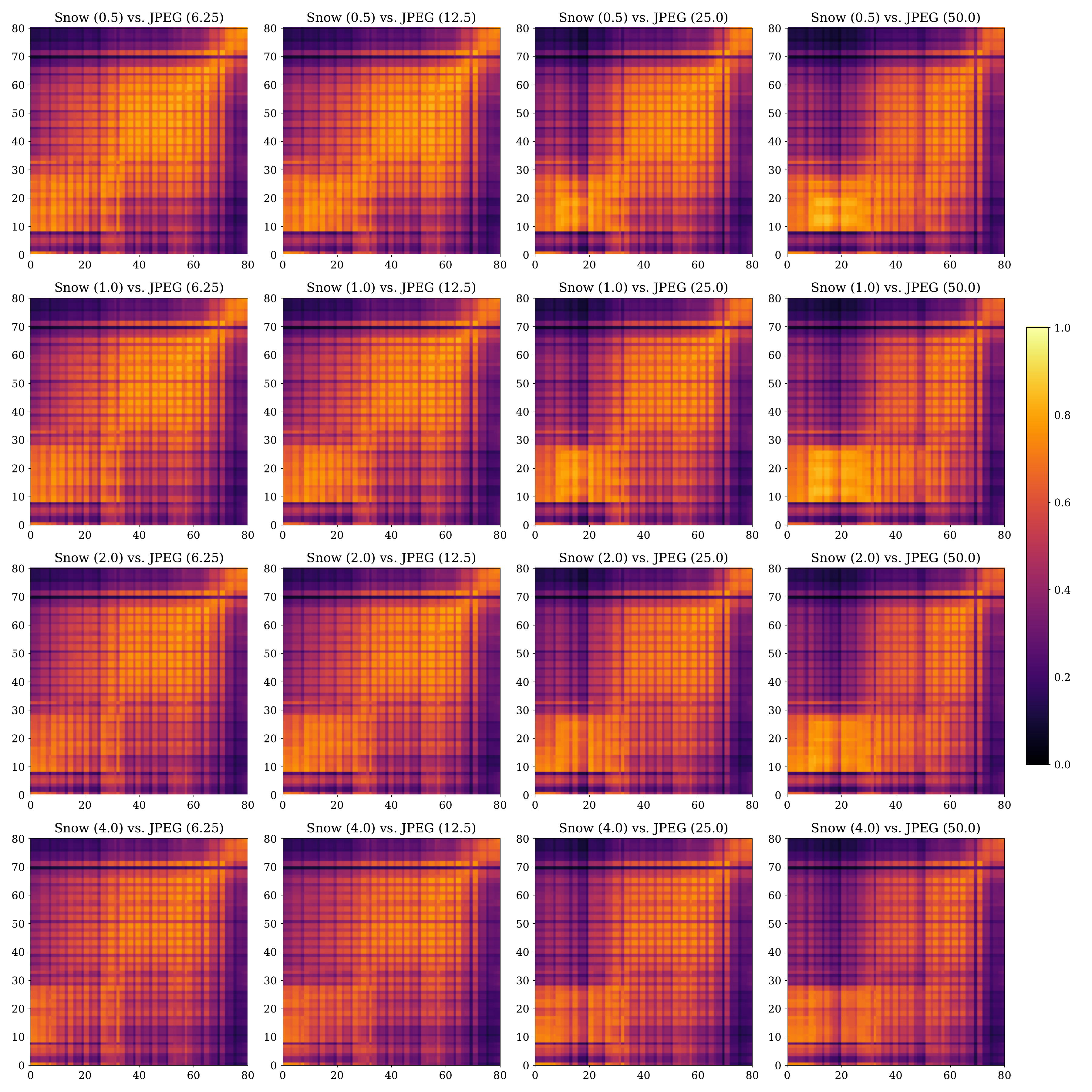}
	\caption{Comparing representations of snow and jpeg robust models at different attack strengths.}
	\label{fig:snow_jpeg_align}
\end{figure}

\begin{figure}[!htb]
	\centering
	\includegraphics[width=\linewidth]{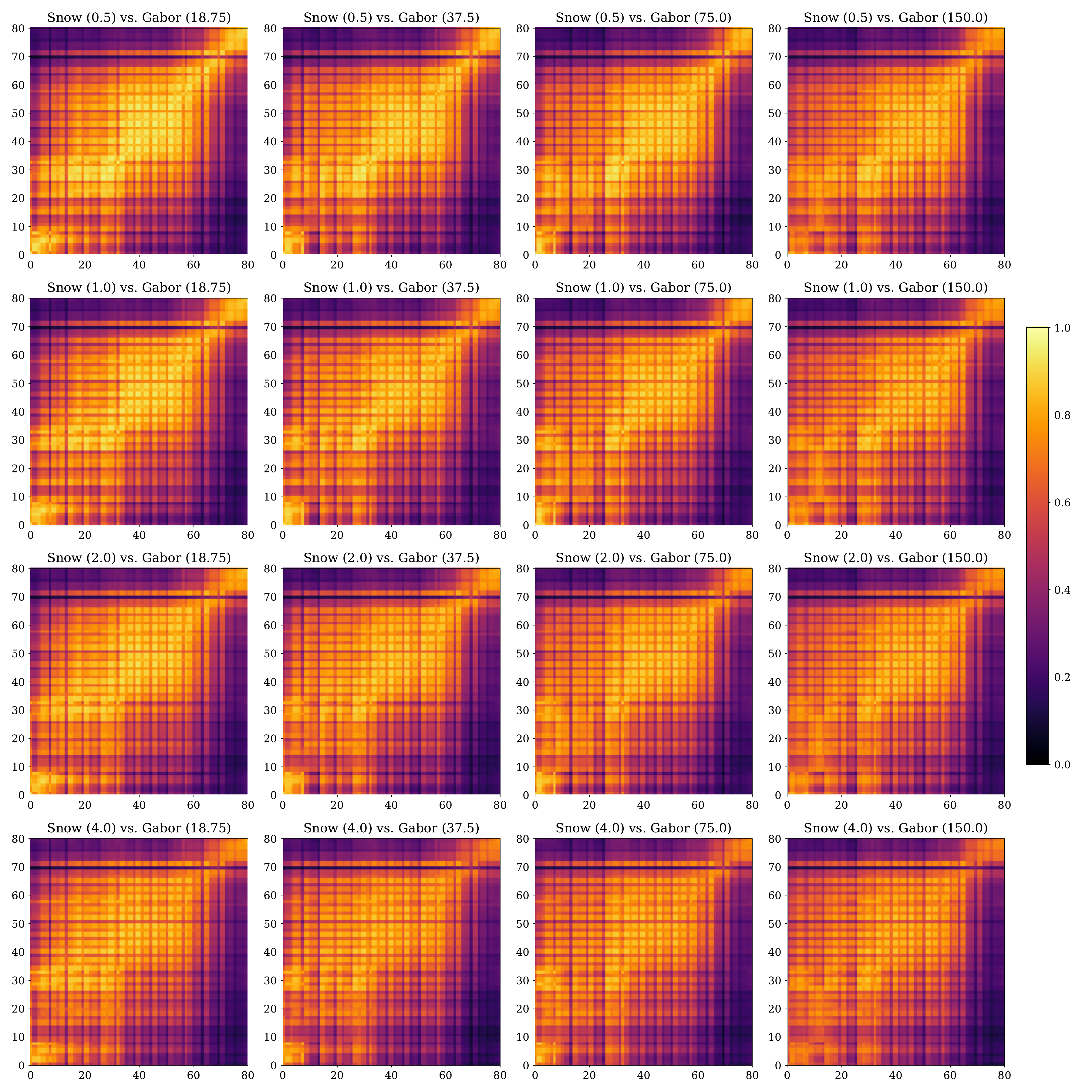}
	\caption{Comparing representations of snow and gabor robust models at different attack strengths.}
	\label{fig:snow_gabor_align}
\end{figure}

Figures \ref{fig:linf_jpeg_align} to \ref{fig:snow_gabor_align} ablate on the results of Figure \ref{fig:threat_model_grid} by varying the perturbation size of each threat model used during training. As expected, models at lower budgets tend to be better aligned. At larger budgets, even for visually similar attacks like JPEG and $\ell_2$, the aligment dips, particularly at deeper layers. This indicates that to train models jointly robust to different threat models, explicit constraints on learned representations during training may be needed. 

\end{document}